\documentclass[12pt]{article}

\newtheorem{lemma}{Lemma}
\newtheorem{theorem}{Theorem}

\usepackage{amssymb}
\usepackage{epsfig, graphicx, amsmath}
\usepackage{hyperref}
\usepackage{natbib}
\usepackage{multirow}
\usepackage{array}
\usepackage{bm, color}
\usepackage{amsmath}

\newcommand{\Mean}{{\mathbb{E}}}
\newcommand{\Var}{{\mbox{Var}}}
\newcommand{\Cov}{{\mbox{cov}}}

\newcommand{\prob}{{\mbox{Pr}}}

\newtheorem{example}{Example}
\let\code=\texttt
\let\proglang=\textsf
\DeclareMathOperator*{\argmin}{arg\,min}
\DeclareMathOperator*{\argmax}{arg\,max}
\DeclareMathOperator*{\sargmax}{sarg\,max}

\def\floor#1{\lfloor #1 \rfloor}

\usepackage{xr}
\usepackage{algpseudocode}
\usepackage[linesnumbered,ruled,vlined]{algorithm2e}

\begin{document}
	
\title{\Large{\textbf{Statistical Inference of the Value Function for Reinforcement Learning in Infinite-Horizon Settings}}}
\author{
	\bigskip
	Chengchun Shi$^\dag$, Sheng Zhang$^\ddag$, Wenbin Lu $^\ddag$ and Rui Song$^\ddag$ \\
	\normalsize{\textit{$^\dag$London School of Economics and Political Science, and $^\ddag$North Carolina State University}}
}
\date{}
\maketitle

\baselineskip=20pt


\begin{abstract}
Reinforcement learning is a general technique that allows an agent to learn an optimal policy and interact with an environment in sequential decision making problems. The goodness of a policy is measured by its value function starting from some initial state. The focus of this paper is to construct confidence intervals (CIs) for a policy's value in infinite horizon settings where the number of decision points diverges to infinity. We propose to model the action-value state function (Q-function) associated with a policy based on series/sieve method to derive its confidence interval. When the target policy depends on the observed data as well, {\color{black}we propose a {\textbf{S}}equenti\textbf{A}l \textbf{V}alue \textbf{E}valuation (SAVE) method} to recursively update the estimated policy and its value estimator. As long as either the number of trajectories or the number of decision points diverges to infinity, we show that the proposed CI achieves nominal coverage even in cases where the optimal policy is not unique. Simulation studies are conducted to back up our theoretical findings. We apply the proposed method to a dataset from mobile health studies and find that reinforcement learning algorithms could help improve patient's health status. A \proglang{Python} implementation of the proposed procedure is available at \url{https://github.com/shengzhang37/SAVE}. 
\end{abstract}

\vspace{0.1in}

\section{Introduction}
Reinforcement learning (RL) is a general technique that allows an agent to learn and interact with an environment. A policy defines the agent's way of behaving. It maps the states of environments to a set of actions to be chosen from.  
RL algorithms have made tremendous achievements and found extensive applications in video games \citep{silver2016}, robotics \citep{kormushev2013}, bidding \citep{jin2018}, ridesharing \citep{xu2018}, etc. In particular, a number of RL methods have been proposed in precision medicine, to derive an optimal policy as a set of sequential treatment decision rules that
optimize patients' clinical outcomes over a fixed period of time (finite horizon). References include \cite{Murphy2003,Zhang2013,Zhao2015,Shi2018,shi2018maximin,Zhang2018}, to name a few. 

Mobile health (or mHealth) technology has recently emerged due to the use of mobile devices such as mobile phones, tablet computers or wearable devices in health care. It allows health-care providers to communicate with patients and manage their illness in real time. It also collects rich longitudinal data (e.g., through mobile health apps) that can be used to estimate the optimal policy. Data from mHealth applications differ from those in finite horizon settings in that the number of treatment decision points for each patient is not necessarily fixed (infinite horizon) while the total number of patients could be limited. Take the OhioT1DM dataset \citep{marling2018} as an example. It contains data for six patients with type 1 diabetes. For all patients, their continuous glucose monitoring (CGM) blood glucose levels, insulin doses including bolus and basal rates, self-reported times of meals and exercises are continually measured and recorded for eight weeks. Developing an optimal policy as functions of these time-varying covariates could potentially assist these patients in improving their health status.  

In this paper, we focus on the infinite horizon setting where the data generating process is modeled by a Markov decision process \citep[MDP,][]{Puterman1994}. {\color{black}Specifically, at each time point, the agent selects an action based on the observed state. The system responds by giving the decision maker a corresponding outcome and moving into a new state in the
next time step.} 
This model is generally applicable to sequential decision making, including applications from mHealth, games, robotics, ridesharing, etc. 
After a policy is being proposed, it is important to examine its benefit prior to recommending it for practical use. 
The goodness of a policy is quantified by its (state) value function, corresponding to the discounted cumulative reward that the agent receives on average, starting from some initial state. The inference of the value function helps a decision maker to evaluate the impact of implementing a policy when the environment is in a certain state. In some applications, it is also important to evaluate the integrated value of a policy aggregated over different initial states. For example, in medical studies, one might wish to know the mean outcome of patients in the population. The integrated value could thus be used as a criterion for comparing different policies. 

In statistics literature, a few methods have been proposed to estimate the optimal policy in infinite horizons. \cite{Ertefaie2018} proposed a variant of gradient Q-learning method. 
\cite{luckett2019} proposed a V-learning to directly search the optimal policy among a restricted class of policies. Inference of the value function under a generic (data-dependent) policy has not been studied in these papers. 
In the computer science literature, \cite{thomas2015} and \cite{jiang2016} proposed (augmented) inverse propensity-score weighted ((A)IPW) estimators for the the value function in infinite horizons and derived their associated CIs. However, these methods are not suitable for settings where only a limited number of trajectories (e.g., plays of a game or patients in medical studies) are available, since (A)IPW estimators become increasingly unstable as the number of decision points diverges to infinity. 
{\color{black}Recently, \cite{kallus2019efficiently} proposed a double reinforcement learning (DRL) method that achieves consistent estimation of the value under a fixed policy even with limited number of trajectories. Their method computes a Q-function and a marginalized density ratio. Learning the density ratio is challenging in general and it remains difficult to investigate the goodness-of-fit of the estimated density ratio in practice.}

The focus of this paper is to construct confidence intervals (CIs) for a (possibly data-dependent) policy's value function at a given state as well as its integrated value with respect to a given reference distribution. 
Our proposed CI is derived by estimating the state-action value function (Q-function) under the target policy. Similar to the value, {\color{black}the Q-function measures the discounted cumulative reward that the agent receives on average, starting from some initial state-action pair.}  We use series/sieve method to approximate the Q-function based on $L$ basis functions, where $L$ grows with the total number of observations. The advances of our proposed method are summarized as follows. First, the proposed inference method is generally applicable. Specifically, it can be applied to any fixed policy (either deterministic or random) and any data-dependent policy whose value converges at a certain rate. The latter includes policies estimated by {\color{black}general Q-learning type algorithms that learns an optimal Q-function from the observed data, such as} gradient Q-learning \citep{maei2010,Ertefaie2018}, fitted Q-iteration \citep[see for example,][]{Ernst2005,riedmiller2005}, etc. See Section \ref{secpolicyapplication} for detailed illustrations. 

Second, when applied to data-dependent policies, our method is valid in nonregular cases where the optimal policy is not uniquely defined. Inference without requiring the uniqueness of the optimal policy is extremely challenging even in the simpler finite-horizon settings \citep[see the related discussions in][]{Alex2016}. The major challenge lies in that the estimated policy may not stabilize as sample size grows, making the variance of the value estimator difficult to estimate (see Section \ref{secchallenge} for details). We achieve valid inference by proposing a \textbf{S}equenti\textbf{A}l \textbf{V}alue \textbf{E}valuation (SAVE) method that splits the data into several blocks and recursively update the estimated policy and its value estimator. It is worth mentioning that the data-splitting rule cannot be arbitrarily determined since the observations are time dependent in infinite horizon settings (see Section \ref{secCI} for details).

Third, 
our CI is valid as long as either the number of trajectories $n$ in the data, or the number of decision points $T$ per trajectory diverges to infinity. It can thus be applied to a wide variety of real applications in infinite horizons ranging from the Framingham heart study \citep{tsao2015} with over two thousand patients to the OhioT1DM dataset that contains eight weeks' worth of data for six people. 
We also allow both $n$ and $T$ to approach infinity, which is the case in applications from video games. 
In contrast, CIs proposed by \cite{thomas2015} and \cite{jiang2016} require $n$ to grow to infinity to achieve nominal coverage.
	
Lastly, we consider both off-policy and on-policy learning methods. In off-policy settings, CIs are derived based on historical data collected by a potentially different behavior policy. Off-policy evaluation is critical in situations where running the target policy could be expensive, risky or unethical. In on-policy settings, the estimated policy is recursively updated as batches of new observations arrive. To our knowledge, this is the first work on statistical inference of a data-dependent policy in on-policy settings {\color{black}in sequential decision making with infinite horizons}.


To study the asymptotic properties of our proposed CI, we focus on tensor-product spline and wavelet series estimators. 
Our technical contributions are described as follows. First, we introduce a bidirectional-asymptotic framework that allows either $n$ or $T$ to approach infinity. 
Our major technical contribution is to derive a nonasymptotic error bound for the spectral norm of sums of mean zero random matrices formed by the data transactions from MDP 
as a function of $n$, $T$ and $L$ (see e.g., Lemma \ref{lemma3}). This result is important in studying the limiting distribution of series estimators under such a theoretical framework. 

Second, for policies that are estimated by Q-learning type algorithms such as the greedy gradient Q-learning, fitted Q-iteration and deep Q-network \citep{mnih2015},  
we relate the convergence rate of their values to the prediction error of the corresponding estimated Q-functions. We show in Theorems \ref{thm3} and \ref{thm4} that the values can converge at faster rates than the estimated Q-functions under certain margin type conditions on the optimal Q-function. {\color{black}To the best of our knowledge, these findings have not been discovered in the reinforcement-learning literature. Our theorems form a basis for researchers to study the value properties of Q-learning type algorithms.} Moreover, our theoretical results are consistent with findings in point treatment studies where there is only one single decision point \citep[see e.g.,][]{qian2011,Alex2016}. However, the derivation of Theorems \ref{thm3} and \ref{thm4} is more involved since the value function in our settings is an infinite series involving both immediate and future rewards.

Third, when these basis functions are used, we mathematically characterize the approximation error of the Q-function as a function of $L$, the dimension of the state variables, and the smoothness of the Markov transition function and the conditional mean of the immediate reward as a function of the state-action pair. 
This offers some guidance to practitioners on the choice of the number of basis functions $L$, when some prior knowledge on the degree of smoothness of the aforementioned functions are available. 

The rest of the paper is organized as follows. We introduce the model setup in Section \ref{secsetup}. In Sections \ref{secoffpolicy} and \ref{seconpolicy}, we present the proposed off-policy and on-policy evaluation methods, respectively.
Simulation studies are conducted to evaluate the empirical performance of the proposed inference methods in Section \ref{secsimu}. We apply the proposed inference method to the OhioT1DM dataset in Section \ref{secrealdata}, Finally,  we conclude our paper by a discussion section. 



\section{Optimal policy in infinite-horizon settings}\label{secsetup}
We begin by introducing the notion of the optimal policy, the Q-function and the value function in infinite-horizon settings. Let $X_{0,t}\in \mathbb{X}$ be the time-varying covariates collected at time point $t$, $A_{0,t}\in \mathcal{A}$ denote the action taken at time $t$, and $Y_{0,t}$ stand for the immediate reward observed. {\color{black}Here, $\mathbb{X}$ and $\mathcal{A}$ denote the state and action space, respectively. We assume $\mathbb{X}$ is a subspace of $\mathbb{R}^d$ where $d$ is the number of state vectors and $\mathcal{A}$ is a discrete space $\{0,1,\cdots,m-1\}$ where $m$ denotes the number of actions.} Suppose the system satisfies the following Markov assumption (MA),
\begin{eqnarray*}
	\prob(X_{0,t+1}\in \mathcal{B}|X_{0,t}=x,A_{0,t}=a,\{Y_{0,j}\}_{0\le j<t},\{X_{0,j}\}_{0\le j<t}, \{A_{0,j}\}_{0\le j<t})=\mathcal{P}(\mathcal{B}|x,a),
\end{eqnarray*}
for some transition function $\mathcal{P}$. {\color{black}Here, $\mathcal{P}$ defines the next state distribution conditional on the current state-action pair.} Moreover, suppose the following conditional mean independence assumption (CMIA) holds
\begin{eqnarray*}
	\Mean (Y_{0,t}|X_{0,t}=x,A_{0,t}=a,\{Y_{0,j}\}_{0\le j<t},\{X_{0,j}\}_{0\le j<t}, \{A_{0,j}\}_{0\le j<t})
	\\= \Mean (Y_{0,t}|X_{0,t}=x,A_{0,t}=a)=r(x,a),
\end{eqnarray*}
for some reward function $r$. 
By MA, CMIA automatically holds when $Y_{0,t}$ is a deterministic function of $X_{0,t},A_{0,t}$ and $X_{0,t+1}$ that measures the system's status at time $t+1$. The latter is satisfied in our real data application (see Section \ref{secrealdata} for details) and {\color{black}is commonly assumed in the reinforcement learning literature. CMIA is thus weaker than this condition.} MA and CMIA are important to guarantee the existence of an optimal policy (see \eqref{existenceoptimal}) and derive the bidirectional-asymptotic theory of the proposed CI (see the discussions below Theorem \ref{thm1}). We assume both assumptions hold throughout this paper.


In the following, we focus on the class of stationary policies
that map the covariate space $\mathbb{X}$ to probability mass functions on $\mathcal{A}$. Let $\pi(\cdot|\cdot)$ denote such a policy. It satisfies $\pi(a|x)\ge 0$, for any $a\in \mathcal{A},x\in \mathbb{X}$ and $\sum_{a\in \mathcal{A}}\pi(a|x)=1$, for any $x\in \mathbb{X}$. For a deterministic policy, we have $\pi(a|x)\in \{0,1\}$, for any $a\in \mathcal{A},x\in \mathbb{X}$. Under $\pi$, a decision maker will set $A_{0,t}=a$ with probability $\pi(a|X_{0,t})$ at time $t$. For such a policy and a given discounted factor $0\le \gamma<1$, let $V(\pi;x)$ denote the value function
\begin{eqnarray*}
	V(\pi;x)=\sum_{t\ge 0} \gamma^{t} \Mean^{\pi} (Y_{0,t}|X_{0,0}=x),
\end{eqnarray*}
where the expectation $\Mean^{\pi}$ is taken by assuming that the system follows the policy $\pi$. The rate $\gamma$ reflects a trade-off between immediate and future rewards. If $\gamma=0$, the agent tends to choose actions that maximize the immediate reward. As $\gamma$ increases, the agent will consider future rewards more seriously. Under CMIA, we have
\begin{eqnarray*}
	V(\pi;x)=\sum_{t\ge 0} \gamma^{t} \Mean^{\pi} \{\Mean (Y_{0,t}|X_{0,t},A_{0,t})|X_{0,0}=x\}=\sum_{t\ge 0} \gamma^{t} \sum_{a\in \mathcal{A}} \Mean^{\pi} \{\pi(a|X_{0,t})  r(X_{0,t},a)|X_{0,0}=x\}.
\end{eqnarray*}
{\color{black}Similar to Theorem 6.2.12 of \cite{Puterman1994}, we can show under the given conditions that there exists at least one optimal policy $\pi^{\tiny{opt}}$ that satisfies}
\begin{eqnarray}\label{existenceoptimal}
	V(\pi^{\tiny{opt}};x)\ge V(\pi;x),\,\,\,\,\,\,\,\,\forall \pi,x.
\end{eqnarray}

To better understand $\pi^{\tiny{opt}}$, we introduce the state-action function (Q-function) under a policy $\pi$ as
\begin{eqnarray*}
	Q(\pi;x,a)=\sum_{t\ge 0} \gamma^{t} \Mean^{\pi} (Y_{0,t}|X_{0,0}=x,A_{0,0}=a).
\end{eqnarray*}
Let $Q^{\tiny{opt}}$ denote the optimal Q-function, i.e, $Q^{\tiny{opt}}(\cdot,\cdot)=\sup_{\pi} Q(\pi;\cdot,\cdot)$. It can be shown that $\pi^{\tiny{opt}}$ satisfies
\begin{eqnarray*}
	\pi^{\tiny{opt}}(a|x)=0\,\,\,\,\hbox{if}\,\,\,\,a\notin \argmax_{a'} Q^{\tiny{opt}}(x,a'),\,\,\,\,\forall x,a.
\end{eqnarray*}
There exist infinitely many optimal policies when $\argmax_{a} Q^{\tiny{opt}}(x,a)$ is not unique for some $x\in \mathbb{X}$. Let $\Pi^{\tiny{opt}}$ denote the set consisting of all these optimal policies. Define 
\begin{eqnarray}\label{pi0}
	\pi^{\tiny{opt}}_0(a|x)=\left\{\begin{array}{ll}
		1, & \hbox{if}\,\,\,\,a= \sargmax_{a'} Q^{\tiny{opt}}(x,a'),\\
		0, & \hbox{otherwise},
	\end{array}
	\right.
\end{eqnarray} 
where $\sargmax$ denotes the smallest maximizer when the argmax is not unique. Such a deterministic optimal policy may be appealing in medical studies. For example, in optimal dose studies, it is preferred to assign each patient the smallest optimal dose level to avoid toxicity. 

\section{Off-policy evaluation}\label{secoffpolicy}
\subsection{Inference of the value under a fixed policy}\label{secvaluefixedpolicy}
Let $n$ denote the number of trajectories in the dataset. 
For the $i$-th trajectory, let $\{A_{i,t}\}_{t\ge 0}$, $\{X_{i,t}\}_{t\ge 0}$ and $\{Y_{i,t}\}_{t\ge 0}$ denote the sequence of actions, states and rewards, respectively. It is worth mentioning that the time points are not necessarily homogeneous across different trajectories. Suppose the data are generated according to a fixed policy $b(\cdot|\cdot)$, better known as the behavior policy such that
\begin{eqnarray*}
	\{(X_{1,t},A_{1,t},Y_{1,t})\}_{t\ge 0}, \{(X_{2,t},A_{2,t},Y_{2,t})\}_{t\ge 0},\cdots, \{(X_{n,t},A_{n,t},Y_{n,t})\}_{t\ge 0},
\end{eqnarray*}
are i.i.d copies of $\{(X_{0,t},A_{0,t},Y_{0,t})\}_{t\ge 0}$. 
The observed data can thus be summarized as $\{(X_{i,t},A_{i,t},Y_{i,t},X_{i,t+1})\}_{0\le t< T_i,1\le i\le n}$, where $T_i$ is the termination time of the $i$-th trajectory. The goal of off-policy evaluation is to learn the value under a target policy $\pi(\cdot|\cdot)$, possibly different from $b(\cdot|\cdot)$. 

\subsubsection{Modelling value or Q-function?}\label{sec:vorQ}

\cite{luckett2019} showed that the value function satisfies
\begin{eqnarray}\label{vBell1}
	\Mean \left[\left.\frac{\pi(A_{i,t},X_{i,t})}{b(A_{i,t},X_{i,t})}\{Y_{i,t}+\gamma V(\pi;X_{i,t+1})-V(\pi;X_{i,t}) \}\right|X_{i,t}\right]=0.
\end{eqnarray}
{\color{black}Based on \eqref{vBell1}, they directly modelled the value function, constructed an estimator for the integrated value under their estimated optimal policy and proved that it is asymptotically normal \citep[see Theorem 4.3,][]{luckett2019}.}

{\color{black}Following their procedure}, for a fixed policy $\pi$, one might estimate $V(\pi;\cdot)$ nonparametrically and construct the CI using the resulting estimate. However, such an approach might not be appropriate for polices that are discontinuous functions of the covariates. 
To better illustrate this, notice that $V(\pi,\cdot)$ satisfies the following Bellman equation
\begin{eqnarray}\label{vBell2}
	V(\pi;x)=\sum_{a\in \mathcal{A}}\pi(a|x) \underbrace{\left\{r(x,a)+\gamma \int_{x'} V(\pi;x') \mathcal{P}(dx'|x,a)\right\}}_{C(\pi;x,a)}.
\end{eqnarray}
{\color{black}When $\mathcal{P}$ satisfies certain smoothness conditions (see Condition A1 below), we have
\begin{eqnarray}\label{vBell3}
	&& \left|\int_{x'} V(\pi;x')\mathcal{P}(dx'|x_1,a)-\int_{x'} V(\pi;x')\mathcal{P}(dx'|x_2,a)\right|\\ \nonumber
	&\le& \int_{x'} |V(\pi;x')| |\mathcal{P}(dx'|x_1,a)-\mathcal{P}(dx'|x_2,a)|\to 0,\,\,\,\,\hbox{as}~\|x_1-x_2\|_2\to 0,
\end{eqnarray}
for any $\pi$. Suppose $r(\cdot,a)$ is continuous for any $a\in \mathcal{A}$. Then $C(\pi;x,a)$ is continuous in $x$ for any $\pi$ and $a$. When $\pi$ is a non-continuous function of $x$, it follows from \eqref{vBell2} that $V(\pi;\cdot)$ is not continuous either.} However, many nonparametric methods, such as kernel smoothers and series estimation, require the underlying function to possess certain degree of smoothness in order to achieve estimation consistency. Notice that any non-constant deterministic policy has jumps and is not continuous at certain points (such as the optimal policy $\pi_0$ given in  \eqref{pi0}). This poses significant challenges in performing inference to these policies. 


To allow valid inference for both deterministic and random policies, we consider modelling the Q-function. Under CMIA, we have
\begin{eqnarray*}
	Q(\pi;x,a)=\sum_{t\ge 0} \gamma^{t} \Mean^{\pi} \{r(X_{0,t},A_{0,t})|X_{0,0}=x,A_{0,0}=a\}.
\end{eqnarray*}
This together with MA yields
\begin{eqnarray*}
	Q(\pi;x,a)=r(x,a)+\gamma  \Mean^{\pi} \left[\left.\sum_{t\ge 0} \gamma^{t}\{\Mean^{\pi} r(X_{0,t},A_{0,t})|X_{0,1},A_{0,1}\}\right|X_{0,0}=x,A_{0,0}=a \right]\\
	=r(x,a)+\gamma \Mean^{\pi} \{Q(X_{0,1},A_{0,1})|X_{0,0}=x,A_{0,0}=a\}.
\end{eqnarray*}
As a result, the Q-function satisfies the following Bellman equation
\begin{eqnarray}\label{QBell1}
	Q(\pi;x,a)=r(x,a)+\gamma \sum_{a'\in \mathcal{A}}\int_{x'} Q(\pi;x',a')\pi(a'|x')\mathcal{P}(dx'|x,a). 
\end{eqnarray}
Similar to \eqref{vBell3}, we can show the second term on the right-hand-side (RHS) of \eqref{vBell3} is a smooth function of $x$ for any $\pi$ and $a$. When $r(\cdot,a)$ is smooth, so is $Q(\pi,\cdot,a)$. To formally establish these results, we introduce the notion of $p$-smoothness ({\color{black}also known as H{\"o}lder smoothness with exponent $p$}) below.

Let $h(\cdot)$ be an arbitrary function on $\mathbb{X}$. For a $d$-tuple $\alpha=(\alpha_1,\dots,\alpha_d)^{\top}$ of nonnegative integers, let $D^{\alpha}$ denote the differential operator:
\begin{eqnarray*}
	D^{\alpha}h(x)=\frac{\partial^{\|\alpha\|_1} h(x)}{\partial x_1^{\alpha_1}\cdots\partial x_d^{\alpha_d}}.
\end{eqnarray*}
Here, $x_j$ denotes the $j$-th element of $x$. For any $p>0$, let $\floor{p}$ denote the largest integer that is smaller than $p$. Define the class of $p$-smooth functions as follows:
\begin{eqnarray*}
	\Lambda(p,c)=\left\{h:\sup_{\|\alpha\|_1\le \floor{p}} \sup_{x\in \mathbb{X}} |D^{\alpha} h(x)|\le c, \sup_{\|\alpha\|_1=\floor{p}} \sup_{\substack{x,y\in \mathbb{X}\\ x\neq y}} \frac{|D^{\alpha} h(x)-D^{\alpha} h(y)|}{\|x-y\|_2^{p-\floor{p}}}\le c \right\}. 
\end{eqnarray*}
{\color{black}When $0<p\le 1$, we have $\floor{p}=0$. It is equivalent to require $h$ to satisfy $\sup_{x,y} |h(x)-h(y)|/\|x-y\|_2^p\le c$. The notion of $p$-smoothness is thus reduced to the H{\"o}lder continuity.}

For any $x\in \mathbb{X}$, $a\in \mathcal{A}$, suppose the transition kernel $\mathcal{P}(\cdot|x,a)$ is absolutely continuous with respect to the Lebesgue measure. Then there exists some transition density function $q$ such that $\mathcal{P}(dx'|x,a)=q(x'|x,a)dx'$. We impose the following condition.

\smallskip

\noindent (A1.) There exist some $p,c>0$ such that $r(\cdot,a), q(x'|\cdot,a) \in \Lambda(p,c)$ for any $a\in \mathcal{A},x'\in \mathbb{X}$.

\begin{lemma}\label{lemma1}
	Under A1, there exists some constant $c'>0$ such that $Q(\pi;\cdot,a)\in \Lambda(p,c')$ for any policy $\pi$ and $a\in \mathcal{A}$.
\end{lemma}

{\color{black}Lemma \ref{lemma1} implies the Q-function has bounded derivatives up to order $\floor{p}$.} This motivates us to first estimate the Q-function and then derive the corresponding value estimators based on the relation $V(\pi;x)=\sum_{a\in \mathcal{A}} \pi(a|x) Q(\pi;x,a)$. 
By the Bellman equation \eqref{QBell1}, we can show the Q-function satisfies
\begin{eqnarray}\label{QBell2}
	\Mean \left[\left.\left\{Y_{i,t}+\gamma \sum_{a\in \mathcal{A}} Q(\pi;X_{i,t+1},a)\pi(a|X_{i,t+1})-Q(\pi;X_{i,t},A_{i,t}) \right\}\right|X_{i,t},A_{i,t}\right]=0.
\end{eqnarray}

{\color{black}The above equation forms a basis of our methods to learn $Q(\pi;\cdot,\cdot)$ (see details in the next section).} In contrast to Equation \eqref{vBell1}, the sampling ratio $\pi(a|x)/b(a|x)$ does not appear in \eqref{QBell2}. This is because $A_{i,t}$ is the only sampling action and no further actions are involved in \eqref{QBell2}. As a result, our method does not require correct specification of the behavior policy. Nor do we need to estimate it from the observed dataset. This is another advantage of modelling the Q-function over the value.

\subsubsection{Method}\label{sec:method}
We describe our procedure in this section. We propose to approximate $Q(\pi;\cdot,\cdot)$ based on linear sieves, which takes the form
\begin{eqnarray*}
	Q(\pi;x,a)\approx \Phi_L^{\top}(x) \beta_{\pi,a},\,\,\,\,\forall x\in \mathbb{X},a\in \mathcal{A},
\end{eqnarray*}
where $\Phi_L(\cdot)=\{\phi_{L,1}(\cdot),\cdots,\phi_{L,L}(\cdot)\}^{\top}$ is a vector consisting of $L$ sieve basis functions, such as splines or wavelet bases \citep[see for example,][for choices of basis functions]{Huang1998}. We allow $L$ to grow with the sample size to reduce the bias of the resulting estimates. 
Under certain mild conditions, there exist some $\{\beta_{\pi,a}^*\}_{a\in \mathcal{A}}$ that satisfy
\begin{eqnarray*}
	\Mean \left\{Y_{i,t}+\gamma \sum_{a\in \mathcal{A}} \Phi_L^{\top}(X_{i,t+1}) \beta_{\pi,a}^*\pi(a|X_{i,t+1})-\Phi_L^{\top}(X_{i,t}) \beta_{\pi,a'}^* \right\} \Phi_L(X_{i,t})\mathbb{I}(A_{i,t}=a')= 0,
\end{eqnarray*}
for any $a'\in \mathcal{A}$. Recall that $\mathcal{A}=\{0,1,\dots,m-1\}$. Define $\bm{\beta}^*_{\pi}=(\beta_{\pi,1}^{*T},\cdots,\beta_{\pi,m}^{*T})^{\top}$, 
\begin{eqnarray*}
	\bm{\xi}(x,a)&=&\{\Phi_L^{\top}(x)\mathbb{I}(a=0),\Phi_L^{\top}(x)\mathbb{I}(a=1),\cdots,\Phi_L^{\top}(x)\mathbb{I}(a=m-1)\}^{\top},\\
	\bm{U}_{\pi}(x)&=&\{\Phi_L^{\top}(x)\pi(0|x),\Phi_L^{\top}(x)\pi(1|x),\cdots,\Phi_L^{\top}(x)\pi(m-1|x)\}^{\top},
\end{eqnarray*}
$\bm{\xi}_{i,t}=\bm{\xi}(X_{i,t},A_{i,t})$, $\bm{U}_{\pi,i,t}=\bm{U}_{\pi}(X_{i,t})$. The above equation can be rewritten as $\Mean \bm{\xi}_{i,t}(\bm{\xi}_{i,t}-\gamma \bm{U}_{\pi,i,t+1})^{\top} \bm{\beta}_{\pi}^* = \Mean \bm{\xi}_{i,t}Y_{i,t}$. Based on the observed data, 
we propose to estimate $\bm{\beta}^*_{\pi}$ by solving
\begin{eqnarray*}
	\widehat{\bm{\beta}}_{\pi}=\biggl\{\underbrace{\frac{1}{\sum_{i} T_i} \sum_{i=1}^n \sum_{t=0}^{T_i-1} \bm{\xi}_{i,t} (\bm{\xi}_{i,t}- \gamma \bm{U}_{\pi,i,t+1})^{\top}}_{\widehat{\bm{\Sigma}}_{\pi}}\biggr\}^{-1} \left( \frac{1}{\sum_{i} T_i}\sum_{i=1}^n \sum_{t=0}^{T_i-1} \bm{\xi}_{i,t}Y_{i,t}  \right).
\end{eqnarray*}
Let $\widehat{\bm{\beta}}_{\pi}=(\widehat{\beta}_{\pi,1}^{\top},\cdots,\widehat{\beta}_{\pi,m}^{\top})^{\top}$, we propose to estimate $V(\pi;x)$ by $$\widehat{V}(\pi;x)=\sum_{a\in \mathcal{A}} \Phi_L^{\top} (x) \widehat{\beta}_{\pi,a} \pi(a|x)=\bm{U}_{\pi}^{\top}(x) \widehat{\bm{\beta}}_{\pi}.$$ A two-side CI is given by 
\begin{eqnarray}\label{CIvalue}
	[\widehat{V}(\pi;x)-z_{\alpha/2} (\sum_{i} T_i)^{-1/2} \widehat{\sigma}(\pi;x), \widehat{V}(\pi;x)+z_{\alpha/2} (\sum_{i} T_i)^{-1/2} \widehat{\sigma}(\pi;x)],
\end{eqnarray}
where $z_{\alpha}$ denotes the upper $\alpha$-th quantile of a standard normal distribution, and
\begin{eqnarray*}
	\widehat{\sigma}^2(\pi;x)=\bm{U}_{\pi}^{\top}(x) \widehat{\bm{\Sigma}}_{\pi}^{-1} \widehat{\bm{\Omega}}_{\pi}(\widehat{\bm{\Sigma}}^{\top}_{\pi})^{-1} \bm{U}_{\pi}(x),
\end{eqnarray*}
where
\begin{eqnarray*}
	\widehat{\bm{\Omega}}_{\pi}=\frac{1}{\sum_{i} T_i}\sum_{i=1}^n \sum_{t=0}^{T_i-1} \bm{\xi}_{i,t} \bm{\xi}_{i,t}^{\top} \{Y_{i,t}+\gamma \sum_{a\in \mathcal{A}} \Phi_L^{\top}(X_{i,t+1}) \widehat{\beta}_{\pi,a}\pi(a|X_{i,t+1})-\Phi_L^{\top}(X_{i,t}) \widehat{\beta}_{\pi,A_{i,t}}\}^2.
\end{eqnarray*}

Let $\mathbb{G}$ be a reference distribution on the covariate space $\mathbb{X}$. Define the following integrated value function 
$$V(\pi;\mathbb{G})=\int_{x\in \mathbb{X}} V(\pi;x)\mathbb{G}(dx).$$
By setting $\mathbb{G}(\cdot)$ to be a Dirac measure $\delta_x(\cdot)$, i.e, $\mathbb{G}(\mathcal{X})=\mathbb{I}(x\in \mathcal{X}),\forall \mathcal{X}\subseteq \mathbb{X}$, $V(\pi;\mathbb{G})$ is reduced to $V(\pi;x)$. Let $\nu_0(\cdot)$ be the probability density function of $X_{0,0}$. By setting $\mathbb{G}(dx)=\nu_0(x)dx$, we obtain
\begin{eqnarray*}
	V(\pi;\mathbb{G})=\int_{x\in \mathbb{X}} V(\pi;x)\nu_0(x)dx.
\end{eqnarray*}
Based on $\widehat{\bm{\beta}}_{\pi}$, a two-side CI for $V(\pi;\mathbb{G})$ is given by
\begin{eqnarray}\label{CIaveragevalue}
	[\widehat{V}(\pi;\mathbb{G})-z_{\alpha/2} (\sum_{i} T_i)^{-1/2} \widehat{\sigma}(\pi;\mathbb{G}),\widehat{V}(\pi;\mathbb{G})+z_{\alpha/2} (\sum_{i} T_i)^{-1/2} \widehat{\sigma}(\pi;\mathbb{G})],
\end{eqnarray}
where
\begin{eqnarray}\label{hatVpiG}
	&&\widehat{V}(\pi;\mathbb{G})=\int_{x\in \mathbb{X}} \widehat{V}(\pi;x)\mathbb{G}(dx),\\\label{hatsigmapiG}
	&&\widehat{\sigma}^2(\pi;\mathbb{G})=\left\{\int_{x\in \mathbb{X}} \bm{U}_{\pi}(x)\mathbb{G}(dx)\right\}^{\top} \widehat{\bm{\Sigma}}_{\pi}^{-1} \widehat{\bm{\Omega}}_{\pi}(\widehat{\bm{\Sigma}}_{\pi}^{\top})^{-1} \left\{\int_{x\in \mathbb{X}} \bm{U}_{\pi}(x)\mathbb{G}(dx)\right\}.
\end{eqnarray}

\subsubsection{Theory}\label{secvaluefixedtheory}
In this section, we focus on proving the validity of the proposed CIs in \eqref{CIaveragevalue}. By setting $\mathbb{G}(\cdot)=\delta_x(\cdot)$, it implies that the CI in \eqref{CIvalue} achieves nominal coverage as well. 
To simplify the presentation, we assume $T_1=T_2=\cdots=T_n=T$, all the covariates are continuous and $\mathbb{X}=[0,1]^d$. Our theory is valid regardless of whether $T$ is bounded or diverges to infinity. We remark that the boundedness of $T$ does not mean we work on a finite-horizon setting, since $T$ is the termination time of the study, not the final time step of each trajectory.

In addition, we restrict our attentions to two particular types of sieve basis functions, corresponding to tensor product of B-splines 
with degree $r$ and dimension $L$ or Wavelets 
with regularity $r$ and dimension $L$. See Section 6 of \cite{Chen2015} for a brief review of these sieve bases. This together with A1 implies that there exists a set of vectors $\{\beta_{\pi,a}^*\}_{a\in \mathcal{A}}$ that satisfy $\sup_{x\in \mathbb{X},a\in \mathcal{A}} |Q(\pi;x,a)-\Phi_L^{\top}(x) \beta_{\pi,a}^*|=O(L^{-p/d})$. See Section 2.2 of \cite{Huang1998} for detailed discussions on the approximation power of these sieve bases. 

Following the behavior policy $b(\cdot|\cdot)$, the set of variables $\{X_{0,t}\}_{t\ge 0}$ forms a time-homogeneous Markov chain. Its transition kernel $\mathcal{P}_X$ is given by
\begin{eqnarray*}
	\mathcal{P}_X(\mathcal{B}|x)=\hbox{Pr}(X_{0,1}\in\mathcal{B}|X_{0,0}=x)=\sum_{a\in \mathcal{A}} \mathcal{P}(\mathcal{B}|x,a)b(a|x),\,\,\,\,\forall \mathcal{B}\in \mathbb{X}.
\end{eqnarray*}
We impose the following assumptions.

\smallskip

\noindent (A2.) The Markov chain $\{X_{0,t}\}_{t\ge 0}$ has an unique invariant distribution with some density function $\mu(\cdot)$. The density functions $\mu$ and $\nu_0$ are uniformly bounded away from $0$ and $\infty$.

\noindent (A3.) Suppose (i) and (ii) hold when $T\to \infty$ and (i) holds when $T$ is bounded. \\
(i)$\lambda_{\min} [\sum_{t=0}^{T-1} \Mean\{\bm{\xi}_{0,t}\bm{\xi}_{0,t}^{\top}-\gamma^2 \bm{u}_{\pi}(X_{0,t},A_{0,t})\bm{u}^{\top}_{\pi}(X_{0,t},A_{0,t})\}]\ge T\bar{c}$ for some constant $\bar{c}>0$, 
where $\bm{u}_{\pi}(x,a)=\Mean \{\bm{U}_{\pi}(X_{0,1})|X_{0,0}=x,A_{0,0}=a\}$ 
and $\lambda_{\min}(K)$ denotes the minimum eigenvalue of a matrix $K$. 

\noindent (ii) The Markov chain $\{X_{0,t}\}_{t\ge 0}$ is geometrically ergodic.

\smallskip




We make a few remarks. First, we do not require the limiting density function $\mu$ to be equal to the initial state density $\nu_0$. 

Second, Condition A3(i) guarantees the matrix $\Mean \widehat{\bm{\Sigma}}_{\pi}$ is invertible. In Section \ref{secaddcondA4} of the supplementary article, we show A3(i) is automatically satisfied when $\mu=\nu_0$, the target policy $\pi$ is deterministic and $b$ is the $\epsilon$-greedy policy with respect to $\pi$ that satisfies $\epsilon\le 1-\gamma^2$. 

Third, we present the detailed definition of geometric ergodicity in Appendix \ref{secsometech} to save space. Suppose the Markov chain $\{X_{0,t}\}_{t\ge 0}$ has a finite state space. Assume $\mathcal{P}_X$ is diagonalizable. Then A3(ii) holds when the second largest eigenvalue of $\mathbb{P}_X$ is strictly smaller than $1$. When $X_{0,t}$'s are generated by the vector autoregressive process $\Mean \{X_{0,t}|X_{0,t-1}\}=f(X_{0,t-1})$ for some function $f$, \cite{saikkonen2001} provided sufficient conditions that ensure the geometric ergodicity of the Markov chain. 

Finally, when $\nu_0=\mu$, $\{X_{0,t}\}_{t\ge 0}$ is stationary. Under Condition A3(ii), it follows from Theorem 3.7 of \cite{Brad2005} that $\{X_{0,t}\}_{t\ge 0}$ is exponentially $\beta$-mixing (see the proof of Lemma \ref{lemma3} for details). When $T\to \infty$, A3(ii) enables us to derive matrix concentration inequalities for $\widehat{\bm{\Sigma}}_{\pi}$. This together with A3(i) implies that $\widehat{\bm{\Sigma}}_{\pi}$ is invertible, with probability approaching $1$ (wpa1). We remark that A3(ii) is not needed when $T$ is bounded. 

For any $x\in \mathbb{X},a\in \mathcal{A}$, define
\begin{eqnarray*}
	\omega_{\pi}(x,a)=\Mean \left[\left\{Y_{0,0}+\gamma \sum_{a\in \mathcal{A}} \pi(a|X_{0,1}) Q(\pi;X_{0,1},a)-Q(\pi;X_{0,0},A_{0,0})\right\}^2 |X_{0,0}=x,A_{0,0}=a\right].
\end{eqnarray*}
 
\begin{theorem}[bidirectional asymptotics]\label{thm1}
	Assume A1-A3 hold. Suppose $L$ satisfies $L=o\{\sqrt{nT}/\log (nT)\}$, $L^{2p/d}\gg nT\{1+\|\int_{x}\Phi_L(x)\mathbb{G}(dx)\|_2^{-2}\}$, 
	and there exists some constant $c_0\ge 1$ such that $\omega_{\pi}(x,a)\ge c_0^{-1}$ for any $x\in\mathbb{X},a\in\mathcal{A}$ and $\prob(\max_{0\le t\le T-1} |Y_{0,t}|\le c_0)=1$. Then as either $n\to \infty$ or $T\to \infty$, we have
	\begin{eqnarray*}
		\sqrt{nT}\widehat{\sigma}^{-1}(\pi;\mathbb{G})\{\widehat{V}(\pi;\mathbb{G})-V(\pi;\mathbb{G})\}\stackrel{d}{\to} N(0,1).
	\end{eqnarray*}
\end{theorem}

A sketch for the proof of Theorem \ref{thm1} is given in Appendix \ref{secsketchproofthm1}. 
Under the conditions in Theorem \ref{thm1}, we can show that $\widehat{\sigma}(\pi;\mathbb{G})$ converges almost surely to some $\sigma(\pi;\mathbb{G})$. The form of $\sigma(\pi;\mathbb{G})$ is given in Section \ref{secproofthm1}. In addition, we have
\begin{eqnarray}\label{CLT}
	\frac{\widehat{V}(\pi;\mathbb{G})-V(\pi;\mathbb{G}) }{(nT)^{-1/2}\widehat{\sigma}(\pi;\mathbb{G})}=\frac{(nT)^{-1/2}}{{\sigma}(\pi;\mathbb{G})} \sum_{i=1}^n\sum_{t=1}^{\top} \left\{ \int_{x\in \mathbb{X}} \bm{U}_{\pi}(x)\mathbb{G}(dx) \right\}^{\top} \bm{\Sigma}_{\pi}^{-1} \bm{\xi}_{i,t}\varepsilon_{i,t}+o_p(1),
\end{eqnarray}
where $\bm{\Sigma}_{\pi}=\Mean \widehat{\bm{\Sigma}}_{\pi}$ and
\begin{eqnarray*}
	\varepsilon_{i,t}=Y_{i,t}+\gamma \sum_{a\in \mathcal{A}}Q(\pi;X_{i,t+1},a)\pi(a|X_{i,t+1})-Q(\pi;X_{i,t},A_{i,t}).
\end{eqnarray*}
By MA, CMIA and \eqref{QBell2}, the leading term on the RHS of \eqref{CLT} forms a mean-zero martingale (details can be found in Section \ref{secproofthm1}). As either $n$ or $T$ grows to infinity, the asymptotic normality follows from the martingale central limit theorem. 

When $\sigma(\pi;\mathbb{G})$ is bounded away from zero, it can be seen from \eqref{CLT} that $\widehat{V}(\pi;\mathbb{G})-V(\pi;\mathbb{G})=O_p(n^{-1/2} T^{-1/2})$. {\color{black}That is, the proposed value estimator converges at a rate of $(nT)^{-1/2}$. In contrast, AIPW-type estimators typically converge at a rate of $n^{-1/2}$ and are thus not suitable for settings with only a few trajectories.}



\subsection{Inference of the value under an (estimated) optimal policy}\label{secvalueestimatedpolicy}
For simplicity, we assume $T_1=T_2=\cdots=T_n=T$ throughout this section. Consider an estimated policy $\widehat{\pi}$, computed based on the data $\{(X_{i,t},A_{i,t},Y_{i,t},X_{i,t+1})\}_{0\le t< T,1\le i\le n}$. The integrated value under $\widehat{\pi}$ is given by
\begin{eqnarray*}
	V(\widehat{\pi};\mathbb{G})=\int_{x\in \mathbb{X}} V(\widehat{\pi};x)\mathbb{G}(dx). 
\end{eqnarray*}
We will require the value of $\widehat{\pi}$ to converge to some fixed policy $\pi^*$ (possibly different from $\pi^{\tiny{opt}}$), i.e,
\begin{eqnarray}\label{Vpiconverge}
V(\widehat{\pi};\mathbb{G})-V(\pi^*;\mathbb{G})\stackrel{P}{\to} 0.
\end{eqnarray}
{\color{black}In this section, we focus on constructing CIs for $V(\widehat{\pi};\mathbb{G})$ and $V(\pi^*;\mathbb{G})$.}
\subsubsection{The challenge}\label{secchallenge}
{\color{black}We begin by outlining the challenge of obtaining inference in the nonregular cases. Suppose $\pi^*\in \Pi^{\tiny{opt}}$. When the optimal policy is not unique, $\widehat{\pi}$ might not converge to a fixed policy, despite that its value converges (see \eqref{Vpiconverge}). 
To better illustrate this, suppose $\widehat{\pi}$ is computed by some Q-learning type algorithms, i.e,
\begin{eqnarray}\label{greedypolicy}
	\widehat{\pi}(a|x)=\left\{\begin{array}{ll}
		1, & \hbox{if}~~a=\sargmax_{a'\in \mathcal{A}} \widehat{Q}(x,a'),\\
		0, & \hbox{otherwise},
	\end{array}
	\right.
\end{eqnarray}
where $\widehat{Q}(\cdot,\cdot)$ denotes some consistent estimator for $Q^{\tiny{opt}}(\cdot,\cdot)$. Assume there exists a subset $\mathbb{X}_0$ of $\mathbb{X}$ with positive Lebesgue measure such that the argmax of $Q^{\tiny{opt}}(x,\cdot)$ is not uniquely defined for any $x\in \mathbb{X}_0$. Then $\widehat{\pi}(\cdot|x)$ might not converge to a fixed quantity for any $x\in \mathbb{X}_0$.}

Consider the plug-in estimator $\widehat{V}(\widehat{\pi};\mathbb{G})$ for $V(\widehat{\pi};\mathbb{G})$. Similar to \eqref{CLT}, we can show
\begin{eqnarray}\label{fakeCLT}
\frac{\widehat{V}(\widehat{\pi};\mathbb{G})-V(\widehat{\pi};\mathbb{G})}{(nT)^{-1/2}\widehat{\sigma}(\widehat{\pi};\mathbb{G})}=\frac{(nT)^{-\frac{1}{2}}}{{\sigma}(\widehat{\pi};\mathbb{G})} \sum_{i,t} \left\{ \int_{x\in \mathbb{X}} \bm{U}_{\widehat{\pi}}(x)\mathbb{G}(dx) \right\}^{\top} \bm{\Sigma}_{\widehat{\pi}}^{-1} \bm{\xi}_{i,t}\widehat{\varepsilon}_{i,t}+o_p(1),
\end{eqnarray}
where
\begin{eqnarray*}
	\widehat{\varepsilon}_{i,t}=Y_{i,t}+\gamma \sum_{a\in \mathcal{A}}Q(\widehat{\pi};X_{i,t+1},a)\widehat{\pi}(a|X_{i,t+1})-Q(\widehat{\pi};X_{i,t},A_{i,t}).
\end{eqnarray*}
When $\widehat{\pi}$ does not converge, 
$\sigma(\widehat{\pi};\mathbb{G})$, $\bm{U}_{\widehat{\pi}}(x)$, $\bm{\Sigma}_{\widehat{\pi}}$ and $\widehat{\varepsilon}_{i,t}$ will fluctuate randomly and might not stabilize. {\color{black}Since these quantities depend on the data as well, the martingale structure is violated. As a result, the leading term on the RHS of \eqref{fakeCLT} does not have a well tabulated limiting distribution. Thus, CIs based on $\widehat{V}(\widehat{\pi};\mathbb{G})$ will fail to maintain the nominal coverage probability. 

To allow for valid inference, we use a sequential value evaluation procedure to construct the CI. That is, we propose sequentially estimating the optimal policy and evaluating its value using different data subsets. This allows us to treat the estimated optimal policy as known conditional on past observations (see Equation \eqref{TrueCLT} in Appendix \ref{secsketchproofthm2}). The martingale CLT can thus be applied to obtain the limiting distribution for our estimator (see \eqref{TrueCLT2} and the related discussions). We detail our procedure in the next section.
}


\subsubsection{SAVE for the value under an (estimated) optimal policy}\label{secCI}
{\color{black}
We begin by dividing $\mathcal{I}_0=\{(i,t): 1\le i\le n,0\le t<T\}$ into $K$ non-overlapping subsets, denoted by $\mathcal{I}_1,\mathcal{I}_2,\cdots,\mathcal{I}_K$. At the $k$-th step, we use the sub-dataset
\begin{eqnarray*}
	O_k=\{(X_{i,t},A_{i,t},Y_{i,t},X_{i,t+1}):(i,t)\in \bar{\mathcal{I}}_k=\mathcal{I}_1\cup \mathcal{I}_2\cup \cdots \cup \mathcal{I}_{k}\},
\end{eqnarray*}
to compute an estimated optimal policy (denoted by $\widehat{\pi}_{\bar{\mathcal{I}}_k}$). Then we apply the proposed procedure in Section \ref{secvaluefixedpolicy} to dataset in the $(k+1)$-th block $O_{k+1}=\{(X_{i,t},A_{i,t},Y_{i,t},X_{i,t+1}):(i,t)\in \mathcal{I}_{k+1}\}$ to compute its value estimator $\widehat{V}_{\mathcal{I}_{k+1}}(\widehat{\pi}_{\bar{\mathcal{I}}_k};\mathbb{G})$ and the associated standard error $\widehat{\sigma}_{\mathcal{I}_{k+1}}(\widehat{\pi}_{\bar{\mathcal{I}}_k};\mathbb{G})$ (Details are given in Appendix \ref{secmoreCI2}). 

As commented in the introduction, the data-splitting rule cannot be arbitrary.
For any of the two tuples $(i_1,t_1)$ and $(i_2,t_2)$, define an order $(i_2,t_2)\succ (i_1,t_1)$ if either $t_2>t_1$ or $i_2>i_1$. For any $(i_2,t_2)\in \mathcal{I}_{k+1}$, we require the following:
\begin{eqnarray}\label{condhold}
	(i_2,t_2)\succ (i_1,t_1),\,\,\,\,\,\,\,\,\,\,\,\,\,\,\,\,\forall (i_1,t_1)\in \bar{\mathcal{I}}_k.
\end{eqnarray}
Then $\widehat{\pi}_{\bar{\mathcal{I}}_k}$ depends on the $i_2$-th patent's trajectory only through $\{(X_{i_2,j},A_{i_2,j},Y_{i_2,j})\}_{j<t_2}$ and $X_{i,t_2}$, $A_{i,t_2}$. Under MA and CMIA, \eqref{QBell2} still holds with $\pi=\widehat{\pi}_{\bar{\mathcal{I}}_k}$ any $(i,t)=(i_2,t_2)$. Similar to \eqref{CLT}, we can show conditional on the observations in $\bar{\mathcal{I}}_{k}$,  $\sqrt{nT/K}\widehat{V}_{\mathcal{I}_{k+1}}(\widehat{\pi}_{\bar{\mathcal{I}}_k};\mathbb{G})$ is asymptotically normal with variance consistently estimated by $\widehat{\sigma}_{\mathcal{I}_{k+1}}^2(\widehat{\pi}_{\bar{\mathcal{I}}_k};\mathbb{G})$.

Our final estimator $\widetilde{V}(\mathbb{G})$ is defined as a weighted average of these $\widehat{V}_{\mathcal{I}_{k+1}}(\widehat{\pi}_{\bar{\mathcal{I}}_k};\mathbb{G})$'s. Specifically, we set 
\begin{eqnarray*}
	\widetilde{V}(\mathbb{G})=\left\{ \sum_{k=1}^{K-1} \frac{1}{\widehat{\sigma}_{\mathcal{I}_{k+1}}(\widehat{\pi}_{\bar{\mathcal{I}}_k};\mathbb{G})} \right\}^{-1} \left\{ \sum_{k=1}^{K-1} \frac{\widehat{V}_{\mathcal{I}_{k+1}}(\widehat{\pi}_{\bar{\mathcal{I}}_k};\mathbb{G})}{\widehat{\sigma}_{\mathcal{I}_{k+1}}(\widehat{\pi}_{\bar{\mathcal{I}}_k};\mathbb{G})} \right\}.
\end{eqnarray*}
The inclusion of the inverse weight $1/\widehat{\sigma}_{\mathcal{I}_{k+1}}(\widehat{\pi}_{\bar{\mathcal{I}}_k};\mathbb{G})$ is necessary for the theoretical development of asymptotic normality of $\widetilde{V}(\mathbb{G})$ (see \eqref{TrueCLT2}). Our CI is given by 
\begin{eqnarray}\label{CI2}
[\widetilde{V}(\mathbb{G})-z_{\alpha/2} \{nT(K-1)/K\}^{-1/2} \widetilde{\sigma}(\mathbb{G}), \widetilde{V}(\mathbb{G})+z_{\alpha/2} \{nT(K-1)/K\}^{-1/2} \widetilde{\sigma}(\mathbb{G})],
\end{eqnarray}
where $\widetilde{\sigma}(\mathbb{G})=(K-1)\{ \sum_{k=1}^{K-1} \widehat{\sigma}_k^{-1}(\widehat{\pi}_{\bar{\mathcal{I}}_k};\mathbb{G}) \}^{-1}$.
}

It remains to specify $\mathcal{I}_1,\mathcal{I}_2,\cdots,\mathcal{I}_K$ that satisfy \eqref{condhold}. Consider some positive integers $n_{\min}\le n$, $T_{\min}\le T$. Assume $n$ and $T$ are divisible by $n_{\min}$ and $T_{\min}$, respectively. Let $K_n=n/n_{\min}$ and $K_T=T/T_{\min}$. We set $K=K_n K_T$. For any $1\le k_n\le K_n$, $1\le k_T\le K_T$, define a set $\mathcal{I}(k_n,k_T)$ by
\begin{eqnarray*}
		\{(i,t):(k_n-1)n_{\min}<i\le k_n n_{\min}, (k_T-1)T_{\min}\le t< k_T T_{\min} \}. 
\end{eqnarray*}
Thus, each block $\mathcal{I}(k_n,k_T)$ contains data from $n_{\min}$ trajectories with $T_{\min}$ decision time points. Below, we introduce two special examples. 

\begin{enumerate}
	\item When only a few trajectories are available, we may set $n_{\min}=n$. Then, the blocks are constructed according to the times that decisions were being made. 
	
	\item When each trajectory contains a very short time period, we may set $T_{\min}=T$. Then, the observations are divided according to the trajectories they belong to.
\end{enumerate} 

We order these blocks by
\begin{eqnarray*}
	\mathcal{I}(1,1),\mathcal{I}(2,1),\dots,\mathcal{I}(K_n,1),\mathcal{I}(1,2),\mathcal{I}(2,2),\dots,\mathcal{I}(K_n,2),\dots,\mathcal{I}(1,k_T),\dots,\mathcal{I}(K_n,K_T).
\end{eqnarray*}
Based on this order, we set $\mathcal{I}_k=\mathcal{I}(n(k),T(k))$ where $n(k)$ and $T(k)$ are the unique positive integers that satisfy $k=n(k)+(T(k)-1)K_n$. 
For any $k_2> k_1$, we have either $n(k_2)> n(k_1)$ or $T(k_2)> T(k_1)$. Thus, the proposed data-splitting rule guarantees \eqref{condhold} holds for any $k$. 

In Theorem \ref{thm2} below, we establish the validity of our CI in \eqref{CI2}. It relies on Condition A3* and A4. A3* is very similar to A3 and we present the detailed definition in Appendix \ref{secsometech} to save space. 

\smallskip

\noindent (A4) $\Mean  |V(\widehat{\pi}_{\bar{\mathcal{I}}_{k}};\mathbb{G})-V(\pi^*;\mathbb{G})|=O(|\bar{\mathcal{I}}_{k}|^{-b_0})$ for some $b_0>1/2$ such that $(nT)^{b_0-1/2}\gg \|\int_x \Phi_L(x)\mathbb{G}(dx)\|_2^{-1}$, where the big-$O$ term is uniform in $k$. 

\smallskip

Set $\mathcal{I}=\mathcal{I}_0$. By Markov's inequality, it is immediate to see A4 implies that Condition \eqref{Vpiconverge} holds. When the tensor-product B-splines are used, we have $\liminf_L \|\int_x \Phi_L(x)\mathbb{G}(dx)\|_2>0$. Thus, it is equivalent to require $\Mean  |V(\widehat{\pi}_{\bar{\mathcal{I}}_k};\mathbb{G})-V(\pi^*;\mathbb{G})|=O(|\bar{\mathcal{I}}_k|^{-b_0})$ for some $b_0>1/2$. In Section \ref{secconvergeVpi}, we discuss the rate $b_0$ in detail  when $\pi^*=\pi^{\tiny{opt}}$ and $\widehat{\pi}$ is a greedy policy derived based on some Q-learning algorithms.

\begin{theorem}[bidirectional asymptotics]\label{thm2}
	Assume A1-A2, A3* and A4 hold. Suppose $K=O(1)$ and $L$ satisfies 
	$L=o\{\sqrt{nT}/\log (nT)\}$, $L^{2p/d}\gg nT\{1+\|\int_{x}\Phi_L(x)\mathbb{G}(dx)\|_2^{-2}\}$.  Suppose $T_{\min}=T$ if $T$ is bounded. Assume there exists some constant $c_0\ge 1$ such that $\omega_{\pi}(x,a)\ge c_0^{-1}$ for any $x,a,\pi$ and $\prob(\max_{0\le t\le T-1} |Y_{0,t}|\le c_0)=1$. Then as either $n\to \infty$ or $T\to \infty$, we have
	\begin{eqnarray*}
		&&\sqrt{nT(K-1)/K}\widetilde{\sigma}^{-1}(\mathbb{G})\{\widetilde{V}(\mathbb{G})-V(\widehat{\pi};\mathbb{G})\}\stackrel{d}{\to} N(0,1),\\
		&&\sqrt{nT(K-1)/K}\widetilde{\sigma}^{-1}(\mathbb{G})\{\widetilde{V}(\mathbb{G})-V(\pi^*;\mathbb{G})\}\stackrel{d}{\to} N(0,1).
	\end{eqnarray*}
\end{theorem}
{\color{black}We provide a sketch for the proof of Theorem \ref{thm2} in Appendix \ref{secsketchproofthm2}. }

\subsubsection{Convergence of the value under an estimated optimal policy}\label{secconvergeVpi}
{\color{black}For any $\mathcal{I}\subseteq \mathcal{I}_0$, we use $\widehat{\pi}_{\mathcal{I}}$ to denote an estimated optimal policy based on observations in $\mathcal{I}$. Let $\widehat{Q}_{\mathcal{I}}(\cdot,\cdot)$ denote some consistent estimator for $Q^{\tiny{opt}}(\cdot,\cdot)$ and $\widehat{\pi}_{\mathcal{I}}$ denote the greedy policy with respect to $\widehat{Q}_{\mathcal{I}}(\cdot,\cdot)$ (see Equation \eqref{greedypolicy}).}

In the following, we focus on relating $|V(\pi^{\tiny{opt}};\mathbb{G})-V(\widehat{\pi}_{\mathcal{I}};\mathbb{G})|$ to the prediction loss $\widehat{Q}_{\mathcal{I}}-Q^{\tiny{opt}}$. By definition,
$V(\pi^{\tiny{opt}};x)\ge V(\widehat{\pi}_{\mathcal{I}};x),\forall x\in \mathbb{X}$. 
Hence, $V(\pi^{\tiny{opt}};\mathbb{G})\ge V(\widehat{\pi}_{\mathcal{I}};\mathbb{G})$. It suffices to provide an upper bound for $V(\pi^{\tiny{opt}};\mathbb{G})- V(\widehat{\pi}_{\mathcal{I}};\mathbb{G})$. 
We introduce a margin-type condition A5 below. 

\noindent (A5) Assume there exist some constants $\alpha,\delta_0>0$ such that
\begin{eqnarray}\label{lambdamargin}
\lambda\left\{x\in \mathbb{X}:\max_{a} Q^{\tiny{opt}}(x,a)-\max_{a'\in \mathcal{A}-\argmax_{a} Q^{\tiny{opt}}(x,a)} Q^{\tiny{opt}}(x,a')\le \varepsilon\right\}=O(\varepsilon^{\alpha}),\\ \label{Gmargin}
\mathbb{G}\left\{x\in \mathbb{X}:\max_{a} Q^{\tiny{opt}}(x,a)-\max_{a'\in \mathcal{A}-\argmax_{a} Q^{\tiny{opt}}(x,a)} Q^{\tiny{opt}}(x,a')\le \varepsilon\right\}=O(\varepsilon^{\alpha}),
\end{eqnarray}
where $\lambda$ denotes the Lebesgue measure, the big-$O$ terms are uniform in $0<\varepsilon\le \delta_0$, and $\max_{a'\in \mathcal{A}-\argmax_{a} Q^{\tiny{opt}}(x,a)} Q^{\tiny{opt}}(x,a')=-\infty$ if the set $\mathcal{A}-\argmax_{a} Q^{\tiny{opt}}(x,a)=\emptyset$.  

For each $x$, the quantity $\max_{a} Q^{\tiny{opt}}(x,a)-\max_{a'\in \mathcal{A}-\argmax_{a} Q^{\tiny{opt}}(x,a)} Q^{\tiny{opt}}(x,a')$ measures the difference in value between $\pi^{\tiny{opt}}$ and the policy that assigns the best suboptimal treatment(s) at the first decision point and follows $\pi^{\tiny{opt}}$ subsequently. In point treatment studies, \cite{qian2011} imposed a similar condition \citep[see Equation (3.3),][]{qian2011} to derive sharp convergence rate for the value under an estimated optimal individualized treatment regime. Here, we generalize their condition in infinite-horizon settings. A5 is also closely related to the margin condition commonly used to bound the excess misclassification error \citep{Alex2004,Alex2007}.

The margin-type condition is mild. In Appendix \ref{sec:moremargin}, we present detailed examples and show the condition holds under these examples. 
The following theorems summarize our results. 


\begin{theorem}\label{thm3}
	Assume A1, \eqref{lambdamargin} and \eqref{Gmargin} hold. Suppose the following event occurs with probability at least $1-O(|\mathcal{I}|^{-\kappa})$ for any finite $\kappa>0$, 
	\begin{eqnarray*}
		\sup_{x\in \mathbb{X},a\in \mathcal{A}} |\widehat{Q}_{\mathcal{I}}(x,a)-Q^{\tiny{opt}}(x,a)|=O(|\mathcal{I}|^{-b_*}),
	\end{eqnarray*}
	for some $b_*>0$. Then $\Mean |V(\pi^{\tiny{opt}};\mathbb{G})-V(\widehat{\pi}_{\mathcal{I}};\mathbb{G})|=O(|\mathcal{I}|^{-b_*(1+\alpha)})$. 
\end{theorem}

In Theorem \ref{thm3}, we require the estimated Q-function to satisfy certain uniform convergence rate. In Theorem \ref{thm4} below, we relax this condition by assuming that the integrated loss converges to zero at certain rate. 

\begin{theorem}\label{thm4}
	Assume A1 and A5 hold. Suppose
	\begin{eqnarray*}
		\left(\Mean \int_{x\in \mathbb{X}} \sum_{a\in \mathcal{A}} |\widehat{Q}_{\mathcal{I}}(x,a)-Q^{\tiny{opt}}(x,a)|^2dx\right)^{1/2}=O(|\mathcal{I}|^{-b_*}),
	\end{eqnarray*}
	for some $b_*>0$. Then $\Mean |V(\pi^{\tiny{opt}};\mathbb{G})-V(\widehat{\pi}_{\mathcal{I}};\mathbb{G})|=O(|\mathcal{I}|^{-b_*(2+2\alpha)/(2+\alpha)})$. 
\end{theorem}

{It can be seen from Theorems \ref{thm3} and \ref{thm4} that the integrated value converges faster compared to the Q-function. We provide a sketch for the proofs of both theorems in Appendix \ref{secsketchproofthm34}.}

\subsubsection{Applications}\label{secpolicyapplication}
In this section, we provide several examples to illustrate the convergence rate of $\widehat{Q}_{\mathcal{I}}$. The proposed methods can be applied to evaluating the values under these estimated policies. {\color{black}The algorithm in Example \ref{ExamGGQ} requires to impose a linear model assumption for the optimal Q-function. The algorithm in Example \ref{exam:FQI} allows more general nonlinear and nonparametric models for the optimal Q-function.}

\begin{example}[Greedy gradient Q-learning]\label{ExamGGQ}
The optimal Q-function satisfies
\begin{eqnarray*}
	Q^{\tiny{opt}}(x,a)=r(x,a)+\gamma \int_{x'} \max_{a'\in \mathcal{A}}Q^{\tiny{opt}}(x',a') \mathcal{P}(dx'|x,a),
\end{eqnarray*}
for any $a$ and $x$, and hence
\begin{eqnarray*}
	\Mean \left[\left. \left\{ Y_{i,t}+\gamma \max_{a'\in \mathcal{A}} Q^{\tiny{opt}}(X_{i,t+1},a')-Q^{\tiny{opt}}(X_{i,t},A_{i,t}) \right\}\right|X_{i,t},A_{i,t} \right]=0.
\end{eqnarray*}
Suppose we model $Q^{\tiny{opt}}(x,a)$ by linear sieves $\Phi_L^{\top}(x) \theta_a$. Then we can compute $\{\widehat{\theta}_{a,\mathcal{I}}\}_{a\in \mathcal{A}}$ by minimizing the following projected Bellman error:
\begin{eqnarray*}
	\argmin_{ \{\theta_a\}_{a\in \mathcal{A}}} \left(\sum_{(i,t)\in \mathcal{I}} \delta_{i,t}(\{\theta_a\}_{a\in \mathcal{A}}) \bm{\xi}_{i,t}\right)^{\top} \left(\sum_{(i,t)\in \mathcal{I}} \bm{\xi}_{i,t}\bm{\xi}_{i,t}^{\top}\right)^{-1} \left(\sum_{(i,t)\in \mathcal{I}} \delta_{i,t}(\{\theta_a\}_{a\in \mathcal{A}}) \bm{\xi}_{i,t}\right),
\end{eqnarray*}
where $\delta_{i,t}(\{\theta_a\}_{a\in \mathcal{A}})=Y_{i,t}+\gamma \max_{a'\in \mathcal{A}} \Phi_L^{\top}(X_{i,t+1}) \theta_{a'}-\Phi_L^{\top}(X_{i,t}) \theta_{A_{i,t}}$. The above loss is non-smooth and non-convex as a function of $\{\theta_a\}_{a\in \mathcal{A}}$. The estimator $\{\widehat{\theta}_{a,\mathcal{I}}\}_{a\in \mathcal{A}}$ can be computed based on the greedy gradient Q-learning algorithm. 

Assuming the optimal Q-function is correctly specified, \cite{Ertefaie2018} established the consistency and asymptotic normality of the parameter estimates under the scenario where both $L$ and $T$ are fixed. Set $\widehat{Q}_{\mathcal{I}}(x,a)=\Phi^{\top}(x) \widehat{\theta}_{a,\mathcal{I}}$. Using similar arguments in proving Theorem \ref{thm1}, we can show that with proper choice of $L$, $\sup_{x\in \mathbb{X},a\in \mathcal{A}} |\widehat{Q}_{\mathcal{I}}(x,a)-Q^{\tiny{opt}}(x,a)|$ coverages at a rate of $O(|\mathcal{I}|^{-p/(2p+d)})$ up to some logarithmic factors, with probability at least $1-O(n^{-2} T^{-2})$. The condition in Theorem \ref{thm3} thus holds for any $b_*<p/(2p+d)$. 
\end{example}

\begin{example}[Fitted $Q$-iteration]\label{exam:FQI}
	In fitted $Q$-iteration (FQI), the optimal Q-function is approximated by some nonparametric models $Q(\cdot,\cdot,\theta)$ indexed by $\theta$. The parameter $\theta$ is iteratively updated by
	\begin{eqnarray*}
	 	\widehat{\theta}_{k+1}=\argmin_{\theta} \sum_{(i,t)\in \mathcal{I}_k} \{Y_{i,t}+\gamma \max_{a}Q(X_{i,t+1},a;\widehat{\theta}_k)-Q(X_{i,t},A_{i,t};\theta) \}^2,
	\end{eqnarray*} 
	for $k=0,1,2,\dots,K-1$, where $\mathcal{I}_k$'s are some subsets of $\mathcal{I}$. When $\mathcal{I}_1=\cdots=\mathcal{I}_K=\mathcal{I}$ and $Q(\cdot,\cdot,\theta)$ is the family of neural networks, this algorithm is the neural FQI proposed by \cite{riedmiller2005}. \cite{yang2019} studied a variant of neural FQI by assuming $\mathcal{I}_k$'s are disjoint and the training samples in $\cup_{k=1}^K \mathcal{I}_k$ are independent. Using similar arguments in the proof of Theorem 4.4 in \cite{yang2019}, we can show $\Mean \int_{x\in \mathbb{X}} \sum_{a\in \mathcal{A}} |\widehat{Q}_{\mathcal{I}}(x,a)-Q^{\tiny{opt}}(x,a)|^2dx$ coverages at a rate of $O(|\mathcal{I}|^{-(2p)/(2p+d)})$ up to some logarithmic factors. The conditions in Theorem \ref{thm4} thus hold for any $b_*<p/(2p+d)$. 
\end{example}


\section{Extensions to on-policy evaluation}\label{seconpolicy}
We now extend our methodology in Section \ref{secoffpolicy} to on-policy settings. The proposed CI is similar to that presented in Section \ref{secCI} and applies to any reinforcement learning algorithms that iteratively update the estimated policy based on batches of observations. Let $\{T(k)\}_{k\ge 1}$ be a monotonically increasing sequence that diverges to infinity. 
At the $k$-th iteration, define $\bar{\mathcal{I}}_k=\{(i,t):1\le i\le n,0\le t< \sum_{j=1}^k T(j) \}$. The data observed so far can be summarized as  
$\{(X_{i,t},A_{i,t},Y_{i,t},X_{i,t+1})\}_{1\le i\le n,0\le t< \sum_{j=1}^k T(j)}$. We compute the estimated policy $\widehat{\pi}_{\bar{\mathcal{I}}_k}$ based on these data. Then we determine the behavior policy $\widehat{b}_{\bar{\mathcal{I}}_k}$ as a function of $\widehat{\pi}_{\bar{\mathcal{I}}_k}$ and generate new observations
\begin{eqnarray}\label{newobs}
\{(A_{i,t},Y_{i,t},X_{i,t+1})\}_{1\le i\le n,\sum_{j=1}^{k} T(j)\le t< \sum_{j=1}^{k+1} T(j) },
\end{eqnarray} 
according to $\widehat{b}_{\bar{\mathcal{I}}_k}$. To balance the exploration-exploitation trade-off, a common choice of $\widehat{b}_{\bar{\mathcal{I}}_k}$ is the $\epsilon$-greedy policy with respect to $\widehat{\pi}_{\bar{\mathcal{I}}_k}$. 

Let $\mathcal{I}_{k+1}=\{1\le i\le n,\sum_{j=1}^{k} T(j)\le t< \sum_{j=1}^{k+1} T(j)\}$. The new observations in \eqref{newobs} are conditionally independent of $\widehat{\pi}_{\bar{\mathcal{I}}_k}$ given those in $\bar{\mathcal{I}}_k$. So the Bellman equation in \eqref{QBell2} is valid with $\pi=\widehat{\pi}_{\bar{\mathcal{I}}_k}$ for any $(i,t)\in \bar{\mathcal{I}}_{k+1}$. We compute $\widehat{V}_{\mathcal{I}_{k+1}}(\widehat{\pi}_{\bar{\mathcal{I}}_k};\mathbb{G})$ and $\widehat{\sigma}_{\mathcal{I}_{k+1}}(\widehat{\pi}_{\bar{\mathcal{I}}_k};\mathbb{G})$ as in Appendix \ref{secmoreCI2} of the supplementary article, where the number of basis $L(k+1)$ depends on both $n$ and $T(k+1)$. We iterate this procedure for $k=1,2,\dots,K-1$. The estimated value and CI for $V(\widehat{\pi}_{\bar{\mathcal{I}}_k};\mathbb{G})$ are given by
\begin{eqnarray*}
	\widetilde{V}(\mathbb{G})=\left\{ \sum_{k=1}^{K-1} \frac{\sqrt{T(k)}}{\widehat{\sigma}_{\mathcal{I}_{k+1}}(\widehat{\pi}_{\bar{\mathcal{I}}_k};\mathbb{G})} \right\}^{-1} \left\{ \sum_{k=1}^{K-1} \frac{\sqrt{T(k)}\widehat{V}_{\mathcal{I}_{k+1}}(\widehat{\pi}_{\bar{\mathcal{I}}_k};\mathbb{G})}{\widehat{\sigma}_{\mathcal{I}_{k+1}}(\widehat{\pi}_{\bar{\mathcal{I}}_k};\mathbb{G})} \right\},
\end{eqnarray*}
and $$\left[\widetilde{V}(\mathbb{G})-z_{\alpha/2} \left\{\sum_{k=2}^K \sqrt{\frac{nT(k) }{K-1}} \right\}^{-1/2} \widetilde{\sigma}(\mathbb{G}), \widetilde{V}(\mathbb{G})+z_{\alpha/2} \left\{\sum_{k=2}^K \sqrt{\frac{nT(k)}{K-1}} \right\}^{-1/2} \widetilde{\sigma}(\mathbb{G})\right],$$ 
where $\widetilde{\sigma}(\mathbb{G})=\{ \sum_{k=2}^K \sqrt{T(k)} \}\{ \sum_{k=2}^K \sqrt{T(k)}\widehat{\sigma}_k^{-1}(\widehat{\pi}_{\bar{\mathcal{I}}_{k-1}};\mathbb{G}) \}^{-1}$.
Similar to Theorem \ref{thm2}, we can show such a CI achieves nominal coverage under certain conditions. To save space, we provide our technical results in Section \ref{secmoreonpolicy} of the supplementary article. 

\section{Simulations}\label{secsimu}

In this section, we conduct Monte Carlo simulations to examine the finite sample performance of the proposed CI. We consider off-policy settings in Sections \ref{Sim_1} and \ref{Sim_2}, where CIs for values under both fixed and optimal policies are reported. In Section \ref{Sim_3}, we report CIs computed in on-policy settings. The state vector $X_{0,t}$ in our settings might not have bounded supports. For $j=1,\dots,d$, we define $X_{0,t}^{(j)*}=\Phi(X_{0,t}^{(j)})$ where $X_{0,t}^{(j)}$ stands for the $j$-th element of $X_{0,t}$ and $\Phi(\cdot)$ is the cumulative distribution function of a standard normal random variable. This gives us a transformed state vector with bounded support. The basis functions are constructed from the tensor product of $d$ one-dimensional cubic B-spline sets where knots are placed at equally spaced sample quantiles of the transformed state variables. For discrete state space $\mathbb{X}=\{x_1,\cdots,x_M\}$, we set $L=M$ and $\Phi_M(\cdot)=\{\mathbb{I}(\cdot=x_1),\cdots,\mathbb{I}(\cdot=x_M)\}^\top$. 
We set the discount factor $\gamma=0.5$ in all settings, and set $L=\lfloor (nT)^{\eta} \rfloor$ with $\eta = 3/7$. Here, for any $z\in \mathbb{R}$, $\lfloor z \rfloor$ denotes the largest integer that is smaller or equal to $z$. We tried several other values of the parameter $\eta$, and the resulting CIs are very similar and not sensitive to the choice of $\eta$. We also tried several other values of $\gamma$. Overall, the proposed CI achieves nominal coverage and performs better than other baseline methods. More details can be found in Appendix \ref{secsen} of the supplementary article. 

\subsection{Off-policy evaluation with a fixed target policy}\label{Sim_1}
\begin{figure}[!t]
	\centering
	\includegraphics[width=9cm]{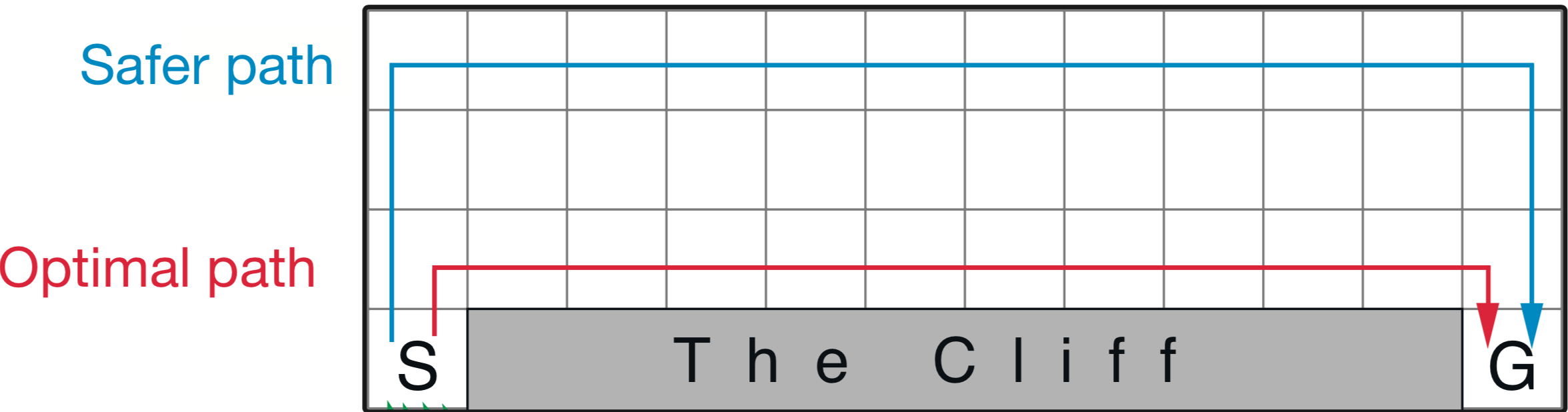}
	\vspace{-0.2cm}
	\caption{Illustration of Cliff Walking}\label{fig:cliff}
\end{figure}
{\color{black}We consider three scenarios. In Scenarios (A) and (B), the system dynamics are given by
\begin{eqnarray}\label{transitiond}
\begin{split}
X_{0,t + 1}&= \begin{bmatrix} 
\frac{3}{4}(2A_{0,t} - 1) & 0 \\
0 & \frac{3}{4}(1- 2A_{0,t}) 
\end{bmatrix} X_{0,t} + z_t,\\
Y_{0,t}&=X^{\top}_{0,t + 1} \begin{bmatrix} 2 \\ 1 \end{bmatrix}- \frac{1}{4}(2A_{0,t} - 1),
\end{split}
\end{eqnarray}
for $t\ge 0$, where $\{z_t\}_{t\ge 0}\stackrel{iid}{\sim}N(0_2, I_2/4)$ and $X_{0,0}\sim N(0_2,I_2)$. In Scenario (A), we consider a completely randomized study and set $\{A_{0,t}\}_{t\ge 0}$ to i.i.d Bernoulli random variables with expectation $0.5$. In Scenario (B), we allow the treatment assignment mechanism to depend on the observed state. Specifically, we set $m=2$ and $\prob(A_{0,t}=1|X_{0,t})=0.5  \textrm{sigmoid}(X_{0,t,1})+0.5 \textrm{sigmoid}(X_{0,t,2})$ where $X_{0,t,i}$ denotes the $i$th element in $X_{0,t}$.  
%
The target policy we consider is designed as follows, 
$$ \pi(a|x) = \left\{
\begin{array}{rcl}
0,      &      & {x_1 > 0 \ \hbox{and} \ x_2 > 0};\\
1,     &      & \hbox{otherwise},
\end{array} \right. $$
where $x_i$ denotes the $i$th element of $x$. The reference distribution $\mathbb{G}$ is set to $N(0_2,I_2)$.


In Scenario (C), we consider a standard RL setting included in OpenAI Gym \citep{brockman2016openai}: Cliff Walking. This RL example is detailed in Example 6.6 in \citet{Sutton2018}. The objective is to identify the optimal path from the starting point S to the destination point G without falling off the cliff (see Figure \ref{fig:cliff}). This scenario corresponds to an episodic task where the agent will be sent instantly to the starting point wherever it steps into the cliff or arrives at the destination. We manually add some noises to the immediate rewards simulated by the OpenAI Gym to ensure that the system dynamics are not deterministic. We remark that this task is considered in \citet{kallus2020double} as well. The target policy is the optimal policy and the behavior policy is a 50-50 mixture of the optimal and uniform random policies.

The true value function $V(\pi,\mathbb{G})$ is computed by Monte Carlo approximations. Specifically, we simulate $N=10^5$ independent trajectories with initial state variable distributed according to $\mathbb{G}$. 
The action at each decision point is chosen according to $\pi$. Then we approximate $V(\pi,\mathbb{G})$ by $\sum_{j = 1}^{N} \sum_{t = 0}^{T_i - 1} \gamma^{t} Y_{j,t}/N$ where $T_i$ is set to 500 in Scenarios (A), (B) and the termination time of each episode in Scenario (C). 
The integrals in \eqref{hatVpiG} and \eqref{hatsigmapiG} are computed via Monte Carlo methods. For Scenarios (A) and (B), we further consider 9 cases by setting $n=25,50,100$ and $T=30,50,70$. For Scenario (C), we consider 3 cases by setting $n=500,1000,1500$. Each trajectory have 13 time points on average, under the behavior policy. 

\begin{figure}[t]
	\caption{Empirical coverage probabilities (ECP) and average lengths (AL) of CIs constructed by the proposed method (colored in blue) and the DRL method (colored in red) as well as the mean-squared errors (MSE) of the corresponding value estimators, with different choices of $n$ and $T$. Settings correspond to Scenario (A) and Scenario (B), from top plots to bottom plots.}\label{fig1}
	\centering
	\includegraphics[width=13cm]{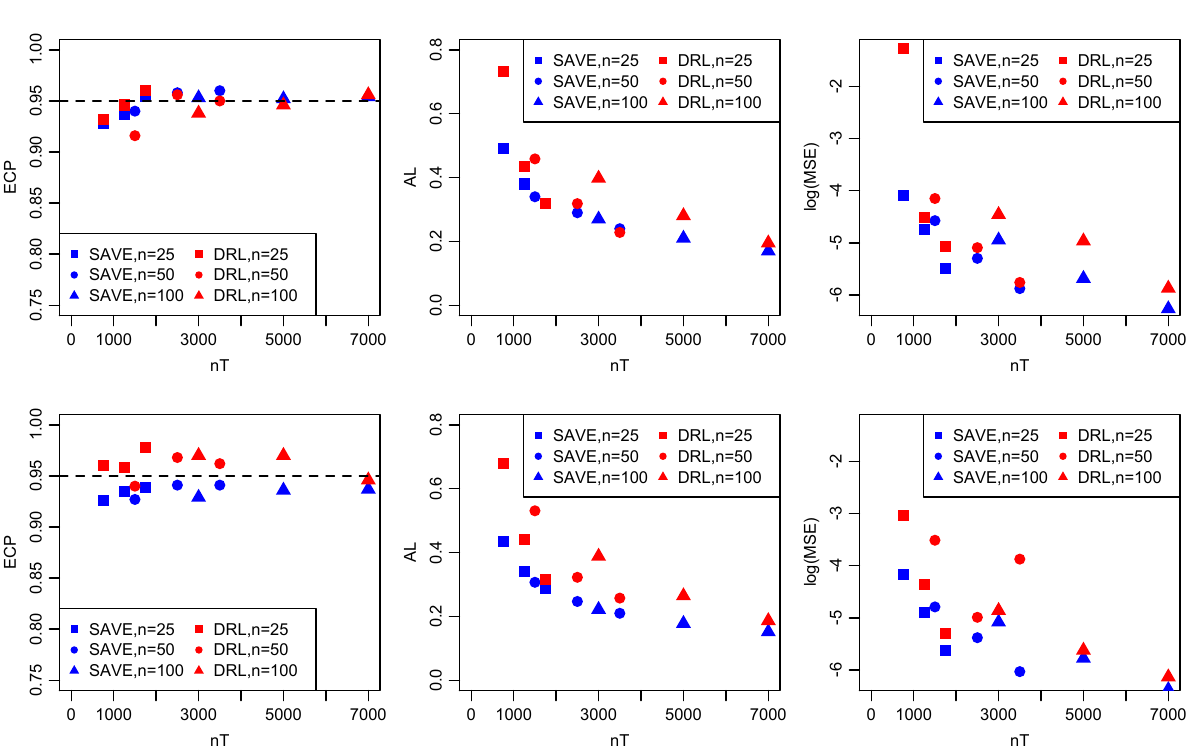}
\end{figure}

\begin{table}[b]
	\caption{Empirical coverage probabilities (ECP) and average lengths (AL) of CIs constructed by the proposed method and DRL as well as the mean-squared errors (MSE) of the corresponding value estimators under Scenario (C).}\label{tab-1}
	\centering
	\begin{tabular}{|c|c|c|c|c|c|c|}\hline
		 & \multicolumn{3}{c|}{SAVE} & \multicolumn{3}{c}{DRL}\\ \hline
		n & ECP & AL & log(MSE) & ECP & AL & log(MSE)\\ \hline 
		500 & 0.96 & 0.11 & -8.11 & 0.89 & 0.13 & -6.50 \\ \hline
		1000 & 0.93 & 0.08 & -7.60 & 0.85 & 0.09 & -7.26\\ \hline
		1500 & 0.96 & 0.06 & -8.11 & 0.87 & 0.07 & -7.82\\\hline 
	\end{tabular}
\end{table}


The DRL estimator has been shown to be much more efficient than AIPW or IPW estimators \citep{thomas2015,jiang2016}. So we focus on comparing our approach with DRL. 
DRL requires the calculation of the Q-function, the marginalized density ratio and the behavior policy. 
Here, we treat the behavior policy as known and estimate the Q-function and the density ratio based on nonparametric sieve regression. 

In Figure \ref{fig1} and Table \ref{tab-1}, we report the empirical coverage probabilities (ECPs) and average lengths (ALs) of CIs constructed by the proposed method, with different choices of $n$ and $T$. It can be seen that our CI achieves nominal coverage in all cases. Its length  decreases as $nT$ increases. This is consistent with our theoretical findings where we show the proposed value estimator converges at a rate of $n^{-1/2}T^{-1/2}$ under certain conditions (see the discussions below Theorem \ref{thm1}).

Comparing our method with DRL, it is clear that our CIs are in general narrower than those constructed by DRL. In addition, MSEs of the proposed value estimates are smaller than those based on DRL. This is consistent with our theoretical analysis in Appendix \ref{sec:varcompare} where we show the variance of our value estimator is strictly smaller than that based on DRL under certain conditions. In addition, it can be seen from Table \ref{tab-1} that ECPs of DRL are below 90\% in Scenario (C). 

In Appendix \ref{sec:addsimu}, we conduct some additional simulation studies under Scenario (A) by setting the reference distribution $\mathbb{G}$ to a Dirac measure. The proposed CI achieves nominal coverage under these settings as well. 

\subsection{Off-policy evaluation with an (estimated) optimal policy}\label{Sim_2}
In this section, we focus on constructing the CI for value under an optimal policy. Specifically, we use a version of fitted Q-iteration (double FQI) to compute the estimated optimal policy. Detailed algorithm can be found in Section 
\ref{secaddFQI} of the supplementary article. 
To implement the proposed CI in Section 
\ref{secvalueestimatedpolicy}, we set $K_n=2$, $K_T=2$ in Scenarios (A), (B) and $K_n=3$, $K_T=1$ in Scenario (C). 
To evaluate our CI, we generate a very large dataset to compute an estimated optimal policy $\widehat{\pi}^*$ based on double FQI and use the Monte Carlo methods described in Section \ref{Sim_1} to evaluate its value $V(\widehat{\pi}^*;\mathbb{G})$. Then we treat $V(\widehat{\pi}^*;\mathbb{G})$ as the true optimal value $V(\pi^*;\mathbb{G})$. 
We consider the same three scenarios detailed in Section \ref{Sim_1}. 
For Scenarios (A) and (B), we fix $\gamma=0.5$ and consider 6 cases by setting $n=100,200$ and $T=60,100,140$. For Scenario (C), we consider 4 cases by setting $n=3000, 4500$ and $\gamma=0.5,0.7$. 
The ECP, AL and MSE of our CI are reported in the top two panels of Figure \ref{fig3} in Table \ref{tab-2}. It can be seen that these ECPs are close to the nominal level in most cases. ALs and MSEs decay as either $n$ or $T$ increases. 

\begin{figure}[!t]
	\caption{Empirical coverage probabilities and average lengths of CIs constructed by the proposed method as well as the mean squared errors of the value estimates, with different choices of $n$ and $T$. Settings correspond to Scenarios (A), (B) and (D), from top panels to bottom panels.}\label{fig3}
	\centering
	\includegraphics[width=15cm]{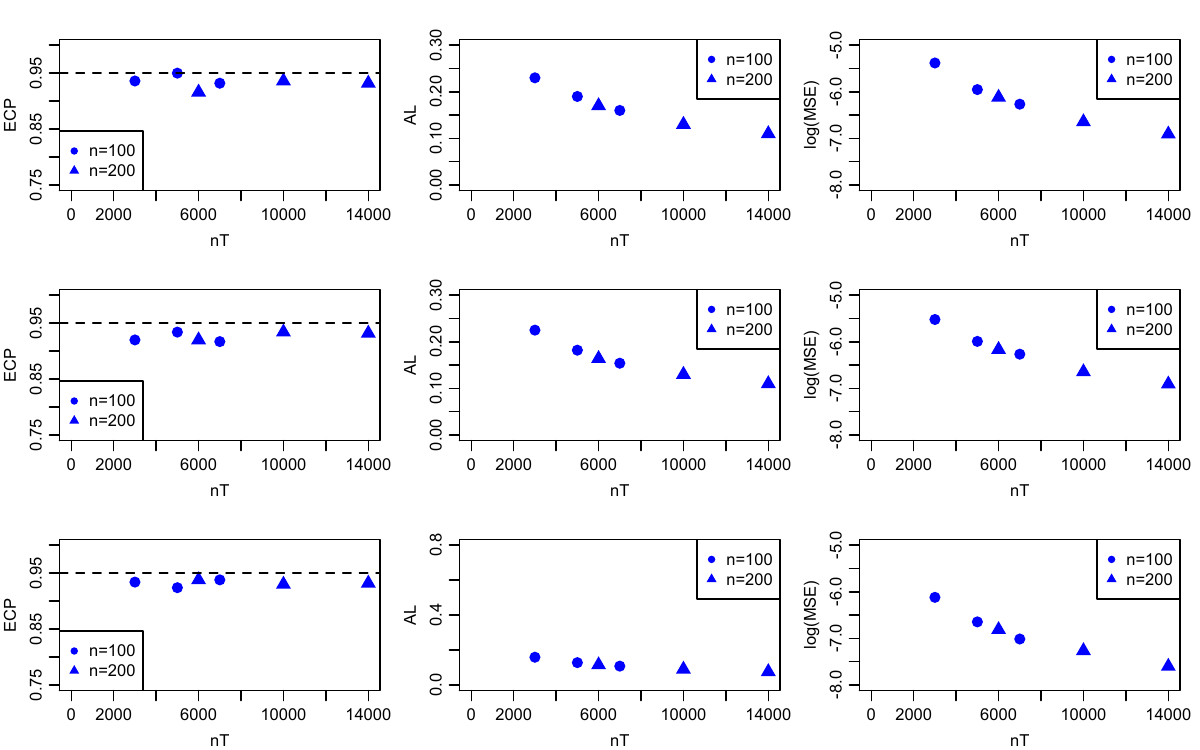}
\end{figure}

\begin{table}[b]
	\caption{Empirical coverage probabilities (ECP) and average lengths (AL) of CIs constructed by the proposed method as well as the mean-squared errors (MSE) of the corresponding value estimators under Scenario (C), with different combinations of $n$ and $\gamma$.}\label{tab-2}
	\centering
	\begin{tabular}{|c|c|c|c|c|c|c|}\hline
		& \multicolumn{3}{c|}{$n=3000$} & \multicolumn{3}{c}{$n=4500$}\\ \hline
		$\gamma$ & ECP & AL & log(MSE) & ECP & AL & log(MSE)\\ \hline 
		0.5 & 0.94 & 0.12 & -6.91 & 0.94 & 0.10 & -7.51\\ \hline
		0.7 & 0.95 & 0.23 & -5.81 & 0.96 & 0.19 & -6.35\\\hline 
	\end{tabular}
\end{table}

%
%

In addition, we design a non-regular setting Scenario (D) where the actions do not have effects on the transition dynamics or the immediate rewards. Specifically, for any $t\ge 0$, we set
$$X_{0,t + 1} = \begin{bmatrix} 
-\frac{3}{4} & 0 \\
0 & \frac{3}{4} 
\end{bmatrix} X_{0,t} + z_t,$$ and
$$Y_{0,t} = X^{\top}_{0,t + 1} \begin{bmatrix} 2 \\ 1 \end{bmatrix},$$
where $\{z_t\}_{t\ge 0}\stackrel{iid}{\sim}N(0_2, I_2/4)$ and $\{A_{0,t}\}_{t\ge 0}$ are i.i.d Bernoulli random variables with expectation $0.5$. Under this setup, any policy will achieve the same value function. As a result, the optimal policy is not unique. We consider the same reference distribution $\mathbb{G}$, and the same combinations of $n$ and $T$ as in the regular setting. ECPs and ALs of the proposed CIs are plotted in the bottom panels of Figure \ref{fig3}. It can be seen that our CIs achieve nominal coverage in the non-regular setting as well. 

In Appendix \ref{sec:addsimu}, we conduct some additional simulation studies under Scenario (A) by setting the reference distribution $\mathbb{G}$ to a Dirac measure. Findings are very similar to those in cases where $\mathbb{G}=N(0_2,I_2)$.}

\subsection{On-policy evaluation with an (estimated) optimal policy}\label{Sim_3}
We consider a setting where the transition dynamics and immediate rewards are defined by \eqref{transitiond}. In the first block of data $\{(X_{i,t},A_{i,t},Y_{i,t},X_{i,t+1})\}_{0\le t< T(1), 1\le i\le n}$, the actions are generated according to i.i.d Bernoulli random variables with expectation 0.5. For $k=2,\cdots,K$, we use double FQI to estimate the optimal policy based on the data observed so far $\{(X_{i,t},A_{i,t},Y_{i,t},X_{i,t+1})\}_{0\le t< T(k-1), 1\le i\le n}$ and use an $\epsilon$-greedy method to generate actions in the next block of data. In our experiments, we set $\epsilon=0.2$, $K=4$ and $T(1)=T(2)=\cdots=T(K)=T$. We fix $n=25$ and consider three choices of $T$, corresponding to $T=120$, $200$ and $280$. We consider three choices of $\mathbb{G}$, corresponding to $\mathbb{G}=\delta_{(0.5,0.5)}$, $\delta_{(-0.5,-0.5)}$ and  $N(0_2,I_2)$. The true optimal value function is approximated by Monte Carlo methods, as in off-policy settings. ECPs and ALs of the proposed CIs are reported in Table \ref{tab0}.  It can be seen that ECPs are close to the nominal level in almost all cases and ALs decrease as $T$ increases.

\begin{table}[!t]
	\caption{Empirical coverage probabilities and average lengths of CIs constructed in on-policy settings with different choices of $T$ and $\mathbb{G}$}\label{tab0}
		\centering
		\begin{tabular}{ |c|c|c|c|c|c|c|} 
			\hline
			&\multicolumn{3}{|c|}{ 
				ECPs
			} & \multicolumn{3}{|c|}{ 
			ALs}
			\\
			\hline
			& T = 120 & 200 & 280 & T = 120 & 200 & 280\\ 
			\hline
			$\mathbb{G}=\delta_{(0.5,0.5)}$ & 0.947  & 0.914  & 0.931  & 0.51  &    0.46 & 0.38	\\ \hline
			$\mathbb{G}=\delta_{-(0.5,0.5)}$ & 0.925 & 0.942 & 0.940  & 0.72 & 0.59 & 0.48	\\ \hline
			$\mathbb{G}=N(0_2,I_2)$ & 0.914 & 0.926 & 0.948 & 0.29  &    0.21 & 0.17 
			\\	
			\hline
	\end{tabular}
\end{table}

%

\section{Application to the OhioT1DM dataset}\label{secrealdata}
As commented in the introduction, this dataset contains eight weeks' records of CGM blood glucose levels, insulin doses and self-reported life-event data for each of six patients with type 1 diabetes. To analyze this data, we divide these eight weeks into three hour intervals. The state variable $X_{i,t}$ is set to be a three-dimensional vector. Specifically, its first element $X_{i,t}^{(1)}$ is the average CGM blood glucose levels during the three hour interval $[t-1,t)$. The second covariate $X_{i,t}^{(2)}$ is constructed based on the $i$-patient's self-reported time and the carbohydrate estimate for the meal. Suppose the patient has meals at time $t_1,t_2,\dots,t_N\in [t-1,t)$ with the carbohydrate estimates $\hbox{CE}_1,\hbox{CE}_2,\dots,\hbox{CE}_N$. Define
\begin{eqnarray*}
	X_{i,t}^{(2)}=\sum_{j=1}^N  \hbox{CE}_j \gamma_c^ {36(t_j-t+1)},
\end{eqnarray*}
where $\gamma_c$ corresponds to the decay rate every five minutes. Here, we set $\gamma_c=0.5$. The third covariate $X_{i,t}^{(3)}$ is defined as an average of the basal rate during the three hour interval. 

We discretize the action according to the amount of insulin injected in the three hour interval. Specifically, $A_{i,t}=1$ when the total amount of insulin delivered to the $i$-th patient is greater than one unit. Otherwise, we set $A_{i,t}=0$. The immediate reward $Y_{i,t}$ is defined according to the Index of Glycemic Control \citep[IGC,][]{rodbard2009}, which is a non-linear function of the blood glucose levels. Specifically, we set
$$ Y_{i,t} =\left\{
\begin{array}{rcl}
&-\frac{1}{30}(80 - X_{i,t+1}^{(1)})^2,            & { X_{i,t+1}^{(1)}     <     80;}\\
&0,          & {80 \le X_{i,t+1}^{(1)}     <     140;}\\
&-\frac{1}{30}(X_{i,t+1}^{(1)} - 140)^{1.35},    & {    140   \le   X_{i,t+1}^{(1)}. }
\end{array} \right. $$ 
A large IGC indicates the patient is in good health status. We set the discount factor $\gamma=0.5$, as in simulations. 
\begin{figure}[!t]
	\caption{Confidence interval of the value difference between an estimated optimal policy and the behavior policy for each of the six patients, with $\gamma=0.5$.}\label{figreal}
	\centering
	\includegraphics[width=12cm]{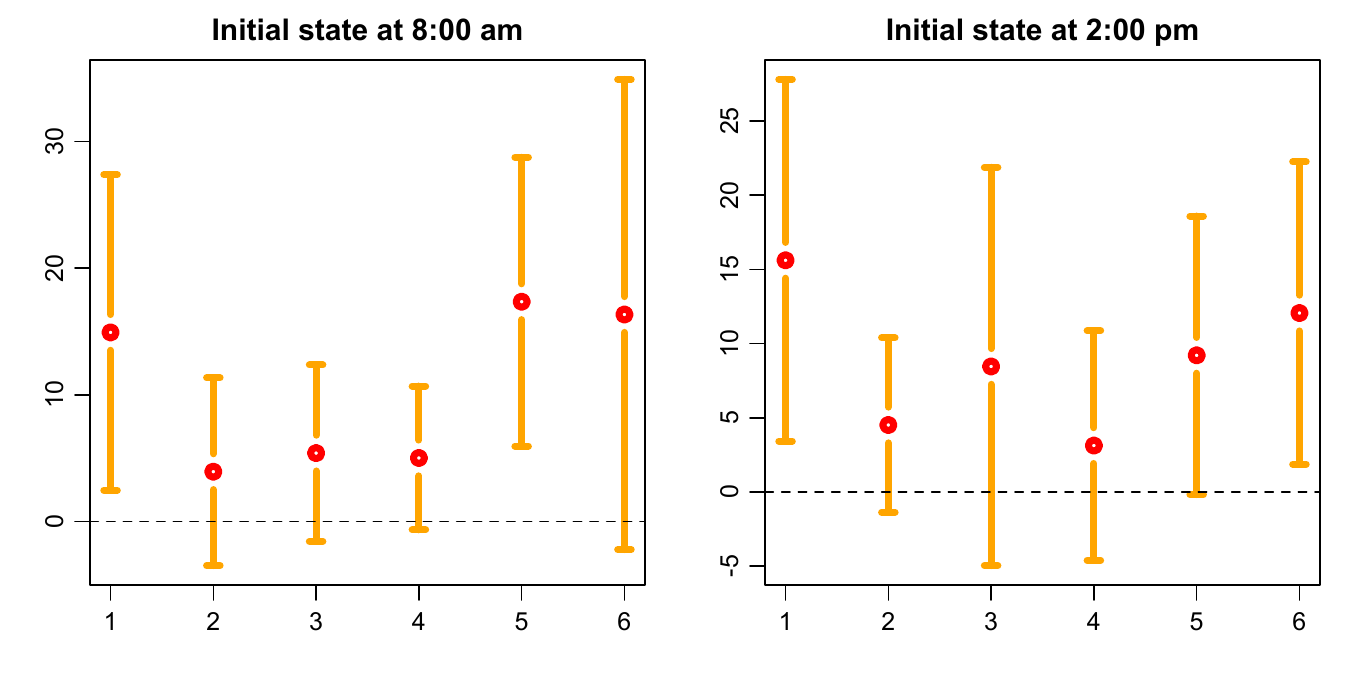}
\end{figure}

\begin{figure}[!t]
	\caption{Confidence intervals of the value difference between an estimated optimal policy and the behavior policy for each of the six patients, with different $\gamma$.}\label{figreal2}
	\centering
	\includegraphics[width=5.5cm]{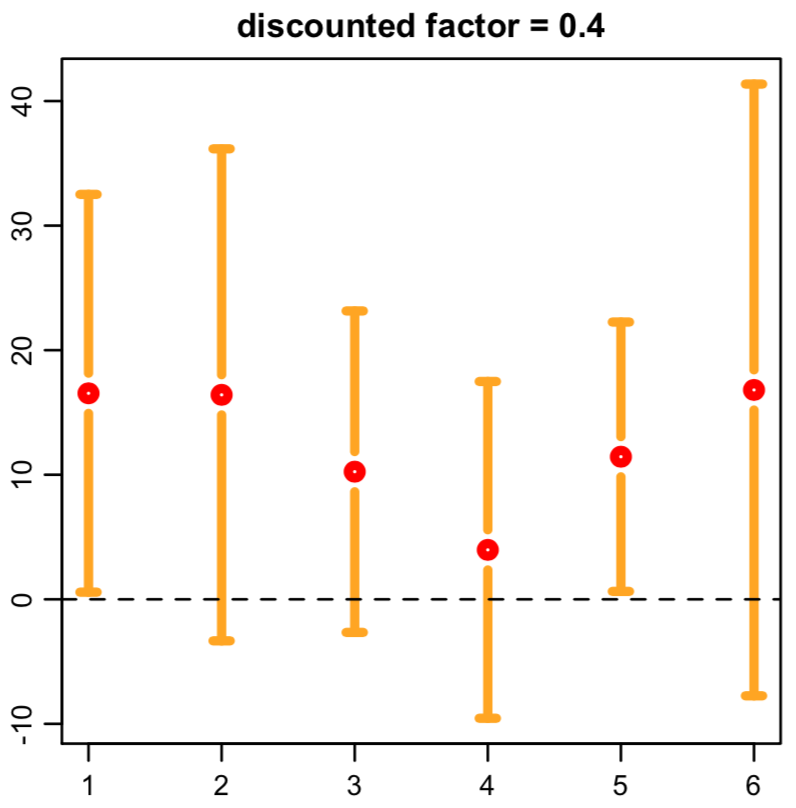}
	\includegraphics[width=5.5cm]{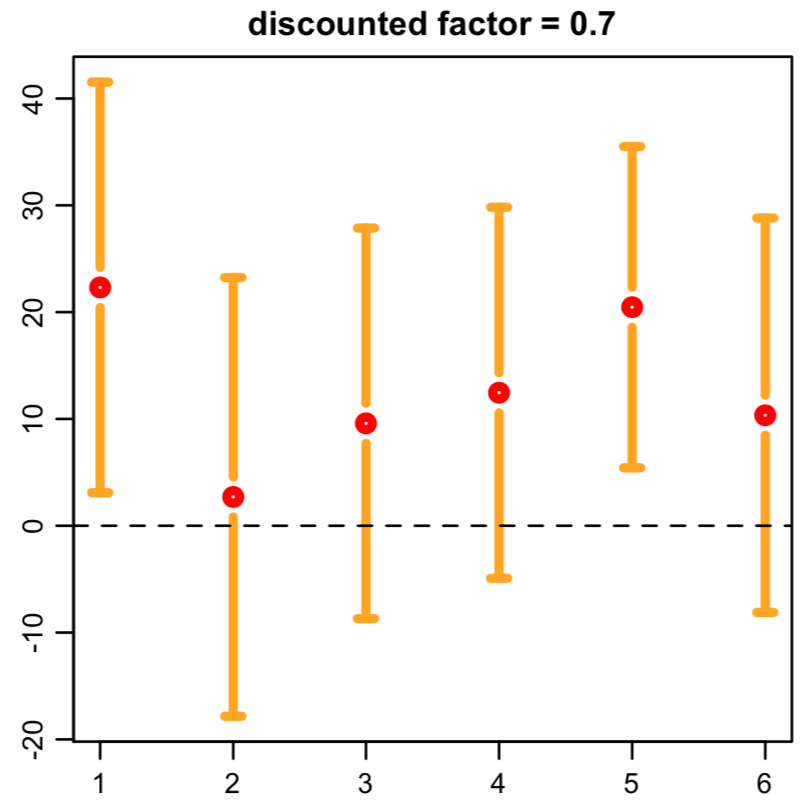}
\end{figure}

For the $i$-th patient, we apply the double FQI algorithm 
to the data $$\{(X_{i,t},A_{i,t},Y_{i,t},X_{i,t+1})\}_{0\le t< T},$$ to estimate a patient-specific optimal policy. 
{\color{black}Then we  compute the estimator for the value function $V(\pi^{\tiny{opt}};X_{i,0})$ starting from the initial state variable $X_{i,0}$. In addition, we extend our methodology in Section \ref{Sim_2} to construct the confidence interval for the value difference $V(\pi^{\tiny{opt}};X_{i,0})-V(b;X_{i,0})$ where $V(b;X_{i,0})$ corresponds to the value under the behavior policy. See Appendix \ref{app:vd} for details. In Figure \ref{figreal}, we plot our proposed CI for the value difference, for each of the six patients, when the initial starting time is either 8:00 am or 2:00 pm in Day 1. It can be seen that the estimated value differences are strictly positive in all cases. This implies that the optimal value is strictly larger than the observed discounted cumulative reward. In some cases, the lower bound of our CI is larger than zero. The difference is thus significant. In Figure \ref{figreal2}, we fix the starting time to 8:00 am and plot the CI of the value difference with different $\gamma$. Results show a similar qualitative pattern. This suggests applying reinforcement learning algorithms could potentially improve some patients' health status.
\section{Discussion}
\subsection{Comparison between DRL}\label{sec:compareDRL}
We discuss the advantages and limitations among the proposed method and DRL when inferring the value under a fixed decision rule. Generally speaking, the proposed method results in narrower CIs and would be preferred in cases where $(nT)^{-1/2}$-consistent estimation of the Q-function is feasible. This includes settings where the dimension of the state-vector is not large, as in our real data applications. In contrast, DRL would be preferred in ergodic environments with high-dimensional covariates where  $(nT)^{-1/2}$-consistent estimation of the Q-function is infeasible. 

Specifically, in Appendix \ref{sec:varcompare}, we consider settings where both the behavior policy and the target policy are nondynamic. Under certain conditions, we prove that the variance of the DRL estimator is 
strictly larger than that of the proposed estimator. 
This in turn implies that our method yields a narrower CI in general. 

In addition, we remark that the CI constructed by DRL requires the data to be generated from an ergodic environment. In the Cliff Walking example, the data are generated by a mixture of the optimal and random policy. Since the agent will be sent instantly to the starting point wherever it steps into the cliff or arrives at the destination, the Markov chain formed by the state-action pair is no longer ergodic. It can be seen from Table \ref{tab-1} where ECP of the CI is well below the nominal level in the Cliff Walking example. Although our procedure also requires the ergodicity assumption (see Condition (A3)(ii)), this assumption is not necessary. It can be seen from the proof of Theorem \ref{thm1} that our CI is valid as long as the random matrix $\widehat{\bm{\Sigma}}_{\pi}$ stabilizes. This is consistent with the findings in Table \ref{tab-1} where our CI achieves nominal coverage in all cases. 

However, to ensure the proposed CI is valid, we require the bias of our Q-estimator to decay at a rate of $o((nT)^{-1/2})$. Consequently, our estimator converges at a rate of $(nT)^{-1/2}$. This rate might not be achievable in high-dimensions. In contrast, DRL requires a weaker condition. The CI based on DRL is valid when both the Q-estimator and the estimated marginalized density ratio converge at a rate of $o((nT)^{-1/4})$. 

Another potential limitation of our method is that in cases where $\widehat{\bm{\Sigma}}_{\pi}$ is close to a singular matrix, the resulting Q-estimator might suffer from over-fitting, leading to an unbounded outcome. In practice, we could add a ridge penalty to reduce over-fitting. We discuss in detail in Appendix \ref{secaddrealdata}. 

\subsection{Number of basis functions}\label{sec:disnumber}
We outline a procedure to choose the number of basis function $L$ in this section. The idea is to simulate the model dynamics and select $L$ such that the resulting confidence interval achieves nominal coverage under the simulated model. Specifically, given the observed data $\{(X_{i,t},A_{i,t},Y_{i,t},X_{i,t+1})\}_{0\le t<T_i,1\le i\le n}$, we propose to learn the conditional density function of $(Y_{i,t},X_{i,t+1})$ given $(A_{i,t},X_{i,t})$. Following \citet{janner2019trust}, we recommend to use a Gaussian distribution to model the conditional density function in practice, 
\begin{eqnarray}\label{eqn:gauden}
(Y_{i,t},X_{i,t+1})|(A_{i,t},X_{i,t})=(a,x)\sim N(\mu(a,x),\Sigma(a,x)).
\end{eqnarray}
The conditional mean $\mu$ can be  estimated via nonparametric regression (e.g., random forest). Let $\widehat{\mu}$ denote the corresponding estimator. Given the set of estimated residuals $\{(Y_{i,t},X_{i,t+1})^\top-\widehat{\mu}(A_{i,t},X_{i,t})\}_{i,t}$ the conditional covariance function $\Sigma$ can be estimated via regression as well. We remark that in addition to the Gaussian function, other density functions could be used to model the system dynamics as well. 

The behavior policy $b$ can be similarly estimated via regression. Given an estimated behavior policy $\widehat{b}$ and $\widehat{\mu}$, $\widehat{\Sigma}$, we generate simulated trajectories to investigate the performance of the proposed CI with different choices of $L$. 

Finally, we choose $L$ such that the resulting CI is the shortest among all CIs whose coverage probabilities are above certain level (e.g., 93\%) under the simulated environment. 

In Appendix \ref{sec}, we investigate the finite sample performance of such a method and find that it performs reasonably well. We remark that alternative to the aforementioned method, cross-validation could be applied to select $L$.
\subsection{Sensitivity to the ordering of trajectories}
The proposed sequential value evaluation procedure in Section \ref{secvalueestimatedpolicy} divides the data into blocks defined both by trajectories and by time. While there is a natural order in time, there does not appear to be a natural order in the trajectories. In Appendix \ref{sec:senordering}, we conduct additional simulation studies to investigate the sensitivity of our CI to the ordering of trajectories under Scenario (B). Results suggest that our CI is not overly sensitive under our simulation setting. 

As suggested by one of the referees, we may aggregate CIs over multiple orderings in cases where the results depend strongly on the ordering of the trajectories. \cite{dezeure2015high} derived a CI for the regression coefficients in high-dimensional models by aggregating results over multiple sample splits using a quantile function. We can adopt their method to aggregate our CIs over multiple orderings. Alternatively, one may average the value estimates over sufficiently many orderings and apply similar methods developed in \cite{wang2020debiased,shi2020breaking} to derive the CI. However, these algorithms are much more time-consuming.

\subsection{More on value-based method}
In Section \ref{sec:vorQ}, we discuss a potential drawback of using nonparametric methods to directly model the value function. We remark that a kernel-type importance sampling estimator for the $V(\pi;x)$ will not suffer from this issue, since it does not directly model the value function, but uses an inverse propensity-score weighted method instead. Both IPW and regression type estimators have their own merits. In general, IPWEs might suffer from a large variance whereas regression-based estimators might suffer from a large bias. There exist methods that combine both for more robust off-policy evaluation \citep[see e.g.,][]{kallus2019efficiently,uehara2020minimax,tang2020doubly,shi2021deep}. However, as commented in Section \ref{sec:compareDRL}, they might yield larger CIs compared to our method. 

\subsection{Rate of convergence of Q-learning type algorithms}
Through authors' communication, we found a recent independent work by \citet{hu2021fast} that derived a nonasymptotic error bound on the value of the estimated optimal policy computed by Q-learning type algorithms under the margin condition. Their results are consistent with our theoretical findings in Theorems \ref{thm3} and \ref{thm4} that show the value of the estimated optimal policy converges to the optimal value at a faster rate than the estimated Q-function.

\bibliographystyle{rss}
\bibliography{mycite}

\appendix

\section{Some technical conditions}\label{secsometech}
\subsection{More on Conditions A3}
Define $\mathcal{P}_X^{t}$ as the $t$-step transition kernel, i.e, $\mathcal{P}_X^{t}(\mathcal{B}|x)=\hbox{Pr}(X_{0,t}\in \mathcal{B}|X_{0,0}=x)$.  
Geometric ergodicity implies that there exists some function $M(\cdot)$ on $\mathbb{X}$ and some constant $\rho<1$ such that $\int_{x\in \mathbb{X}} M(x) \mu(x)dx<+\infty$ and 
\begin{eqnarray*}
	\|\mathcal{P}^t_X(\cdot|x)-\mu(\cdot)\|_{TV}\le M(x) \rho^t,\,\,\,\,\,\,\,\,\forall t\ge 0,
\end{eqnarray*} 
where $\|\cdot\|_{TV}$ denotes the total variation norm. 




\subsection{Conditions A3*}\label{seccondA4A5}
We present the technical condition (A3*) below.  
We assume the estimated policy satisfies $\widehat{\pi}_{\mathcal{I}}\in \Pi$ with probability $1$, for any $\mathcal{I}$. For example, if Q-learning type algorithms are used and we approximate the optimal Q-function based on a linear model $\Phi^{\top}(x,a) \beta$ with some basis function $\Phi(x,a)\in \mathbb{R}^{\mathbb{M}}$. Then for any $\beta$, we can define a policy $\pi(\beta)$ as follows:
\begin{eqnarray*}
	\pi_{\beta}(a|x)=\left\{\begin{array}{ll}
		1, &\hbox{if}~~a=\sargmax_{a'\in \mathcal{A}} \Phi^{\top}(x,a') \beta,\\
		0, &\hbox{otherwise}.
	\end{array}\right.
\end{eqnarray*}
Then we have $\Pi=\{ \pi_{\beta}:\beta \in \mathbb{R}^{\mathbb{M}}\}$. 

\noindent (A3*.) Assume (i) and (ii) hold if $T\to \infty$ and (i) holds if $T$ is bounded. \\
(i) $\inf_{\pi\in \Pi}  \lambda_{\min} [ \sum_{t=0}^{T-1} \Mean\{\bm{\xi}_{0,t}\bm{\xi}_{0,t}^{\top}-\gamma^2 \bm{u}_{\pi}(X_{0,t},A_{0,t})\bm{u}^{\top}_{\pi}(X_{0,t},A_{0,t})\}]\ge \bar{c}T$ 
for some constant $\bar{c}>0$.  

\noindent (ii) The Markov chain $\{X_{0,t}\}_{t\ge 0}$ is geometrically ergodic.


We remark that Condition A3*(ii) is the same as A3(ii). 


\subsection{More on the margin condition}\label{sec:moremargin}
To better understand Condition A5, we consider a simple scenario where $\mathcal{A}=\{0,1\}$. Define $\tau(x)=Q^{\tiny{opt}}(x,1)-Q^{\tiny{opt}}(x,0)$. It follows that
\begin{eqnarray*}
	\max_{a} Q^{\tiny{opt}}(x,a)-\max_{a'\in \mathcal{A}-\argmax_{a} Q^{\tiny{opt}}(x,a)} Q^{\tiny{opt}}(x,a')=\left\{\begin{array}{ll}
		|\tau(x)|, & \hbox{if}~\tau(x)\neq 0,\\
		+\infty, & \hbox{otherwise}.
	\end{array}\right.
\end{eqnarray*}
As a result, \eqref{lambdamargin} and \eqref{Gmargin} are equivalent to the followings:
\begin{eqnarray}\label{advantage1}
\lambda\left\{ x\in\mathbb{X}: 0<|\tau(x)|\le \varepsilon \right\}=O(\varepsilon^{\alpha}),\\\label{advantage2}
\mathbb{G}\left\{ x\in\mathbb{X}: 0<|\tau(x)|\le \varepsilon \right\}=O(\varepsilon^{\alpha}).
\end{eqnarray}
Apparently, these two conditions hold when $\inf_{x\in \mathbb{X}}|\tau(x)|>0$. They are satisfied in many other cases. For example, let $d=1$. Consider
\begin{eqnarray*}
	\tau(x)=\left\{\begin{array}{ll}
		x^{1/\alpha}, & \hbox{if}~x>0,\\
		0, & \hbox{otherwise},
	\end{array}
	\right.
\end{eqnarray*}
for some $\alpha>0$. Then, with some calculations, we can show
\begin{eqnarray*}
	\lambda\left\{ x\in\mathbb{X}: 0<|\tau(x)|\le \varepsilon \right\}\le \lambda\{x: 0<x<\varepsilon^{\alpha} \}=\varepsilon^{\alpha}.
\end{eqnarray*}
This verifies \eqref{advantage1}. When $\mathbb{G}$ has a bounded density function on $\mathbb{X}$, \eqref{advantage2} is reduced to \eqref{advantage1}. If $\mathbb{G}(\cdot)$ equals the Dirac measure $\delta_x(\cdot)$, then \eqref{advantage2} automatically holds for any $\alpha>0$.
}

\section{Additional details regarding the method}
\subsection{More on the CI in \eqref{CI2}}\label{secmoreCI2}
We begin by providing more details on the estimators $\widehat{V}_{\mathcal{I}_{k+1}}(\widehat{\pi}_{\bar{\mathcal{I}}_k};\mathbb{G})$ and its standard error $\widehat{\sigma}_{\mathcal{I}_{k+1}}(\widehat{\pi}_{\bar{\mathcal{I}}_k};\mathbb{G})$. In general, for a given $\mathcal{I}\subseteq \mathcal{I}_0$ and any policy $\pi$, we define $\widehat{V}_{\mathcal{I}}(\pi;\mathbb{G})$ and $\widehat{\sigma}_{\mathcal{I}}(\pi;\mathbb{G})$ as
\begin{eqnarray*}
	&&\widehat{V}_{\mathcal{I}}(\pi;\mathbb{G})=\left\{\int_{x\in \mathbb{X}} \bm{U}_{\pi}(x) \mathbb{G}(dx) \right\}^{\top} \widehat{\bm{\beta}}_{\mathcal{I},\pi},\\
	&&\widehat{\sigma}^2_{\mathcal{I}}(\pi;\mathbb{G})=\left\{\int_{x\in \mathbb{X}} \bm{U}_{\pi}(x) \mathbb{G}(dx) \right\}^{\top} \widehat{\bm{\Sigma}}_{\mathcal{I},\pi}^{-1} \widehat{\bm{\Omega}}_{\mathcal{I},\pi} (\widehat{\bm{\Sigma}}_{\mathcal{I},\pi}^{\top})^{-1} \left\{\int_{x\in \mathbb{X}} \bm{U}_{\pi}(x) \mathbb{G}(dx) \right\},
\end{eqnarray*}
where
\begin{eqnarray*}
	\widehat{\bm{\beta}}_{\mathcal{I},\pi}=(\widehat{\beta}_{\mathcal{I},\pi,1}^{\top},\cdots,\widehat{\beta}_{\mathcal{I},\pi,m}^{\top})^{\top}=  \frac{1}{|\mathcal{I}|}\sum_{(i,t)\in \mathcal{I}} \widehat{\bm{\Sigma}}_{\mathcal{I},\pi}^{-1} \bm{\xi}_{i,t}Y_{i,t} ,\,\,\,\,
	\widehat{\bm{\Sigma}}_{\mathcal{I},\pi}=\frac{1}{|\mathcal{I}|}\sum_{(i,t)\in \mathcal{I}} \bm{\xi}_{i,t}(\bm{\xi}_{i,t}-\gamma \bm{U}_{\pi,i,t+1})^{\top},\\
	\widehat{\bm{\Omega}}_{\mathcal{I},\pi}=\frac{1}{|\mathcal{I}|}\sum_{(i,t)\in \mathcal{I}} \bm{\xi}_{i,t} \bm{\xi}_{i,t}^{\top} \{Y_{i,t}+\gamma \sum_{a\in \mathcal{A}} \Phi_L^{\top}(X_{i,t+1}) \widehat{\beta}_{\mathcal{I},\pi,a}\pi(a|X_{i,t+1})-\Phi_L^{\top}(X_{i,t}) \widehat{\beta}_{\pi,A_{i,t}}\}^2,
\end{eqnarray*}
and $|\mathcal{I}|$ stands for the number of elements in $\mathcal{I}$.

{\color{black}
	\subsection{Value difference between the target and behavior policy}\label{app:vd}
	In this section, we outline a method to evaluate the value difference function between the target and behavior policy. We first consider the scenario where the target policy is a fixed policy. We next consider the scenario where the target policy is an estimated optimal policy. To simplify the presentation, we assume $T_1=\cdots=T_n=T$. The proposed method can be similarly extended to on-policy settings.
	\subsubsection{Inference of the value difference under a fixed policy}
	Consider a data-independent policy $\pi$. We aim to evaluate the value difference function $\textrm{VD}(\pi;x)=V(\pi;x)-V(b;x)$ where $b$ is the unknown behavior policy. We first apply our method in Section \ref{sec:method} to compute an estimator value function $\widehat{V}(\pi;x)$ for $V(\pi;x)$. 
	
	To estimate $V(b;x)$, we observe that the Q-function $Q(b;x,a)$ satisfies the Bellman equation, $\Mean [\{Y_{i,t}+\gamma Q(b;X_{i,t+1},A_{i,t+1})-Q(b;X_{i,t},A_{i,t}) \}|X_{i,t},A_{i,t}]=0$. We approximate $Q(b;x,a)$ based on linear sieves $\Phi_L^\top (x) \beta_{b,a}^*$. Similar to Section \ref{sec:method}, $\bm{\beta}_b=(\beta_{b,1}^{*\top},\cdots,\beta_{b,m}^{*\top})^\top$ can be estimated by
	\begin{eqnarray*}
		\widehat{\bm{\beta}}_b=(\widehat{\beta}_{b,1}^\top,\cdots,\widehat{\beta}_{b,m}^\top)^\top=\biggl\{\underbrace{\frac{1}{nT} \sum_{i=1}^n \sum_{t=0}^{T-1} \bm{\xi}_{i,t} (\bm{\xi}_{i,t}- \gamma \bm{\xi}_{i,t+1})^{\top}}_{\widehat{\bm{\Sigma}}_b}\biggr\}^{-1} \left( \frac{1}{nT}\sum_{i=1}^n \sum_{t=0}^{T-1} \bm{\xi}_{i,t}Y_{i,t}  \right).
	\end{eqnarray*}
	The resulting estimates for $Q(b;x,a)$ can be derived as $\Phi_L^\top(x) \widehat{\beta}_{b,a}$. The corresponding estimator for $V(b,x)$ is given by $\widehat{V}(b,x)=\sum_a \widehat{b}(a|x)\Phi_L^\top(x) \widehat{\beta}_{b,a}$ where $\widehat{b}(a|x)$ denotes the sieve estimator $\Phi_L^\top(x)\widehat{\alpha}_a$ for $b(a|x)$ where
	\begin{eqnarray*}
		\widehat{\alpha}_a=\biggl\{\underbrace{\frac{1}{nT} \sum_{i=1}^n \sum_{t=0}^{T-1} \Psi_L(X_{i,t})\Psi_L^\top(X_{i,t})}_{\widehat{\bm{\Psi}}}\biggr\}^{-1}\left\{\frac{1}{nT}\sum_{i=1}^n \sum_{t=0}^{T-1} \Psi_L(X_{i,t})\mathbb{I}(A_{i,t}=a)  \right\}.
	\end{eqnarray*}
	This yields the estimator for the value difference $\widehat{\textrm{VD}}(\pi;x)=\widehat{V}(\pi;x)-\widehat{V}(b;x)$. 
	
	We next derive a confidence interval for VD$(\pi;x)$ based on $\widehat{\textrm{VD}}(\pi;x)$. Similar to the proof of Theorem \ref{thm1}, we can show $\sqrt{nT}\{\widehat{\textrm{VD}}(\pi;x)-\textrm{VD}(\pi;x)\}$ is equivalent to
	\begin{eqnarray}\label{eqn:valuediff123}
	\begin{split}
	&\bm{U}_{\pi}^\top(x) \bm{\Sigma}_{\pi}^{-1}\left(\frac{1}{\sqrt{nT}}\sum_{i=1}^n\sum_{t=0}^{T-1} \bm{\xi}_{i,t}\varepsilon_{i,t} \right)-\bm{U}_b^\top(x)\bm{\Sigma}_b^{-1} \left(\frac{1}{\sqrt{nT}}\sum_{i=1}^n\sum_{t=0}^{T-1} \bm{\xi}_{i,t}\varepsilon^*_{i,t} \right)\\-&\sum_{a}Q(b;a,x)\Psi_L^\top(x)\bm{\Psi}^{-1}\left[ \frac{1}{\sqrt{nT}}\sum_{i=1}^n \sum_{t=0}^{T-1} \Psi_L(X_{i,t})\{\mathbb{I}(A_{i,t}=a)-b(a|X_{i,t})\}  \right],
	\end{split}	
	\end{eqnarray}
	where $\varepsilon_{i,t}^*$ denotes the temporal difference error $Y_{i,t}+\gamma Q(b;X_{i,t+1},A_{i,t+1})-Q(b;X_{i,t},A_{i,t})$ and $\bm{\Psi}$ denotes the population limit of $\widehat{\Psi}$. 
	Note that the RHS can be rewritten as $(nT)^{-1/2}\sum_{i=1}^n\sum_{t=0}^{T-1} \psi_{i,t}$ that corresponds to a sum of martingale difference. Its variance can be consistently estimated by $\widehat{\sigma}^{*2}(\pi;x)=(nT)^{-1} \sum_{i=1}^n\sum_{t=0}^{T-1} \widehat{\psi}_{i,t}^2$ where $\widehat{\psi}_{i,t}$ denotes some consistent estimator for $\psi_{i,t}$ based on $\widehat{Q}(\pi;\cdot,\cdot)$, $\widehat{Q}(b;\cdot,\cdot)$ and $\widehat{b}$. The confidence interval for VD$(\pi;x)$ is given by $$[\widehat{\textrm{VD}}(\pi;x)-z_{\alpha/2} (nT)^{-1/2} \widehat{\sigma}^*(\pi;x),\widehat{\textrm{VD}}(\pi;x)+z_{\alpha/2} (nT)^{-1/2} \widehat{\sigma}^*(\pi;x)].$$

	\subsubsection{Inference of the value difference under an estimated optimal policy}
	We begin by dividing the data into $K$ non-overlapping subsets $\cup_{k=1}^K \mathcal{I}_k$. Similar to Section \ref{secCI}, we construct the value difference estimator by\begin{eqnarray*}
		\widetilde{\textrm{VD}}(x)=\left\{ \sum_{k=1}^{K-1} \frac{1}{\widehat{\sigma}^*_{\mathcal{I}_{k+1}}(\widehat{\pi}_{\bar{\mathcal{I}}_k};x)} \right\}^{-1} \left\{ \sum_{k=1}^{K-1} \frac{\widehat{\textrm{VD}}_{\mathcal{I}_{k+1}}(\widehat{\pi}_{\bar{\mathcal{I}}_k};x)}{\widehat{\sigma}^*_{\mathcal{I}_{k+1}}(\widehat{\pi}_{\bar{\mathcal{I}}_k};x)} \right\},
	\end{eqnarray*}
	where $\widehat{\textrm{VD}}_{\mathcal{I}_{k}}$ and $\widehat{\sigma}^*_{\mathcal{I}_{k}}$ denote the versions of VD and $\widehat{\sigma}^*$ based on samples in $\mathcal{I}_k$ only. The corresponding confidence interval is given by 
	\begin{eqnarray*}
		[\widetilde{\textrm{VD}}(x)-z_{\alpha/2} \{nT(K-1)/K\}^{-1/2} \widetilde{\sigma}^*(x), \widetilde{\textrm{VD}}(x)+z_{\alpha/2} \{nT(K-1)/K\}^{-1/2} \widetilde{\sigma}^*(x)],
	\end{eqnarray*}
	where $\widetilde{\sigma}^*(x)=(K-1)\{ \sum_{k=1}^{K-1} \widehat{\sigma}_k^{*-1}(\widehat{\pi}_{\bar{\mathcal{I}}_k};x) \}^{-1}$.  
	
	Finally, we remark that such a confidence interval might not be valid in the extreme case where the behavior policy is equal to a deterministic optimal policy. To elaborate, notice that when the behavior policy is deterministic, the second line of \eqref{eqn:valuediff123} equal zero. In addition, when $\pi=b=\pi^{\tiny{opt}}$ for some optimal policy $\pi^{\tiny{opt}}$, the first line equals zero as well. In that case, $\widehat{\textrm{VD}}(\pi;x)$ would have a degenerate distribution. Suppose the estimated optimal policy is consistent for $\pi^{\tiny{opt}}$. Then $\widetilde{\textrm{VD}}(x)$ might not have a tractable limiting distribution, leading to an invalid confidence interval. 
	
	To address this concern, we could redefine the inverse weights $\widehat{\sigma}^*_{\mathcal{I}_{k+1}}(\widehat{\pi}_{\bar{\mathcal{I}}_k};x)$ by $\widehat{\sigma}^*_{\mathcal{I}_{k+1}}(\widehat{\pi}_{\bar{\mathcal{I}}_k};x,\delta)=\max\{\widehat{\sigma}^*_{\mathcal{I}_{k+1}}(\widehat{\pi}_{\bar{\mathcal{I}}_k};x),\delta\}$ for some $\delta>0$, as in \cite{luedtke2017evaluating}. This guarantees that these inverse weights are strictly greater than zero. A similar approach is employed by \cite{shi2020sparse} for testing the overall qualitative treatment effects in single-stage decision making. 
	In addition, one could allow $\delta$ to depend on $N$ and $T$. The resulting confidence interval would be valid as long as $\delta\gg (NT)^{-1/6}$\citep[see e.g., Theorem 3.1 of][]{shi2020sparse}. However, a potential limitation is that it would yield a conservative confidence interval when the truncation is active, as discussed in \cite{luedtke2017evaluating}. 
}

\subsection{Double fitted $Q$-iteration}\label{secaddFQI}
In this section, we introduce our algorithm for computing the estimated optimal policy in our numerical studies. The proposed algorithm is based on FQI that recursively updates the estimated optimal Q-function by some supervised learning method (see Example 2 in Section \ref{secconvergeVpi}). In FQI, at each iteration, a maximization over estimated Q-function is used as an estimate of the maximum of the true Q-function. This can lead to a significant positive bias \citep{Sutton2018}. \cite{hasselt2010double} proposed a double Q-learning method to reduce the maximization bias. Here, we apply similar ideas to FQI to compute the estimated optimal policy. We use a pseudocode to summarize our algorithm below. 

\begin{algorithm}
	\KwIn{$\{(X_{i,t},A_{i,t},Y_{i,t},X_{i,t+1})\}_{0\le t< T_i,1\le i\le n}$, initialize parameters $\widehat{\theta}_{A}$, $\widehat{\theta}_{B}$.}
	\While{not convergence}{
		For $0\le t<T_i,1\le i\le n$, construct the target value: $\widehat{Q}^{A}_{i,t} =  Y_{i,t} + \gamma Q(X_{i, t + 1}, \argmax_{a'}Q(X_{i, t + 1}, a';\widehat{\theta}_{A} ), \widehat{\theta}_{B})$ \
		$\widehat{Q}^{B}_{i,t} =  Y_{i,t} + \gamma Q(X_{i, t + 1}, \argmax_{a'}Q(X_{i, t + 1}, a'; \widehat{\theta}_{B} ), 
		\widehat{\theta}_{A} )$ \\
		Update $\widehat{\theta}_A$ : $\widehat{\theta}_A \leftarrow \arg\min_{\theta_A} \sum_i\sum_t\|Q(X_{i,t}, A_{i,t};\theta_A) - \widehat{Q}^{A}_{i,t}\|^2$ \\
		Update $\widehat{\theta}_B$ : $\widehat{\theta}_B \leftarrow \arg\min_{
			\theta_B} \sum_i\sum_t\|Q(X_{i,t}, A_{i,t};\theta_B) - \widehat{Q}^{B}_{i,t}\|^2$
	}
	\caption{Double Fitted Q-iteration Algorithm}
	\label{algo: double-FQI}
\end{algorithm}

In Algorithm \ref{algo: double-FQI}, we can apply any non-parametric models $Q(\cdot,\code;\theta)$ indexed by $\theta$ to model the optimal Q-function. In our implementation, we set $Q(\cdot,\cdot;\cdot)$ to be a linear combination of tensor product B-spline basis functions. 

\section{Additional technical details}
\subsection{Additional details regarding Condition A3}\label{secaddcondA4}
When $\nu_0=\mu$, the density function of $X_{0,1}$ equals $\mu$ as well. By Jensen's inequality, we have for any $\bm{v}\in \mathbb{R}^{mL}$ that
\begin{eqnarray*}
	\{\bm{v}^{\top} \bm{u}_{\pi}(x,a)\}^2\le \Mean [\{\bm{v}^{\top} \bm{U}_{\pi}(X_{0,1}) \}^2|X_{0,0}=x,A_{0,0}=a],
\end{eqnarray*}
and hence
\begin{eqnarray*}
	\int_{x\in \mathbb{X}} \{\bm{v}^{\top} \bm{u}_{\pi}(x,a)\}^2b(a|x)\mu(x)dx\le \int_{x\in \mathbb{X}} \{\bm{v}^{\top} \bm{U}_{\pi}(x)\}^2\mu(x)dx.
\end{eqnarray*}
The matrix
\begin{eqnarray*}
	\int_{x\in \mathbb{X}} \sum_{a\in \mathcal{A}}\{\bm{U}_{\pi}(x)\bm{U}^{\top}_{\pi}(x)-\bm{u}_{\pi}(x,a)\bm{u}^{\top}_{\pi}(x,a)\}b(a|x)\mu(x)dx
\end{eqnarray*} 
is positive semidefinite. It follows that
\begin{eqnarray*}
	\lambda_{\min} \left[\int_{x\in \mathbb{X}} \sum_{a\in \mathcal{A}}\{\bm{\xi}(x,a)\bm{\xi}^{\top}(x,a)-\gamma^2\bm{u}_{\pi}(x,a)\bm{u}_{\pi}^{\top}(x,a)\}b(a|x)\mu(x)dx\right]\\
	\ge \lambda_{\min} \left[\int_{x\in \mathbb{X}} \sum_{a\in \mathcal{A}}\{\bm{\xi}(x,a)\bm{\xi}^{\top}(x,a)-\gamma^2\bm{U}_{\pi}(x)\bm{U}_{\pi}^{\top}(x)\}b(a|x)\mu(x)dx\right].
\end{eqnarray*}

When $\pi$ is a deterministic policy, $\sum_{a\in \mathcal{A}}\bm{\xi}(x,a)\bm{\xi}^{\top}(x,a)b(a|x)-\gamma^2\bm{U}_{\pi}(x)\bm{U}_{\pi}^{\top}(x)$ is a block diagonal matrix. To show A3(i) holds, it suffices to show
\begin{eqnarray*}
	\lambda_{\min}[\Phi_L(x)\Phi_L^{\top}(x)\{b(a|x)-\gamma^2 \pi(a|x)\}\mu(x)dx]>0,\,\,\,\,\,\,\,\,\forall a\in \mathcal{A}.
\end{eqnarray*}
Suppose $b$ is the $\epsilon$-greedy policy with respect to $\pi$, i.e, $b(a|x)=\epsilon m^{-1}+(1-\epsilon) \pi(a|x)$, for any $a\in \{1,\dots,m\}$ and $\epsilon$ satisfies $\epsilon\le 1-\gamma^2$, we have
\begin{eqnarray*}
	\lambda_{\min}\left[ \int_{x\in \mathbb{X}} \Phi_{L}(x)\Phi_L^{\top}(x) \{b(a|x)-\gamma^2\pi(a|x)\}\mu(x)dx \right]\ge \frac{1}{m}\lambda_{\min} \left\{\int_{x\in \mathbb{X}} \Phi_L(x)\Phi^{\top}_L(x)\mu(x)dx \right\}.
\end{eqnarray*} 
Suppose A2 holds. It suffices to require 
\begin{eqnarray}\label{condA4another}
\lambda_{\min}\left\{ \int_{x\in \mathbb{X}}  \Phi_L(x)\Phi^{\top}_L(x)dx \right\}>0.
\end{eqnarray}
The condition in \eqref{condA4another} is automatically satisfied \citep[see, e.g.,][]{Burman1989, Chen2015}.
{\color{black}\subsection{Additional details on the variance comparison}\label{sec:varcompare}
	We consider a randomized study where $b(a|x)$ is a constant function of $x$. In addition, we assume the target policy is nondynamic, i.e., $\pi(a^*|x)=1$ for some $1\le a^*\le m$ and any $x$. We impose the following conditions. 
	
	\noindent (C1) The process $\{X_{0,t}\}_{t\ge 0}$ is stationary.
	
	\noindent (C2) The temporal difference error $\varepsilon_{0,t}$ is independent of $(X_{0,t},A_{0,t})$.
	
	\noindent (C3) $\Mean \{\Phi_L(X_{0,t+1})\Phi_L^\top(X_{0,t+1})|A_{0,t}=a^*\}=\Mean \Phi_L(X_{0,t+1})\Phi_L^\top(X_{0,t+1})$.
	
	
	We make some remarks. First, Condition (C1) is imposed to simplify the presentation. The same results hold as long as $\{X_{0,t}\}_t$ will converge to its stationary distribution. Second, the variances of our estimator and DRL are very difficult to analyse in general. Conditions (C2)-(C3) are imposed to simplify the calculation. Even when these conditions are violated, we expect the variance of the proposed estimator will be smaller in general, as reflected in our numerical study. 
	
	\begin{theorem}\label{thm5}
		Assume (C1)-(C3) hold. Then the asymptotic variance of the DRL estimator is at least $h^{-1}(\Phi_L)\{1+\Var(w(\pi,X_{0,t}))\}$ times larger than the proposed estimator where $h(\Phi_L)= \int_{x}\Phi_L(x)\mathbb{G}(dx)\{\Mean \Phi_L(X_{0,t})\Phi_L^\top(X_{0,t})\}^{-1} \int_x \Phi_L(x)\mathbb{G}(dx)$ and $w$ is the marginalized density ratio \citep{kallus2019efficiently}.
	\end{theorem}
	
	We next investigate the value of the factor $h(\Phi_L)$. We set $\mathbb{G}$ and the stationary distribution of $X_{0,t}$, i.e., $\mu$ to a uniform distribution on $[0,1]$. We consider a polynomial basis function and a B-spline basis function. Figure \ref{fig:E1} depicts the value of this factor with different choices of $L$. We also tried several other combinations of $\mathbb{G}$ and $\mu$, and find this factor is in general smaller than or very close to $1$. Since $\Var(w(\pi,X_{0,t}))>0$, Theorem \ref{thm5} implies that the proposed estimator achieves smaller variance. 
	
	\begin{figure}
		\centering
		\includegraphics[width=10cm]{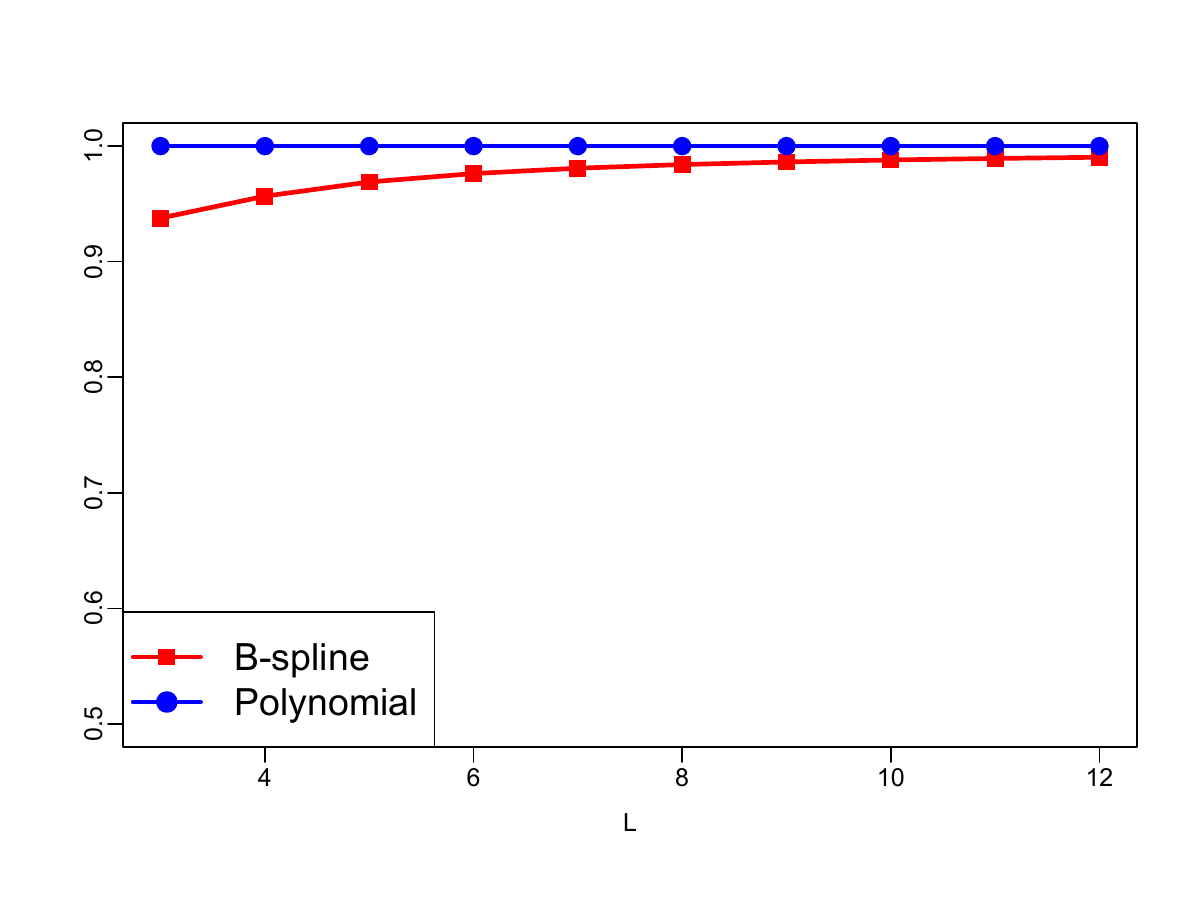}
		\vspace{-1cm}
		\caption{$h(\Phi_{0,L})$ with different choices of $\Phi_{0,L}$.}\label{fig:E1}
	\end{figure}

	We next sketch a few lines to prove Theorem \ref{thm5}. Based on Theorem \ref{thm1}, the asymptotic variance of our estimator is given by
	\begin{eqnarray}\label{eqn:varour}
	(nT)^{-1} \left\{ \int_{x\in \mathbb{X}} \bm{U}_{\pi}(x)\mathbb{G}(dx) \right\}^{\top}  \bm{\Sigma}_{\pi}^{-1} \bm{\Omega} (\bm{\Sigma}_{\pi}^{-1})^\top\left\{ \int_{x\in \mathbb{X}} \bm{U}_{\pi}(x)\mathbb{G}(dx) \right\},
	\end{eqnarray}
	where $\bm{\Omega}$ is defined in Step 2 of the proof of Theorem \ref{thm1}. Under (C1) and (C2), we have $\bm{\Omega}=\Mean \bm{\xi}_{0,t} \bm{\xi}_{0,t}^\top \varepsilon_{0,t}^2=\sigma_*^2 \Mean \bm{\xi}_{0,t} \bm{\xi}_{0,t}^\top$ where $\sigma_*^2$ is the variance of $\varepsilon_{0,t}$. 
	
	Using similar arguments in the proof of Theorem 16 of \cite{kallus2019efficiently}, we can show that the asymptotic variance of the DRL estimator equals
	\begin{eqnarray*}
		\frac{1}{nT(1-\gamma)^2}\Mean w^2(X_{0,0}) \frac{\mathbb{I}(A_{0,0}=a^*)}{b^2(A_{0,0}|X_{0,0})}\varepsilon_{0,0}^2.
	\end{eqnarray*}
	Under the given conditions, the above variance is equal to $\{nT(1-\gamma)^2 p_{a^*}\}^{-1} \sigma_*^2 \{1+\Var(\omega(X_{0,0}))\}$ where $p_{a^*}=\prob(A_{0,t}=a^*)$. Consequently, it suffices to show
	\begin{eqnarray}\label{eqn:0}
	\left\{ \int_{x\in \mathbb{X}} \bm{U}_{\pi}(x)\mathbb{G}(dx) \right\}^{\top}  \bm{\Sigma}_{\pi}^{-1} \Mean \bm{\xi}_{0,t}\bm{\xi}_{0,t}^\top (\bm{\Sigma}_{\pi}^{-1})^\top\left\{ \int_{x\in \mathbb{X}} \bm{U}_{\pi}(x)\mathbb{G}(dx) \right\}\le \frac{h(\Phi_{0,L})}{(1-\gamma)^2 p_{a^*}}. 
	\end{eqnarray}
	Since $\pi$ is a nondynamic policy, $\int_{x\in \mathbb{X}} \bm{U}_{\pi}(x)\mathbb{G}(dx)$ is a sparse vector that takes the following form:
	\begin{eqnarray*}
		\Big\{\underbrace{0,\cdots,0}_{(a^*-1)L~\textrm{many}~0s},\int_{ \mathbb{X}} \Phi_L(x)\mathbb{G}(dx),\underbrace{0,\cdots,0}_{(m-a^*)L~\textrm{many}~0s} \Big\}.
	\end{eqnarray*}
	By the definition of $\bm{\Sigma}_{\pi}$ and $\bm{\xi}_{0,t}$, the left-hand-side of \eqref{eqn:0} is equal to
	\begin{eqnarray*}
		\left\{\int_{ \mathbb{X}} \Phi_L(x)\mathbb{G}(dx)\right\}^{\top} [\Mean \Phi_{L}(X_{0,t})\mathbb{I}(A_{0,t}=a^*)\{\Phi_L(X_{0,t})-\gamma \Phi_L(X_{0,t+1})\}^\top]^{-1\top } \bm{\Omega}_{a^*}\\
		\times [\Mean \Phi_{L}(X_{0,t})\mathbb{I}(A_{0,t}=a^*)\{\Phi_L(X_{0,t})-\gamma \Phi_L(X_{0,t+1})\}^\top]^{-1} \left\{\int_{ \mathbb{X}} \Phi_L(x)\mathbb{G}(dx)\right\},
	\end{eqnarray*} 
	where $\bm{\Omega}_{a^*}=\Mean \Phi_{L}(X_{0,t})\mathbb{I}(A_{0,t}=a^*) \Phi_L^\top(X_{0,t})$. Using similar arguments in \ref{secaddcondA4}, we can show
	\begin{eqnarray*}
		&&\bm{a}^\top \bm{\Omega}_{a^*}^{-1/2} [\Mean \Phi_{L}(X_{0,t})\mathbb{I}(A_{0,t}=a^*)\{\Phi_L(X_{0,t})-\gamma \Phi_L(X_{0,t+1})\}] \bm{\Omega}_{a^*}^{-1/2} \bm{a}\\
		&=& \bm{a}^\top \bm{\Omega}_{a^*}^{-1/2} \Mean \Phi_{L}(X_{0,t})\mathbb{I}(A_{0,t}=a^*)\Phi_L^\top(X_{0,t}) \bm{\Omega}_{a^*}^{-1/2} \bm{a}\\
		&-& \gamma\bm{a}^\top \bm{\Omega}_{a^*}^{-1/2}\Mean \Phi_{L}(X_{0,t})\mathbb{I}(A_{0,t}=a^*)\Phi_L^\top(X_{0,t+1})\bm{\Omega}_{a^*}^{-1/2} \bm{a}\\
		&\ge& \|\bm{a}\|_2^2-\gamma \|\bm{a}\|_2 \sqrt{\bm{a}^\top\bm{\Omega}_{a^*}^{-1/2}\Mean\mathbb{I}(A_{0,t}=a^*)\Phi_L(X_{0,t+1})\Phi_L^\top(X_{0,t+1})\bm{\Omega}_{a^*}^{-1/2}\bm{a}}=(1-\gamma) \|\bm{a}\|_2^2,
	\end{eqnarray*}
	under (C3). Consequently, we have $$\|\bm{\Omega}_{a^*}^{1/2} [\Mean \Phi_{L}(X_{0,t})\mathbb{I}(A_{0,t}=a^*)\{\Phi_L(X_{0,t})-\gamma \Phi_L(X_{0,t+1})\}]^{-1} \bm{\Omega}_{a^*}^{1/2}\|_2\le (1-\gamma).$$ As such, the left-hand-side of \eqref{eqn:0} is upper bounded by 
	\begin{eqnarray*}
		(1-\gamma)^{-2} \left\{\Mean \int_{ \mathbb{X}} \Phi_L(x)\mathbb{G}(dx)\right\}^{\top} \bm{\Omega}_{a^*}^{-1} \left\{\int_{ \mathbb{X}} \Phi_L(x)\mathbb{G}(dx)\right\}=(1-\gamma)^{-2} p_{a^*}^{-1} \left\{\int_{ \mathbb{X}} \Phi_L(x)\mathbb{G}(dx)\right\}^{\top} \\
		\times \{\Mean \Phi_{0,L}(X_{0,t})\Phi_{0,L}^\top(X_{0,t}) \}^{-1} \left\{\int_{ \mathbb{X}} \Phi_L(x)\mathbb{G}(dx)\right\}=(1-\gamma)^{-2} p_{a^*}^{-1} h(\Phi_{0,L}). 
	\end{eqnarray*}
	This completes the proof. }

\subsection{Additional details on on-policy evaluation}\label{secmoreonpolicy}
In this section, we show our proposed CI in Section \ref{seconpolicy} achieves nominal converge. To simplify the analysis, we focus on the setting where $K$ is finite, $T(1)=\cdots=T(K)=T$ and $L(1)=\cdots=L(K)=L$. When $K$ diverges, the sequences $\{T(k)\}_{k\ge 1}$ and $\{L(k)\}_{k\ge 1}$ shall be properly chosen to reduce the bias of the value estimates. We leave this for future research. 

Similar to Appendix \ref{seccondA4A5}, we assume the estimated policy $\widehat{\pi}_{\mathcal{I}}\in \Pi$ with probability $1$, for any $\mathcal{I}$. In on-policy settings, the behavior policy $b_{\pi}$ is a function of the estimated policy $\pi \in \Pi$. For instance, when an $\epsilon$-greedy policy is used to determine the behavior policy, then we have $b_{\pi}= (1-\epsilon)\pi+\epsilon \pi^*$ where $\pi^*$ denotes a uniform random policy. Let $\mathcal{B}=\{b_{\pi}:\pi\in\Pi\}$. 

For any behavior policy $b\in \mathcal{B}$, consider a Markov chain $\{(X_{0,t,b},A_{0,t,b})\}_{t\ge 0}$ generated by this behavior policy. Let $Y_{0,t,b}$ be the realization of the immediate reward at time $t$. Let $\mu_b$ denote the limiting distribution of the Markov chain $\{X_{0,t,b}\}_{t\ge 0}$, and $\mathcal{P}^{t,b}_X(\cdot|x)$ be its $t$-step transition kernel. For any $x\in \mathbb{X},a\in \mathcal{A}$, define $\omega_{\pi,b}(x,a)$ as
\begin{eqnarray*}
	\Mean \left[\left\{Y_{0,0,b}+\gamma \sum_{a\in \mathcal{A}} \pi(a|X_{0,1,b}) Q(\pi;X_{0,1,b},a)-Q(\pi;X_{0,0,b},A_{0,0,b})\right\}^2 |X_{0,0,b}=x,A_{0,0,b}=a\right].
\end{eqnarray*}

We introduce the following conditions. 

\smallskip
\noindent (A2'.) Assume $\nu_0$ and $q$ are uniformly bounded away from $0$ and $\infty$ on their supports.

\noindent (A3'.) Assume (i) and (ii) hold if $T\to \infty$ and (iii) holds if $T$ is bounded. \\
\noindent (i) $\inf_{\pi\in \Pi,b\in \mathcal{B}}\lambda_{\min} [\int_{x\in \mathbb{X}} \sum_{a\in \mathcal{A}}\{\bm{\xi}(x,a)\bm{\xi}^{\top}(x,a)-\gamma^2\bm{u}_{\pi}(x,a)\bm{u}_{\pi}^{\top}(x,a)\}b(a|x)\mu_{b}(x)dx]\ge \bar{c} $ for some constant $\bar{c}>0$.  

\noindent (ii) There exists some function $M(\cdot)$ on $\mathbb{X}$ and some constant $\rho<1$ such that $$\sup_{\pi\in\Pi}\int_{x\in \mathbb{X}} M(x) \mu_{b_\pi}(x)dx<+\infty,$$ and 
\begin{eqnarray*}
	\sup_{\pi\in \Pi}\|\mathcal{P}^{t,b_\pi}_X(\cdot|x)-\mu_{b_\pi}(\cdot)\|_{TV}\le M(x) \rho^t,\,\,\,\,\,\,\,\,\forall t\ge 0.
\end{eqnarray*} 

\noindent (iii) There exists some constant $\bar{c}>0$ such that
\begin{eqnarray*}
	\inf_{\pi\in \Pi,b \in \mathcal{B}}  \lambda_{\min} [ \sum_{t=0}^{T-1} \Mean\{\bm{\xi}(X_{0,t,b},A_{0,t,b})\bm{\xi}(X_{0,t,b},A_{0,t,b})^{\top}-\gamma^2 \bm{u}_{\pi}(X_{0,t,b},A_{0,t,b})\bm{u}^{\top}_{\pi}(X_{0,t,b},A_{0,t,b})\}]\ge \bar{c}T.
\end{eqnarray*}

\noindent (A4') For any $1\le k\le K$, we have
$\Mean  |V(\widehat{\pi}_{\bar{\mathcal{I}}_k};\mathbb{G})-V(\pi^*;\mathbb{G})|=O(|\bar{\mathcal{I}}_k|^{-b_0})$, 
for some $b_0>1/2$ such that $(nT)^{b_0-1/2}\gg \|\int_x \Phi_L(x)\mathbb{G}(dx)\|_2^{-1}$, where the big-$O$ term is uniform in $\mathcal{I}$. 

\begin{theorem}\label{thm2.5}
	Assume A1, A2'-A4' hold. Suppose 
	$L=o\{\sqrt{nT}/\log (nT)\}$ and $L^{2p/d}\gg nT\{1+\|\int_{x}\Phi_L(x)\mathbb{G}(dx)\|_2^{-2}\}$. Assume there exists some constant $c_0\ge 1$ such that $\omega_{\pi,b}(x,a)\ge c_0^{-1}$ for any $x,a,\pi,b$ and $\prob(\max_{0\le t\le T-1} |Y_{0,t}|\le c_0)=1$. Then as either $n\to \infty$ or $T\to \infty$, 
	\begin{eqnarray*}
		&&\sqrt{nT(K-1)}\widetilde{\sigma}^{-1}(\mathbb{G})\{\widetilde{V}(\mathbb{G})-V(\widehat{\pi};\mathbb{G})\}\stackrel{d}{\to} N(0,1),\\
		&&\sqrt{nT(K-1)}\widetilde{\sigma}^{-1}(\mathbb{G})\{\widetilde{V}(\mathbb{G})-V(\pi^*;\mathbb{G})\}\stackrel{d}{\to} N(0,1).
	\end{eqnarray*}
\end{theorem}
Proof of Theorem \ref{thm2.5} is omitted for brevity.

\section{Additional numerical results}
{\color{black}\subsection{Data-adaptive selection of $L$}\label{sec}
	We apply the proposed method detailed in Section \ref{sec:disnumber} to Scenario (B) where the treatment assignment mechanism depends on the observed state, to investigate the finite sample performance of the resulting CI. Specifically, we apply the random forest algorithm to learn the conditional mean function $\mu$ and the behavior policy $b$. 
	We assume $\Sigma(a,x)$ is a constant function of $(a,x)$ and estimated it by
	\begin{eqnarray*}
		\widehat{\Sigma}=\frac{1}{nT} \sum_{i,t} \{X_{i,t+1}-\widehat{\mu}(A_{i,t},X_{i,t})\}\{X_{i,t+1}-\widehat{\mu}(A_{i,t},X_{i,t})\}^\top.
	\end{eqnarray*}
	We use the tensor product B-spline basis for $\Phi_L$, as in Section \ref{secsimu}. Note that the state is a two-dimensional vector, $L$ is selected among the set $\{4^2,5^2,6^2,7^2,8^2,9^2\}$. Specifically, we choose $L$ such that the resulting CI is the shortest among all CIs whose coverage probabilities are above 93\%. If no such CI exists, we select the CI with the highest coverage probability. 
	
	We report the ECP and AL of the resulting CI in the left and middle panels of Figure \ref{fig0}. It can been seen that ECP is close to the nominal level in all cases and AL decays as either $n$ or $T$ increases. In the right panel of Figure \ref{fig0}, we report the number of basis functions that is being selected most by the proposed method as a function of $n$ and $T$ (denote by $L(n,T)$). It is clear from Figure \ref{fig0} that $L(n,T)$ increases with the total number of observations $nT$. This is consistent with the following intuition: as $nT$ increases, more basis functions are needed to reduce the approximation error and guarantee the nominal coverage of the resulting CI. 
	\begin{figure}
		\centering
		\includegraphics[width=15cm]{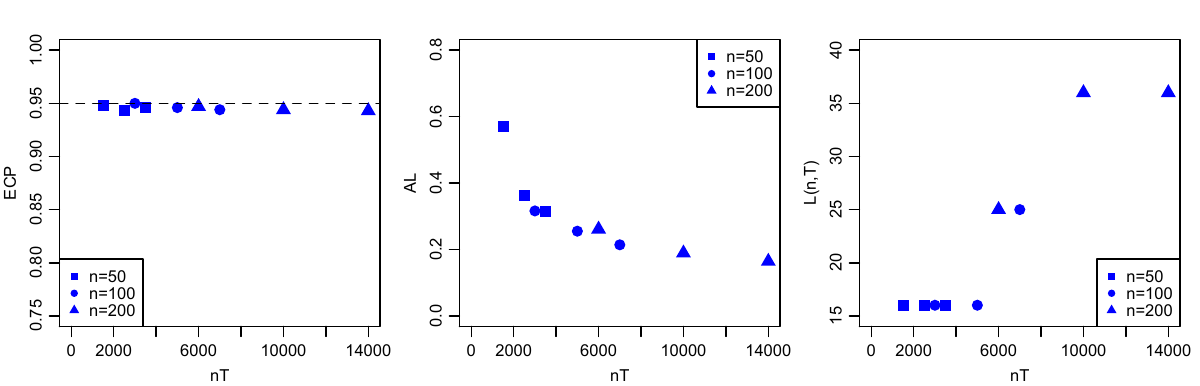}
		\caption{Empirical coverage probabilities (ECP) and average lengths (AL) of CIs constructed by the proposed method detailed in Section \ref{sec:disnumber} as well as the number of basis functions being selected most, with different combinations of $n$ and $T$. }\label{fig0}
	\end{figure}
}

\subsection{Sensitivity analysis}\label{secsen}
\subsubsection{Sensitivity test for $\eta$} \label{secsensitivity}

\begin{figure}[!t]
	\caption{}\label{figsensi}
	\centering
	\includegraphics[width=12cm]{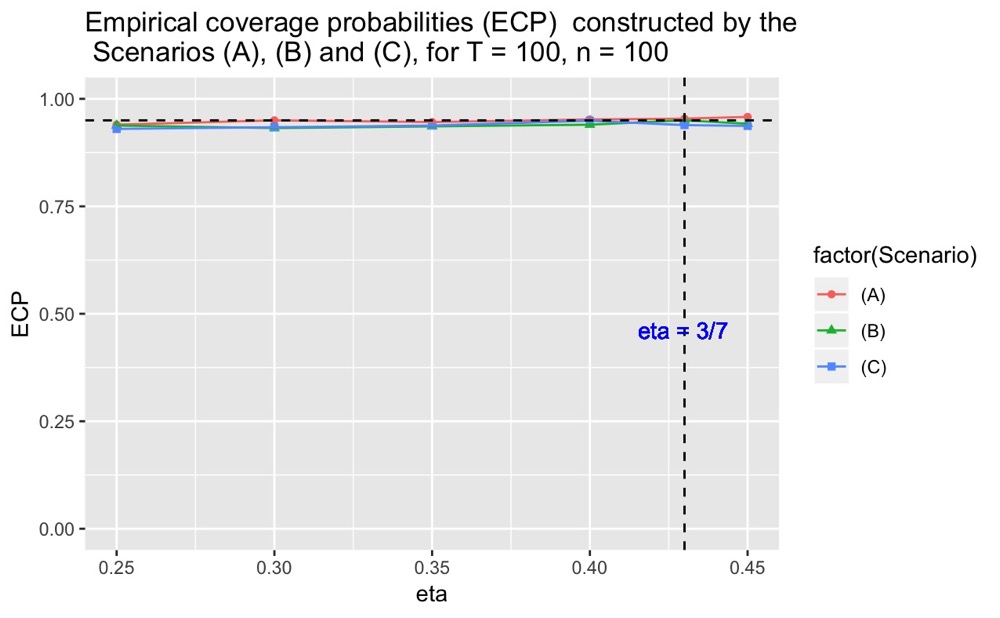}
\end{figure}

In this section, we conduct the sensitivity test for the parameter $\eta$ in the number of basis $L = \lfloor (nT)^{\eta} \rfloor$. We consider the simulation of the off-policy evaluation with a fixed target policy in Section 5.1. For scenario (A), (B) and (C), we set $n = 100$, $T = 100$ and the different $\eta$'s are chosen from $(0.25,0.30,0.35,0.40,3/7,0.45)$. The result of the ECPs are plotted in Figure \ref{figsensi} where all the ECPs are close to the nominal coverage rate 0.95. It shows that the results of the coverage are not sensitive to the different choices of $\eta$. 

{\color{black}\subsubsection{Sensitivity test for $\gamma$} \label{secsensitivity2}
	In Figure \ref{fig10}, we report the ECP and AL of the proposed CI and the MSE of our value estimate under Scenario B where the target policy is fixed, with $\gamma=0.3$ and $0.7$. In Figure \ref{fig11}, we report the ECP and AL of the proposed CI and the MSE of our value estimate under Scenario B where the target policy is an estimated optimal policy, with $\gamma=0.3$ and $0.7$. It can be seen that findings are very similar to those with $\gamma=0.5$. 
	
	In Tables \ref{tab4} and \ref{tab5}, we report the ECP, AL and MSE of the proposed method and DRL under Scenario (C) where the target policy is fixed, with $\gamma=0.3$ and $0.7$. It can be seen that the proposed CI achieves nominal coverage in all cases. When $\gamma=0.3$, ECP of the DRL method is well below the nominal level in all cases. When $\gamma=0.7$, the AL and MSE of the proposed CI are much smaller than those based on DRL. 
	
	\begin{figure}[!t]
		\caption{Empirical coverage probabilities and average lengths of CIs constructed by the proposed method as well as the mean squared errors of the value estimates, under Scenario (B), with different choices of $n$ and $T$. The target policy is fixed. $\gamma=0.3$ and $0.7$, from top to bottom panels.}\label{fig10}
		\centering
		\includegraphics[width=12.5cm]{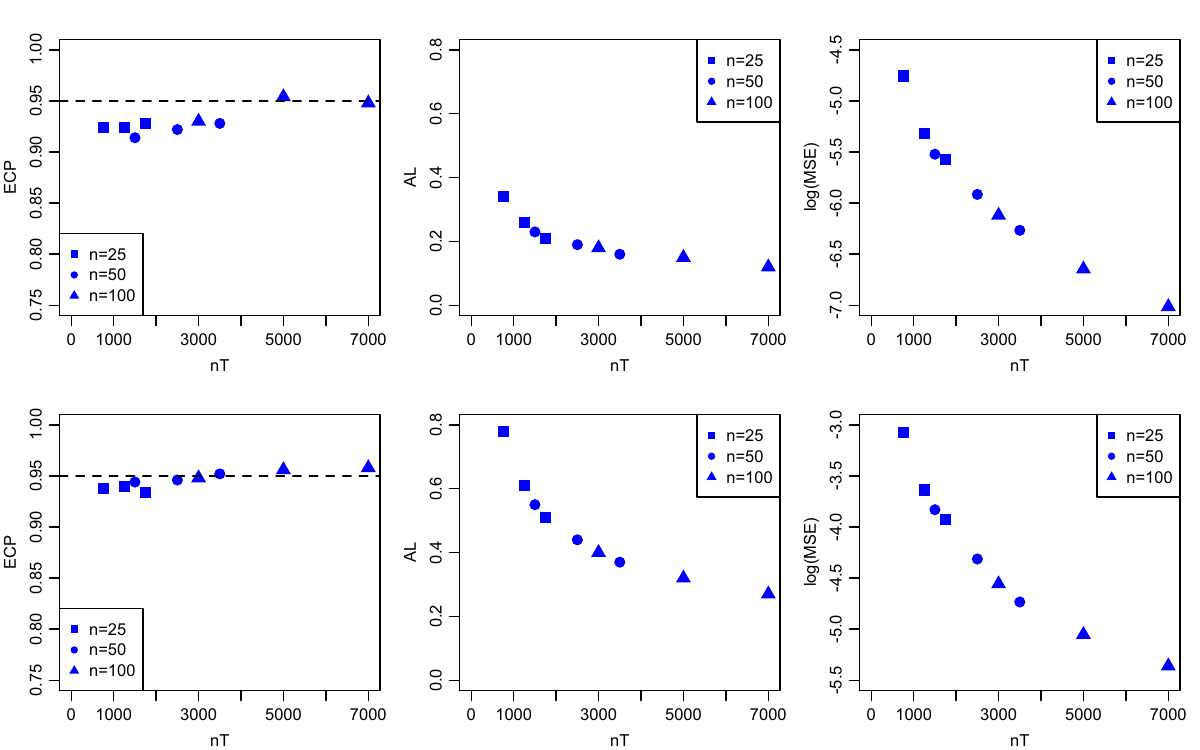}
	\end{figure}
	
	\begin{figure}[!t]
		\caption{Empirical coverage probabilities and average lengths of CIs constructed by the proposed method as well as the mean squared errors of the value estimates, under Scenario (B), with different choices of $n$ and $T$. The target policy is an estimated optimal policy. $\gamma=0.3$ and $0.7$, from top to bottom panels.}\label{fig11}
		\centering
		\includegraphics[width=12.5cm]{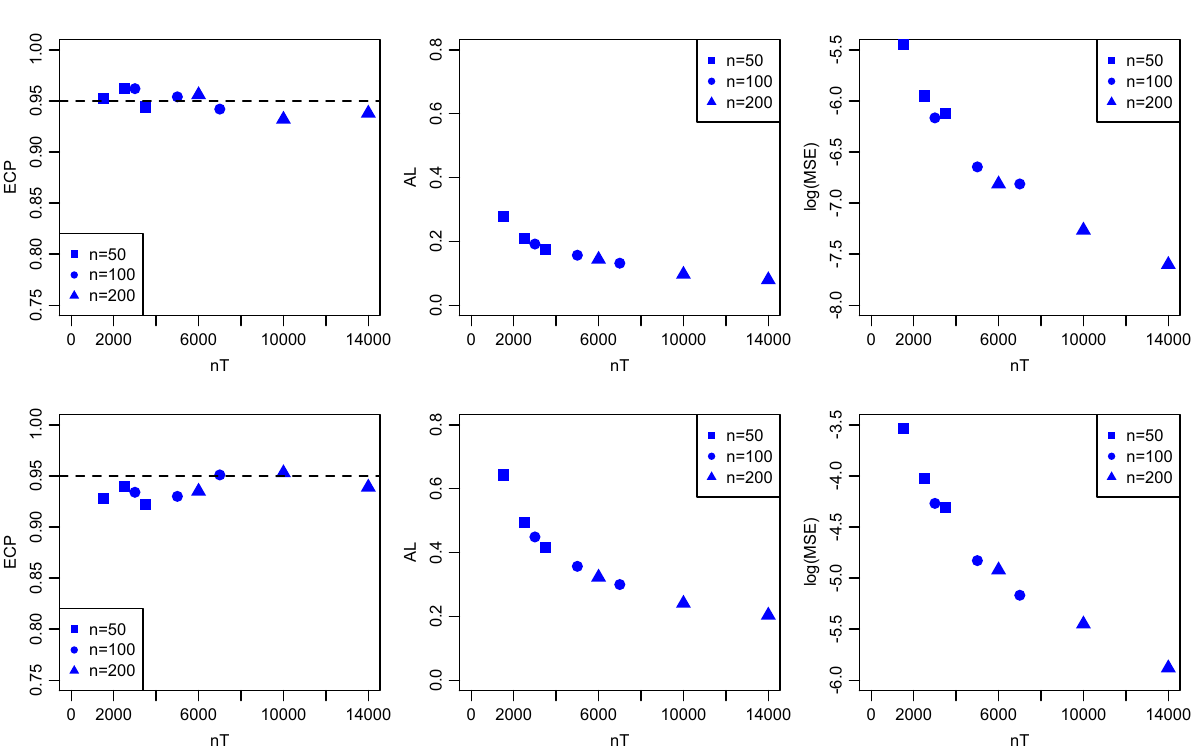}
	\end{figure}
	
	\begin{table}
		\caption{Empirical coverage probabilities (ECP) and average lengths (AL) of CIs constructed by the proposed method and DRL as well as the mean-squared errors (MSE) of the corresponding value estimators under Scenario (C), with $\gamma=0.3$.}\label{tab4}
		\centering
		\begin{tabular}{|c|c|c|c|c|c|c|}\hline
			& \multicolumn{3}{c}{SAVE} & \multicolumn{3}{c}{DRL}\\ \hline
			n & ECP & AL & log(MSE) & ECP & AL & log(MSE)\\ \hline 
			500 & 0.93 & 0.09 & -7.13 & 0.76 & 0.08 & -7.01 \\ \hline
			1000 & 0.95 & 0.07 & -8.22 & 0.76 & 0.06 & -7.60\\ \hline
			1500 & 0.94 & 0.05 & -8.52 & 0.79 & 0.05 & -7.82\\\hline 
		\end{tabular}
	\end{table}
	
	\begin{table}
		\caption{Empirical coverage probabilities (ECP) and average lengths (AL) of CIs constructed by the proposed method and DRL as well as the mean-squared errors (MSE) of the corresponding value estimators under Scenario (C), with $\gamma=0.7$.}\label{tab5}
		\centering
		\begin{tabular}{|c|c|c|c|c|c|c|}\hline
			& \multicolumn{3}{c}{SAVE} & \multicolumn{3}{c}{DRL}\\ \hline
			n & ECP & AL & log(MSE) & ECP & AL & log(MSE)\\ \hline 
			500 & 0.94 & 0.11 & -8.11 & 0.99 & 0.34 & -5.36 \\ \hline
			1000 & 0.96 & 0.11 & -7.60 & 0.99 & 0.24 & -5.99\\ \hline
			1500 & 0.94 & 0.08 & -8.11 & 0.99 & 0.19 & -6.81\\\hline 
		\end{tabular}
	\end{table}
	
	\subsubsection{Sensitivity test for the ordering of trajectories}\label{sec:senordering}
	We focus on Scenario (B), detailed in Section \ref{Sim_1}, to examine the sensitivity of the proposed CI to the ordering of trajectories. Specifically, we first randomly permute all trajectories with some fixed random seed. We next apply our SAVE procedure to construct the CI. We use three random seeds to generate different random permutations and depict the corresponding results in Figure \ref{fig12}. It can be seen that our method is not overly sensitive to the ordering of trajectories. 
	
	\begin{figure}[!t]
		\caption{Empirical coverage probabilities and average lengths of CIs constructed by the proposed method under Scenarios (B), with different choices of $n$ and $T$. From top plots to bottom plots, we use three different random seeds to generate the random permutation applied to all trajectories. }\label{fig12}
		\centering
		\includegraphics[width=12cm]{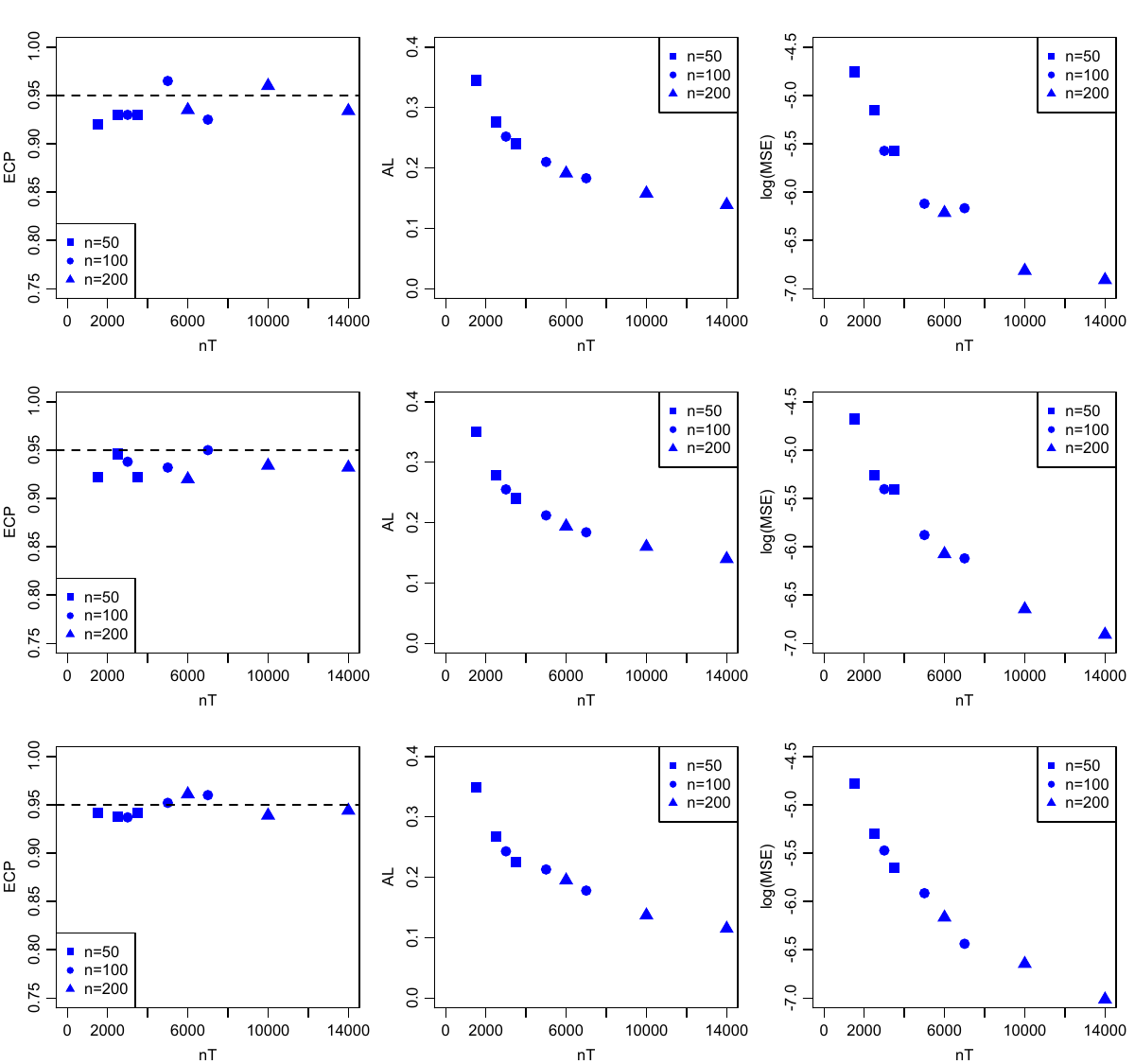}
	\end{figure}
}
\subsection{Additional settings}\label{sec:addsimu}
\begin{figure}[!t]
	\caption{Empirical coverage probabilities and average lengths of CIs constructed by the proposed method under Scenarios (A), with different choices of $n$ and $T$. The target policy is fixed. The reference distribution $\mathbb{G}$ corresponds to $(0.5, 0.5)^{\top}$ and $(-0.5, -0.5)^\top$, from top plots to bottom plots.}\label{figS1}
	\centering
	\includegraphics[width=9cm]{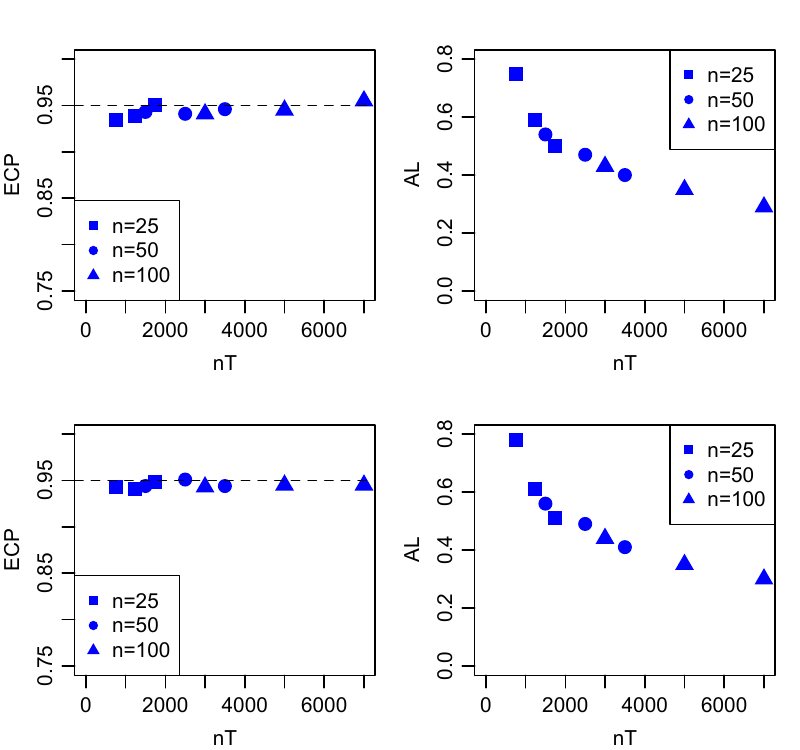}
\end{figure}

\begin{figure}[!t]
	\caption{Empirical coverage probabilities and average lengths of CIs constructed by the proposed method under Scenarios (A), with different choices of $n$ and $T$. The target policy is an estimated optimal policy. The reference distribution $\mathbb{G}$ corresponds to $(0.5, 0.5)^{\top}$ and $(-0.5, -0.5)^\top$, from top plots to bottom plots.}\label{figS2}
	\centering
	\includegraphics[width=9cm]{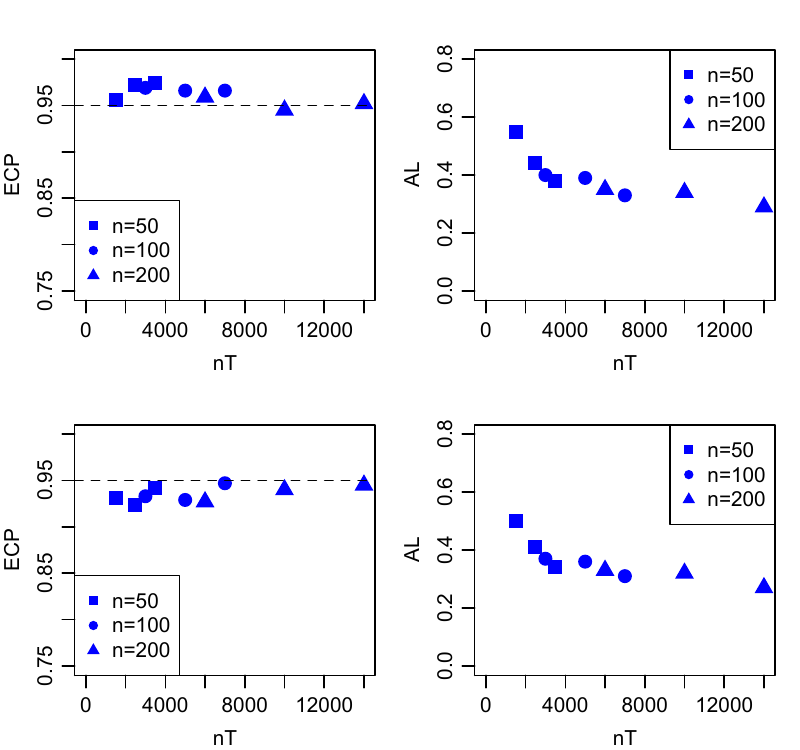}
\end{figure}

In this section, we conduct additional simulation studies to investigate the finite sample performance of the proposed method under settings where the reference distribution $\mathbb{G}$ is a Dirac delta function. Specifically, we consider the settings in Scenario (A) and set $\mathbb{G}$ to $(0.5, 0.5)^{\top}$ and $(-0.5, -0.5)^\top$. It can be seen from Figures \ref{figS1} and \ref{figS2} that our CIs achieve nominal coverage and their lengths decrease as $nT$ increases. 

{\color{black}\subsection{Additional real data results}\label{secaddrealdata}
	We use our real data example to discuss the issue of over-fitting in this section. Specifically, we apply the proposed method in Section \ref{secCI} to evaluate the optimal value $V(\pi^{\tiny{opt}};X_{i,0})$ starting from the initial state variable $X_{i,0}$, for $i=1,2,\cdots,6$. When the initial starting time is 8:00 am in Day 1, CIs for Patient 5 and Patient 6 are $[-6.288,3.287]$ and $[-6.313, 10.644]$, respectively. Both upper bounds are positive. However, according to our definition, the immediate reward is nonpositive. As such, the value and Q-function shall be nonpositive as well. This reflects one of the drawback of the proposed method. The resulting Q-estimator might suffer from over-fitting, leading to an unbounded outcome.
	
	Specifically, it is due to that the matrix $\widehat{\bm{\Sigma}}_{\pi}$ is close to singular. Note that the regression coefficients $\widehat{\bm{\beta}}_{\pi}$ are computed by solving the linear equation 
	\begin{eqnarray*}
		\widehat{\bm{\Sigma}}_{\pi}\widehat{\bm{\beta}}_{\pi}=\frac{1}{\sum_i T_i} \sum_{i=1}^n \sum_{t=0}^{T_i-1} \bm{\xi}_{i,t}Y_{i,t}.
	\end{eqnarray*}
	In our data example, the number of basis function equals 12. As such, $\widehat{\bm{\Sigma}}_{\pi}$ is a $12$ by $12$ matrix. When it is close to singular, the resulting Q-estimator might be unbounded. 
	
	To avoid offer-fitting, we note that in theory, $\widehat{\bm{\Sigma}}_{\pi}$ is a positive definite matrix under Condition (A3)(i). This motivates us to compute $\widehat{\bm{\beta}}_{\pi}$ by solving
	\begin{eqnarray*}
		(\widehat{\bm{\Sigma}}_{\pi}+\lambda I) \widehat{\bm{\beta}}_{\pi}=\frac{1}{\sum_i T_i} \sum_{i=1}^n \sum_{t=0}^{T_i-1} \bm{\xi}_{i,t}Y_{i,t},
	\end{eqnarray*}
	where $I$ denotes the identity matrix. As long as $\lambda$ satisfies $\lambda=O(N^{-1} T^{-1})$, the proposed CI remains valid. In our real data example, we set $\lambda=1\times 10^{-9}$. The resulting CIs for Patient 5 and Patient 6 are $[-8.034, -5.667]$ and $[-9.866, -7.104]$. Both upper bounds are strictly negative. 
}
\section{Technical proofs}\label{secproof}
For any two positive sequences $\{a_t\}_{t\ge 1}$ and $\{b_t\}_{t\ge 1}$, we write $a_t\preceq b_t$ if there exists some constant $C>0$ such that $a_t\le Cb_t$ for any $t$. The notation $a_t\preceq 1$ means $a_t=O(1)$. We will use $C,\bar{C}>0$ to denote some universal constants whose values are allowed to change from place to place. Let $q_X(\cdot|x)$ denote the density function of $\mathcal{P}_X(\cdot|x)$. Define $\mathbb{S}^{mL-1}$ as the unit sphere $\{\bm{v}\in \mathbb{R}^{mL}:\|\bm{v}\|_2=1\}$. When splines are used to estimate the Q-function, we assume the internal knots are equally spaced. 

The rest of the section is organized as follows. We first present the proof sketches for Theorems \ref{thm1}-\ref{thm4}. We next present the detailed technical proofs.
\subsection{A sketch for the proof of Theorem \ref{thm1}}\label{secsketchproofthm1}
We provide an outline for the proof in this section. The detailed proof can be found in Section \ref{secproofthm1} of the supplementary article. We break the proof into three steps. In the first step, we show the estimator $\widehat{\bm{\beta}}_{\pi}$  satisfies
\begin{eqnarray}\label{firststep}
\widehat{\bm{\beta}}-\bm{\beta}^*=\bm{\Sigma}_{\pi}^{-1} \left( \frac{1}{nT}\sum_{i=1}^n \sum_{t=0}^{T-1} \bm{\xi}_{i,t}\varepsilon_{i,t} \right)+O_p(L^{-p/d})+O_p\{L(nT)^{-1}\log(nT)\},
\end{eqnarray}
where $\bm{\Sigma}_{\pi}=\Mean \widehat{\bm{\Sigma}}_{\pi}$. The proof of \eqref{firststep} relies on some random matrix inequalities established in Lemma \ref{lemma3} of the supplementary article. 

In the second step, we show the linear representation in \eqref{CLT} holds. The proof of \eqref{CLT} relies on the convergence rate of $\widehat{\bm{\beta}}$ established in the first step and some additional random matrix nequalities  in Lemma \ref{lemma4}. 

In the last step, we show the leading term on the RHS of \eqref{CLT} is asymptotically normal, based on the martingale central limit theorem. The completes the proof of Theorem \ref{thm1}.

\subsection{A sketch for the proof of Theorem \ref{thm2}}\label{secsketchproofthm2}
Similar to \eqref{fakeCLT}, for each $1\le k\le K$, we have
\begin{eqnarray}\label{TrueCLT}
&&\frac{\widehat{V}_{\mathcal{I}_{k+1}}(\widehat{\pi}_{\bar{\mathcal{I}}_k};\mathbb{G})-V(\widehat{\pi}_{{\bar{\mathcal{I}}_k}};\mathbb{G})}{(nT/K)^{-1/2}\widehat{\sigma}_{\mathcal{I}_{k+1}}(\widehat{\pi}_{\bar{\mathcal{I}}_k};\mathbb{G})}\\ \nonumber
&=&\frac{(nT/K)^{-\frac{1}{2}}}{{\sigma}(\widehat{\pi}_{\bar{\mathcal{I}}_k};\mathbb{G})} \sum_{(i,t)\in \mathcal{I}_{k+1}} \left\{ \int_{x\in \mathbb{X}} \bm{U}_{\widehat{\pi}_{\bar{\mathcal{I}}_k}}(x)\mathbb{G}(dx) \right\}^{\top} \bm{\Sigma}_{\widehat{\pi}_{\bar{\mathcal{I}}_k}}^{-1} \bm{\xi}_{i,t}\varepsilon_{i,t}(\widehat{\pi}_{\bar{\mathcal{I}}_k})+R_k^{(1)},
\end{eqnarray}
where $R_k^{(1)}$ denotes the remainder term and
\begin{eqnarray*}
	\varepsilon_{i,t}(\widehat{\pi}_{\bar{\mathcal{I}}_k})=Y_{i,t}+\gamma \sum_{a\in \mathcal{A}}Q(\widehat{\pi}_{\bar{\mathcal{I}}_k};X_{i,t+1},a)\widehat{\pi}_{\bar{\mathcal{I}}_k}(a|X_{i,t+1})-Q(\widehat{\pi}_{\bar{\mathcal{I}}_k};X_{i,t},A_{i,t}).
\end{eqnarray*}
Since \eqref{condhold} is satisfied, we have $\Mean \{\varepsilon_{i,t}(\widehat{\pi}_{\bar{\mathcal{I}}_k})|O_k\}=0$. Conditional on the data in $O_k$, $\widehat{\pi}_{\bar{\mathcal{I}}_k}$ is a deterministic rule. The RHS of \eqref{TrueCLT} is thus equivalent to a mean-zero martingale. 

When (A5) is satisfied, we have
\begin{eqnarray}\label{TrueCLT1}
\frac{\widehat{V}_{\mathcal{I}_{k+1}}(\widehat{\pi}_{\bar{\mathcal{I}}_k};\mathbb{G})-V(\widehat{\pi}_{\bar{\mathcal{I}}_k};\mathbb{G})}{(nT/K)^{-1/2}\widehat{\sigma}_{\mathcal{I}_{k+1}}(\widehat{\pi}_{\bar{\mathcal{I}}_k};\mathbb{G})}=\frac{\widehat{V}_{\mathcal{I}_{k+1}}(\widehat{\pi}_{\bar{\mathcal{I}}_k};\mathbb{G})-V(\pi^*;\mathbb{G})}{(nT/K)^{-1/2}\widehat{\sigma}_{\mathcal{I}_{k+1}}(\widehat{\pi}_{\bar{\mathcal{I}}_k};\mathbb{G})}+R_k^{(2)},
\end{eqnarray}
for some remainder term $R_k^{(2)}$. Suppose $R_k^{(1)}$ and $R_k^{(2)}$ satisfy certain convergence rates. Combining \eqref{TrueCLT1} together with \eqref{TrueCLT} yields,
\begin{eqnarray}\label{TrueCLT2}
&&\frac{\widetilde{V}(\mathbb{G})-V(\pi^*;\mathbb{G})}{\{nT(K-1)/K\}^{-1/2}\widetilde{\sigma}(\mathbb{G})}\\ \nonumber
&=& \sqrt{\frac{nT(K-1)}{K}} \sum_{k=1}^{K-1}\sum_{(i,t)\in \mathcal{I}_{k+1}} \frac{1}{{\sigma}(\widehat{\pi}_{\bar{\mathcal{I}}_k};\mathbb{G})} \left\{ \int_{x\in \mathbb{X}} \bm{U}_{\widehat{\pi}_{\bar{\mathcal{I}}_k}}(x)\mathbb{G}(dx) \right\}^{\top} \bm{\Sigma}_{\widehat{\pi}_{\bar{\mathcal{I}}_k}}^{-1} \bm{\xi}_{i,t}\varepsilon_{i,t}(\widehat{\pi}_{\bar{\mathcal{I}}_k})+o_p(1),
\end{eqnarray}
due to our use of the inverse weighting trick. Theorem \ref{thm2} thus follows from the martingale central limit theorem.

\subsection{A sketch for the proofs of Theorems \ref{thm3} and \ref{thm4}}\label{secsketchproofthm34}
Proofs of Theorems \ref{thm3} and \ref{thm4} are divided into two steps. In the first step, we decompose the value difference $V(\pi^{\tiny{opt}};\mathbb{G})-V(\widehat{\pi}_{\mathcal{I}};\mathbb{G})$ into the sum of an infinite series and provide upper bounds for all the terms in the series. In the second step, we use the margin-type condition A5 to further characterize these upper bounds. We only present the first step in this section.

For $j=0,1,2,\dots$, define a time-dependent policy $\widehat{\pi}_{\mathcal{I}}^{(j)}$ that executes $\widehat{\pi}_{\mathcal{I}}$ at the first $j$ time points and then follows $\pi^{\tiny{opt}}$. By definition, we have $\pi^{\tiny{opt}}=\widehat{\pi}_{\mathcal{I}}^{(0)}$ and $\widehat{\pi}_{\mathcal{I}}=\widehat{\pi}_{\mathcal{I}}^{(\infty)}$. Notice that
\begin{eqnarray}\label{Vpifirststep}
V(\pi^{\tiny{opt}};\mathbb{G})-V(\widehat{\pi}_{\mathcal{I}}^{(1)};\mathbb{G})=\int_x \sum_{a\in \mathcal{A}} Q^{\tiny{opt}}(x,a) \{\pi^{\tiny{opt}}(a|x)-\widehat{\pi}_{\mathcal{I}}(a|x)\}\mathbb{G}(dx).
\end{eqnarray}
Moreover, for any $j\ge 1$, 
\begin{eqnarray*}
	V(\widehat{\pi}_{\mathcal{I}}^{(j)};\mathbb{G})-V(\widehat{\pi}_{\mathcal{I}}^{(j+1)};\mathbb{G})=\int_x\sum_{t\ge 0} \gamma^{t} \{\Mean^{\widehat{\pi}_{\mathcal{I}}^{(j)}} (Y_{0,t}|X_{0,0}=x)-\Mean^{\widehat{\pi}_{\mathcal{I}}^{(j+1)}} (Y_{0,t}|X_{0,0}=x)\} \mathbb{G}(dx)\\
	=\int_x\sum_{t\ge j} \gamma^{t} \{\Mean^{\widehat{\pi}_{\mathcal{I}}^{(j)}} (Y_{0,t}|X_{0,0}=x)-\Mean^{\widehat{\pi}_{\mathcal{I}}^{(j+1)}} (Y_{0,t}|X_{0,0}=x)\} \mathbb{G}(dx). 
\end{eqnarray*}
Let $q_X^{(j)}(\cdot|x)$ be the density function of $X_{0,j}$ conditional on $X_{0,0}=x$, following the estimated policy $\widehat{\pi}_{\mathcal{I}}$ at the first $j$ time points, we have
\begin{eqnarray*}
	\sum_{t\ge j}\gamma^{t} \Mean^{\widehat{\pi}_{\mathcal{I}}^{(j)}} (Y_{0,t}|X_{0,j})=\gamma^j \sum_{a\in \mathcal{A}} Q^{\tiny{opt}}(X_{0,j},a)\pi^{\tiny{opt}}(a|X_{0,j}),\\
	\sum_{t\ge j}\gamma^{t} \Mean^{\widehat{\pi}_{\mathcal{I}}^{(j+1)}} (Y_{0,t}|X_{0,j})=\gamma^j \sum_{a\in \mathcal{A}} Q^{\tiny{opt}}(X_{0,j},a)\widehat{\pi}_{\mathcal{I}}(a|X_{0,j}).
\end{eqnarray*}
It follows that
\begin{eqnarray*}
	V(\widehat{\pi}_{\mathcal{I}}^{(j)};\mathbb{G})-V(\widehat{\pi}_{\mathcal{I}}^{(j+1)};\mathbb{G})=\gamma^j \int_{x,x'\in \mathbb{X}} \sum_{a\in \mathcal{A}} Q^{\tiny{opt}}(x',a)\{\pi^{\tiny{opt}}(a|x')-\widehat{\pi}_{\mathcal{I}}(a|x') \}  q_X^{(j)}(x'|x)dx' \mathbb{G}(dx).
\end{eqnarray*}
By A1, we have $\sup_{x,x',a} q(x'|x,a)\le c$. Under the Markov assumption,
\begin{eqnarray*}
	q_X^{(j)}(x'|x)=\int_y \sum_{a\in \mathcal{A}} q(x'|y,a)\widehat{\pi}_{\mathcal{I}}(a|y)q_X^{(j-1)}(y|x)dy\le c \int_y q_X^{(j-1)}(y|x)dy=c. 
\end{eqnarray*}
In addition, $\sum_{a\in \mathcal{A}} Q^{\tiny{opt}}(x',a)\{\pi^{\tiny{opt}}(a|x')-\widehat{\pi}_{\mathcal{I}}(a|x') \}\ge 0$ for any $x'$, by the definition of $\pi^{\tiny{opt}}$. Therefore, we obtain
\begin{eqnarray}\label{Vpijthstep}
V(\widehat{\pi}_{\mathcal{I}}^{(j)};\mathbb{G})-V(\widehat{\pi}_{\mathcal{I}}^{(j+1)};\mathbb{G})\le c \gamma^j \int_x \sum_{a\in \mathcal{A}} Q^{\tiny{opt}}(x,a)\{\pi^{\tiny{opt}}(a|x)-\widehat{\pi}_{\mathcal{I}}(a|x) \} dx.
\end{eqnarray}
By A1, the reward function $r(\cdot,\cdot)$ is uniformly bounded. This further implies that $Q^{\tiny{opt}}(\cdot,\cdot)$ is uniformly bounded. Therefore, $\sum_{j\ge t} V(\widehat{\pi}_{\mathcal{I}}^{(j)};\mathbb{G})-V(\widehat{\pi}_{\mathcal{I}}^{(j+1)};\mathbb{G})\to 0$ as $t\to \infty$. It follows from \eqref{Vpifirststep} and \eqref{Vpijthstep} that
\begin{eqnarray}\nonumber
&&V(\pi^{\tiny{opt}};\mathbb{G})-V(\widehat{\pi}_{\mathcal{I}};\mathbb{G})=\sum_{j=0}^{+\infty} \{V(\widehat{\pi}_{\mathcal{I}}^{(j)};\mathbb{G})-V(\widehat{\pi}_{\mathcal{I}}^{(j+1)};\mathbb{G})\}\\ \label{boundG}
&\le& \int_x \sum_{a\in \mathcal{A}} Q^{\tiny{opt}}(x,a) \{\pi^{\tiny{opt}}(a|x)-\widehat{\pi}_{\mathcal{I}}(a|x)\}\mathbb{G}(dx)\\ \label{boundlambda}
&+& \frac{c\gamma}{1-\gamma} \int_x \sum_{a\in \mathcal{A}} Q^{\tiny{opt}}(x,a) \{\pi^{\tiny{opt}}(a|x)-\widehat{\pi}_{\mathcal{I}}(a|x)\}dx. 
\end{eqnarray}
Let $\lambda$ denote the Lebesgue measure on $\mathbb{R}^d$. In Sections \ref{secproofthm3} and \ref{secproofthm4}, we use A5 to further bound \eqref{boundG} and \eqref{boundlambda}. 
\subsection{Proof of Lemma \ref{lemma1}}
Since $\mathbb{X}$ is compact, Condition (A1) implies that $\sup_{x\in \mathbb{X},a\in \mathcal{A}}|r(x,a)|\le R$ for some $0<R<+\infty$. Under CMIA, we have
\begin{eqnarray*}
	Q(\pi;x,a)=\sum_{t\ge 0}\gamma^{t} \Mean^{\pi}(Y_{0,t},X_{0,0}=x,A_{0,0}=a)=\sum_{t\ge 0} \gamma^{t} \Mean^{\pi} \{r(X_{0,t},A_{0,t})|X_{0,0}=x,A_{0,0}=a\}. 
\end{eqnarray*}
As a result, we obtain 
\begin{eqnarray}\label{Qbound}
\sup_{\pi,x,a} |Q(\pi;x,a)|\le \frac{R}{1-\gamma}.
\end{eqnarray}
By the Bellman equation, we obtain
\begin{eqnarray*}
	Q(\pi;x,a)=r(x,a)+\gamma \int_{x'} \sum_{a'\in \mathcal{A}} Q(\pi;x',a')\pi(a'|x') q(x'|x,a)dx'.
\end{eqnarray*}
Since $r(\cdot,a)$ is $p$-smooth for any $a\in \mathcal{A}$, it suffices to show
\begin{eqnarray*}
	T(\pi;x,a)=\int_{x'} \sum_{a'\in \mathcal{A}} Q(\pi;x',a') \pi(a'|x') q(x'|x,a)dx',
\end{eqnarray*}
is $p$-smooth for any $a\in \mathcal{A}$ and any policy $\pi$. 

For any function $h(\cdot)$ defined on $\mathbb{X}$, let $\partial_j h(x)$ denote the partial derivative $\partial h(x)/\partial x_j$. Without loss of generality, suppose $p>1$ such that $\partial_j p(x'|x,a)$ exists for any $j$. In the following, we show $\partial_j T(\pi;x,a)$ exists for any $j$. Let 
\begin{eqnarray*}
	e_j=(\underbrace{0,\dots,0}_{j-1},1,\underbrace{0,\dots,0}_{d-j})^{\top},\,\,\,\,\forall j=1,\dots,d.
\end{eqnarray*}
For any $\delta\in \mathbb{R}$, consider the limit
\begin{eqnarray*}
	\hbox{Re}(j,\delta)=\frac{T(\pi;x+e_j \delta,a)-T(\pi;x,a)}{\delta}-\int_{x'} \sum_{a'\in \mathcal{A}} Q(\pi;x',a')\pi(a'|x')\partial_j q(x'|x,a)dx',
\end{eqnarray*}
as $j\to \infty$. By the mean value theorem, we have
\begin{eqnarray*}
	|\hbox{Re}(j,\delta)|\le \int_{x'}\sum_{a'\in \mathcal{A}} |Q(\pi;x',a')|\pi(a'|x') \left| \frac{q(x'|x+e_j \delta,a)-q(x'|x,a)}{\delta}-\partial_jq(x'|x,a) \right|dx'\\
	\le \int_{x'}\sum_{a'\in \mathcal{A}} |Q(\pi;x',a')|\pi(a'|x') |\partial_j q(x'|x+e_j \theta_x \delta,a)-\partial_j q(x'|x,a)|dx',
\end{eqnarray*}
where $0\le \theta_x\le 1$ for all $x$. When $1<p\le 2$, we have $\floor{p}=1$. It follows from Condition A1 that 
\begin{eqnarray*}
	|\partial_j q(x'|x+e_j \theta_x \delta,a)-\partial_j q(x'|x,a)|\le c\delta^{p-1}. 
\end{eqnarray*}
When $p>2$, $|\partial_j \partial_j q(x'|x,a)|$ exists and is bounded by $c$ for any $x'$, $x$ and $a$. It follows from the mean value theorem that
\begin{eqnarray*}
	|\partial_j q(x'|x+e_j \theta_x \delta,a)-\partial_j q(x'|x,a)|\le c\delta. 
\end{eqnarray*}
In either case, we have that
\begin{eqnarray*}
	|\hbox{Re}(j,\delta)|\le c\max(\delta,\delta^{p-1})\int_{x'}\sum_{a'\in \mathcal{A}} |Q(\pi;x',a')|\pi(a'|x') dx'.
\end{eqnarray*}
By \eqref{Qbound} and that $\mathbb{X}$ is compact, we obtain $\hbox{Re}(j,\delta)\to 0$ as $\delta\to 0$. This implies that $\partial_j T(\pi;x,a)$ exists for any $x\in\mathbb{X}$, $a\in \mathcal{A}$ and equals
\begin{eqnarray*}
	\int_{x'} \sum_{a'\in \mathcal{A}} Q(\pi;x',a')\pi(a'|x')\partial_j q(x'|x,a)dx'.
\end{eqnarray*}
In addition, it follows from A1 and \eqref{Qbound} that
\begin{eqnarray*}
	|\partial_j T(\pi;x,a)|\le \frac{R c}{1-\gamma} \lambda(\mathbb{X}),\,\,\,\,\,\,\,\,\forall j,x,a,\pi,
\end{eqnarray*}
where $\lambda(\cdot)$ denotes the Lebesgue measure. Using the same arguments, we can show for any $d$-tuple $\alpha=(\alpha_1,\dots,\alpha_d)^{\top}$ of nonnegative integers that satisfies $\|\alpha\|_1\le \floor{p}$,
\begin{eqnarray}\label{lemma1eq1}
D^{\alpha} T(\pi;\cdot,a)=\int_{x'} \sum_{a'\in \mathcal{A}} Q(\pi;x',a')\pi(a'|x')D^{\alpha} q(x'|\cdot,a)dx',
\end{eqnarray}
and
\begin{eqnarray}\label{lemma1eq2}
\sup_{\pi}\sup_{x\in \mathbb{X},a\in \mathcal{A}}|D^{\alpha} T(\pi;x,a)|\le \frac{R c}{1-\gamma} \lambda(\mathbb{X}).
\end{eqnarray}
Moreover, by A1, \eqref{Qbound} and \eqref{lemma1eq1}, we have for any $d$-tuple $\alpha$ with $\|\alpha\|_1=\floor{p}$ that
\begin{eqnarray*}
	&&\|D^{\alpha} T(\pi;x,a)-D^{\alpha} T(\pi;y,a)\|_2\\
	&\le&  \int_{x'} \sum_{a'\in \mathcal{A}} |Q(\pi;x',a')|\pi(a'|x')|D^{\alpha} q(x'|x,a)-D^{\alpha} q(x'|y,a)|dx'\le \frac{R c}{1-\gamma} \lambda(\mathbb{X}) \|x-y\|_2^{p-\floor{p}}. 
\end{eqnarray*}
This together with \eqref{lemma1eq1} implies that $T(\pi;\cdot,a)\in \Lambda(p, Rc\lambda(\mathbb{X})(1-\gamma)^{-1})$ for any $\pi$ and $a$. The proof is thus completed. 

\subsection{Proof of Theorem \ref{thm1}}\label{secproofthm1}
We introduce the following lemmas before proving Theorem \ref{thm1}. In the proof of Theorem \ref{thm1} and Lemma \ref{lemma2}-\ref{lemma5}, we will omit the subscript $\pi$ in $\bm{U}_{\pi}(\cdot)$, $\bm{u}_{\pi}$, $\bm{\Sigma}_{\pi}$, $\widehat{\bm{\Sigma}}_{\pi}$, $\widehat{\bm{\beta}}_{\pi},\bm{\beta}_{\pi}^*$, $\omega_{\pi}$, etc, for brevity. 

\begin{lemma}\label{lemma2}
	There exists some constant $c^*\ge 1$ such that
	\begin{eqnarray}\label{lemma2eq1}
	(c^*)^{-1}\le \lambda_{\min}\left\{\int_{x\in \mathbb{X}}  \Phi_L(x)\Phi^{\top}_L(x)dx\right\}\le \lambda_{\max}\left\{\int_{x\in \mathbb{X}} \Phi_L(x)\Phi^{\top}_L(x)dx\right\} \le c^*,
	\end{eqnarray}
	and $\sup_{x\in \mathbb{X}}\|\Phi_L(x)\|_2\le c^*\sqrt{L}$. 
\end{lemma}

\begin{lemma}\label{lemma3}
	Suppose the conditions in Theorem \ref{thm1} hold. We have as either $n\to \infty$ or $T\to \infty$ that $\|\bm{\Sigma}^{-1}\|_2\le 3\bar{c}^{-1}$, $\|\bm{\Sigma}\|_2=O(1)$, $\|\widehat{\bm{\Sigma}}-\bm{\Sigma}\|_2=O_p\{L^{1/2} (nT)^{-1/2}\log (nT)\}$, $\|\widehat{\bm{\Sigma}}^{-1}-\bm{\Sigma}^{-1}\|_2=O_p\{L^{1/2} (nT)^{-1/2}\log (nT)\}$ and $\|\widehat{\bm{\Sigma}}^{-1}\|\le 6\bar{c}^{-1}$ wpa1. 	
\end{lemma}

\begin{lemma}\label{lemma4}
	Suppose the conditions in Theorem \ref{thm1} hold. We have as either $n\to \infty$ or $T\to \infty$ that $\lambda_{\max}(T^{-1}\sum_{t=0}^{T-1}\Mean \bm{\xi}_{0,t}\bm{\xi}_{0,t}^{\top})=O(1)$, $\lambda_{\max}\{(nT)^{-1} \sum_{i=1}^n\sum_{t=0}^{T-1} \bm{\xi}_{i,t}\bm{\xi}_{i,t}^{\top}\}=O_p(1)$, $\lambda_{\min}(T^{-1}\sum_{t=0}^{T-1}\Mean \bm{\xi}_{0,t}\bm{\xi}_{0,t}^{\top})\ge \bar{c}/2$ and $\lambda_{\min}\{(nT)^{-1} \sum_{i=1}^n \sum_{t=0}^{T-1} \bm{\xi}_{i,t}\bm{\xi}_{i,t}^{\top}\}\ge \bar{c}/3$ wpa1.
\end{lemma}

\begin{lemma}\label{lemma5}
	$\|\int_x \bm{U}(x)\mathbb{G}(dx)\|_2\ge m^{-1/2} \|\int_x \Phi_L(x)\mathbb{G}(dx)\|_2$. 
\end{lemma}

\textit{Step 1. }
Since $L^{2p/d}\gg nT\{1+ \|\int_x \Phi_L(x)\mathbb{G}(dx)\|_2^{-2}\}$, it follows from Lemma \ref{lemma5} that $L^{2p/d}\gg  nT\{1+ \|\int_x \bm{U}(x)\mathbb{G}(dx)\|_2^{-2}\}$. 
By Lemma \ref{lemma1}, there exist a set of vector $\{\beta_a^*\}$ that satisfy \citep[see Section 2.2 of][for details]{Huang1998}
\begin{eqnarray}\label{biasbound}
\sup_{x\in \mathbb{X},a\in \mathcal{A}} |Q(\pi;x,a)-\Phi_L^{\top}(x) \beta_a^*|\le CL^{-p/d},
\end{eqnarray}
for some constant $C>0$. 
Let $\bm{\beta}^*=(\beta_1^{*T},\dots,\beta_m^{*T})^{\top}$, and
\begin{eqnarray*}
	r_{i,t}=\gamma \sum_{a\in \mathcal{A}}\{\Phi_L^{\top}(X_{i,t+1})\beta_{a}^*-Q(\pi;X_{i,t+1},a)\}\pi(a|X_{i,t+1})-\{\Phi_L^{\top}(X_{i,t}) \beta_{A_{i,t}}^*-Q(\pi;X_{i,t},A_{i,t})\}.
\end{eqnarray*}
The condition $\prob(\max_{0\le t\le T-1} |Y_{i,t}|\le c_0)=1$ implies that $|Y_{i,t}|\le c_0,\forall i,t$, almost surely. By Lemma \ref{lemma1} and the definition of $p$-smooth functions, we obtain that $|Q(\pi;x,a)|\le c'$ for any $\pi,x,a$. It follows that
\begin{eqnarray}\label{maxresponse}
\max_{0\le t\le T-1,0\le i\le n} |\varepsilon_{i,t}|\le c_0+(\gamma+1) c'\le c_0+2c',
\end{eqnarray}
almost surely. In addition, it follows from \eqref{biasbound} that
\begin{eqnarray}\label{maxresidual}
\max_{0\le t\le T-1,1\le i\le n} |r_{i,t}|\le 2\sup_{x\in \mathbb{X},a\in \mathcal{A}} |Q(\pi;x,a)-\Phi_L^{\top}(x) \beta_a^*|\le 2CL^{-p/d}.
\end{eqnarray}
By definition, we have
\begin{eqnarray*}
	&&\widehat{\bm{\beta}}-\bm{\beta}^*=\widehat{\bm{\Sigma}}^{-1} \left[ \frac{1}{nT} \sum_{i=1}^n \sum_{t=0}^{T-1} \bm{\xi}_{i,t} \{Y_{i,t}- (\bm{\xi}_{i,t}-\gamma \bm{U}_{i,t+1})^{\top} \bm{\beta}^*\} \right]\\
	&=&\widehat{\bm{\Sigma}}^{-1} \left[ \frac{1}{nT} \sum_{i=1}^n \sum_{t=0}^{T-1} \bm{\xi}_{i,t} \left\{Y_{i,t}-\Phi_L^{\top}(X_{i,t})\beta_{A_{i,t}}^*+\gamma \sum_{a\in \mathcal{A}} \Phi_L^{\top}(X_{i,t+1})\beta_a^*\pi(a|X_{i,t+1}) \right\} \right]\\
	&=&\widehat{\bm{\Sigma}}^{-1} \left\{\frac{1}{nT}\sum_{i=1}^n \sum_{t=0}^{T-1} \bm{\xi}_{i,t}(\varepsilon_{i,t}-r_{i,t}) \right\}=\underbrace{\bm{\Sigma}^{-1} \left( \frac{1}{nT}\sum_{i=1}^n \sum_{t=0}^{T-1} \bm{\xi}_{i,t}\varepsilon_{i,t} \right)}_{\zeta_1}\\
	&+&\underbrace{(\widehat{\bm{\Sigma}}^{-1}-\bm{\Sigma}^{-1}) \left( \frac{1}{nT}\sum_{i=1}^n \sum_{t=0}^{T-1} \bm{\xi}_{i,t}\varepsilon_{i,t} \right)}_{\zeta_2}-\underbrace{\widehat{\bm{\Sigma}}^{-1} \left( \frac{1}{nT}\sum_{i=1}^n \sum_{t=0}^{T-1} \bm{\xi}_{i,t}r_{i,t} \right)}_{\zeta_3}.
\end{eqnarray*}
In the following, we show $\zeta_2=O_p\{L (nT)^{-1}\log (nT)\}$ and $\zeta_3=O_p(L^{-p/d})$ as either $n\to \infty$, or $T\to \infty$. 

\noindent \textit{Error bound for $\|\zeta_2\|_2$: }Let $\mathcal{F}_{i,t}$ denote the sub-dataset $\{X_{i,t},A_{i,t}\}\cup\{(Y_{i,j},A_{i,j},X_{i,j})\}_{1\le j<t}$. By the Bellman equation in \eqref{QBell2}, MA and CMIA, we have 
\begin{eqnarray*}
	\Mean (\varepsilon_{i,t}| \mathcal{F}_{i,t})=\Mean (\varepsilon_{i,t}|X_{i,t},A_{i,t})=0.
\end{eqnarray*}
Notice that $\bm{\xi}_{i,t}$ is a function of $X_{i,t}$ and $A_{i,t}$ only, we have for any $0\le t_1<t_2\le T-1$ that
\begin{eqnarray*}
	\Mean \varepsilon_{i,t_1}\varepsilon_{i,t_2}\bm{\xi}_{i,t_1}^{\top} \bm{\xi}_{i,t_2}=\Mean \{\varepsilon_{i,t_1} \bm{\xi}_{i,t_1}^{\top} \bm{\xi}_{i,t_2} \Mean (\varepsilon_{i,t_2}|\mathcal{F}_{i,t_2})\}=0.
\end{eqnarray*}
By the independence assumption, we have for any $0\le t_1<t_2\le T-1$ and $1\le i_1<i_2\le n$ that $\Mean \varepsilon_{i_1,t_1}\varepsilon_{i_2,t_2}\bm{\xi}_{i_1,t_1}^{\top} \bm{\xi}_{i_2,t_2}=0$. It follows that
\begin{eqnarray*}
	\Mean \left\|\sum_{i=1}^n \sum_{t=0}^{T-1} \bm{\xi}_{i,t}\varepsilon_{i,t}\right\|_2^2=\sum_{i=1}^n \sum_{t=0}^{T-1} \Mean \varepsilon_{i,t}^2 \bm{\xi}_{i,t}^{\top} \bm{\xi}_{i,t}=n\sum_{t=0}^{T-1} \Mean \varepsilon_{0,t}^2 \bm{\xi}_{0,t}^{\top} \bm{\xi}_{0,t}.
\end{eqnarray*}
By \eqref{maxresponse} and Lemma \ref{lemma2}, we obtain
\begin{eqnarray*}
	\Mean \left\|\sum_{i=1}^n \sum_{t=0}^{T-1} \bm{\xi}_{i,t}\varepsilon_{i,t}\right\|_2^2\le (c_0+2c')^2 n\sum_{t=0}^{T-1} \Mean\bm{\xi}_{0,t}^{\top} \bm{\xi}_{0,t}\le (c_0+2c')^2 nT\sup_{x\in \mathbb{X}} \|\Phi_L(x)\|_2^2 \preceq nTL.
\end{eqnarray*}
By Markov's inequality, we obtain $(nT)^{-1} \sum_{i=1}^n \sum_{t=0}^{T-1} \bm{\xi}_{i,t}\varepsilon_{i,t}=O_p\{\sqrt{L/(nT)}\}$. Combining this together with Lemma \ref{lemma3} yields that $\zeta_2=O_p\{\sqrt{L/nT}\log (nT)\}O_p\{\sqrt{L/(nT)}\}=O_p\{L(nT)^{-1}\log (nT)\}$. 

\noindent \textit{Error bound for $\|\zeta_3\|_2$: }For any $\bm{a}\in \mathbb{R}^{mL}$, it follows from \eqref{maxresidual} and Cauchy-Schwarz inequality that
\begin{eqnarray*}
	\left|\bm{a}^{\top} \left( \frac{1}{nT} \sum_{i=1}^n \sum_{t=0}^{T-1}\bm{\xi}_{i,t}r_{i,t} \right)\right|\le \frac{1}{nT}\sum_{i=1}^{n}\sum_{t=0}^{T-1} |\bm{a}^{\top} \bm{\xi}_{i,t}| |r_{i,t}|\le \max_{i,t}|r_{i,t}| \left( \frac{1}{nT}\sum_{i=1}^{n}\sum_{t=0}^{T-1} |\bm{a}^{\top} \bm{\xi}_{i,t}| \right)\\
	\le 2CL^{-p/d} \left( \frac{1}{nT}\sum_{i=1}^{n}\sum_{t=0}^{T-1} |\bm{a}^{\top} \bm{\xi}_{i,t}| \right)\le 2CL^{-p/d} \left( \frac{1}{nT}\sum_{i=1}^{n}\sum_{t=0}^{T-1} \bm{a}^{\top} \bm{\xi}_{i,t}\bm{\xi}^{\top}_{i,t} \bm{a} \right)^{1/2}.
\end{eqnarray*}
Therefore, we obtain
\begin{eqnarray*}
	\left\| \frac{1}{nT} \sum_{i=1}^n \sum_{t=0}^{T-1}\bm{\xi}_{i,t}r_{i,t} \right\|_2\le 2CL^{-p/d} \lambda_{\max}^{1/2}\left( \frac{1}{nT}\sum_{i=1}^{n}\sum_{t=0}^{T-1} \bm{\xi}_{i,t}\bm{\xi}^{\top}_{i,t} \right).
\end{eqnarray*}
By Lemma \ref{lemma4}, we obtain 
\begin{eqnarray}\label{proofthm1eq0}
\left\| \frac{1}{nT} \sum_{i=1}^n \sum_{t=0}^{T-1}\bm{\xi}_{i,t}r_{i,t}\right\|_2=O_p(L^{-p/d}).
\end{eqnarray}
By Lemma \ref{lemma3}, we have $\|\widehat{\bm{\Sigma}}^{-1}\|_2=O_p(1)$. Combining this together with \eqref{proofthm1eq0} yields that $\zeta_3=O_p(L^{-p/d})$. 

To summarize, we have shown
\begin{eqnarray}\label{betaerrorbound0}
\widehat{\bm{\beta}}-\bm{\beta}^*=\zeta_1+O_p(L^{-p/d})+O_p\{L(nT)^{-1}\log(nT)\}.
\end{eqnarray}  
This completes the first step of the proof.

\textit{Step 2:} Using similar arguments in bounding $\|\zeta_2\|_2$ in Step 1, we can show that $\|\zeta_1\|_2=O_p\{L^{1/2}(nT)^{-1/2}\}$ and thus
\begin{eqnarray}\label{betaerrorbound1}
\|\widehat{\bm{\beta}}-\bm{\beta}^*\|_2=O_p(L^{-p/d})+O_p\{L^{1/2}(nT)^{-1/2}\},
\end{eqnarray}
under the condition that $L\ll \sqrt{nT}/\log (nT)$. 
Notice that $\widehat{V}(\pi;\mathbb{G})$ can be presented as $\{\int_{x} \bm{U}(x)\mathbb{G}(dx)\}^{\top} \widehat{\bm{\beta}}$. As a result, 
\begin{eqnarray}\label{proofthm1eq1}
\left|\widehat{V}(\pi;\mathbb{G})- \left\{\int_{x} \bm{U}(x)\mathbb{G}(dx)\right\}^{\top} (\bm{\beta}^*+\zeta_1) \right|\le \left\|\int_{x} \bm{U}(x)\mathbb{G}(dx)\right\|_2 \|\widehat{\bm{\beta}}-\bm{\beta}^*-\zeta_1\|_2.
\end{eqnarray}
By \eqref{biasbound}, we have
\begin{eqnarray*}
	|\bm{U}^{\top}(x) \bm{\beta}^*-V(\pi;x)|=\left|\bm{U}^{\top}(x) \bm{\beta}^*-\sum_{a\in \mathcal{A}}Q(\pi;x,a)\pi(a|x) \right|\\
	\le \sum_{a\in \mathcal{A}} |Q(\pi;x,a)-\Phi_L^{\top}(x)\beta_a^*|\pi(a|x)\le CL^{-p/d},
\end{eqnarray*}
and hence
\begin{eqnarray*}
	\left|\left\{\int_x \bm{U}(x)\mathbb{G}(dx)\right\}^{\top} \bm{\beta}^*-\int_x V(\pi;x)\mathbb{G}(dx)\right|\le CL^{-p/d}.
\end{eqnarray*}
This together with \eqref{proofthm1eq1} yields that
\begin{eqnarray}\label{proofthm1eq2}
|\widehat{V}(\pi;\mathbb{G})-V(\pi;\mathbb{G})-\bm{U}^{\top}(x) \zeta_1|\le \left\|\int_x\bm{U}(x)\mathbb{G}(dx)\right\|_2 \|\widehat{\bm{\beta}}-\bm{\beta}^*-\zeta_1\|_2+CL^{-p/d}.
\end{eqnarray}
Let $\bm{\Omega}=T^{-1}\sum_{t=0}^{T-1} \Mean \omega(X_{0,t},A_{0,t})\bm{\xi}_{0,t}\bm{\xi}_{0,t}^{\top}$ and
\begin{eqnarray*}
	\sigma^2(\pi;\mathbb{G})=\left\{\int_x\bm{U}(x)\mathbb{G}(dx)\right\}^{\top} \bm{\Sigma}^{-1} \bm{\Omega} (\bm{\Sigma}^{\top})^{-1} \left\{\int_x\bm{U}(x)\mathbb{G}(dx)\right\}.
\end{eqnarray*}  
Since $\inf_{x,a} \omega(x,a)\ge c_0^{-1}$, it follows from Lemma \ref{lemma4} that
\begin{eqnarray*}
	\lambda_{\min}(\bm{\Omega})\ge c_0^{-1} \lambda_{\min} \left( \frac{1}{T}\sum_{t=0}^{T-1}\Mean \bm{\xi}_{0,t}\bm{\xi}_{0,t}^{\top} \right)\ge 3^{-1} c_0^{-1} \bar{c}. 
\end{eqnarray*}
Hence, we obtain
\begin{eqnarray}\label{lowerbound1}
\sigma^2(\pi;\mathbb{G})\ge \frac{\bar{c}}{3c_0} \left\{\int_x\bm{U}(x)\mathbb{G}(dx)\right\}^{\top} \bm{\Sigma}^{-1} (\bm{\Sigma}^{\top})^{-1} \left\{\int_x\bm{U}(x)\mathbb{G}(dx)\right\}.
\end{eqnarray}
By Lemma \ref{lemma3}, we have $\|\bm{\Sigma}\|_2=O(1)$, or equivalently, $\lambda_{\max}(\bm{\Sigma}^{\top} \bm{\Sigma})=O(1)$. This implies that $\lambda_{\min}\{\bm{\Sigma}^{-1} (\bm{\Sigma}^{\top})^{-1}\}\ge \bar{C}$ for some constant $\bar{C}>0$ and hence 
\begin{eqnarray}\label{lowerbound1.5}
\sigma^2(\pi;\mathbb{G})\ge (3c_0)^{-1}\bar{c}\bar{C} \left\|\int_x\bm{U}(x)\mathbb{G}(dx)\right\|_2^2,
\end{eqnarray}
by \eqref{lowerbound1}. Combining \eqref{lowerbound1.5} together with \eqref{proofthm1eq2} yields that
\begin{eqnarray*}
	\frac{1}{\sigma(\pi;\mathbb{G})}\left|\widehat{V}(\pi;\mathbb{G})-V(\pi;\mathbb{G})-\left\{\int_x \bm{U}(x)\mathbb{G}(dx)\right\}^{\top} \zeta_1\right|\le \frac{\sqrt{3c_0}}{\sqrt{\bar{c}\bar{C}}}\|\widehat{\bm{\beta}}-\bm{\beta}^*-\zeta_1\|_2\\
	+\frac{\sqrt{3c_0} C L^{-p/d}}{\sqrt{\bar{c}\bar{C}}\|\int_x\bm{U}(x)\mathbb{G}(dx)\|_2}.
\end{eqnarray*}
By \eqref{betaerrorbound0} and that $L\ll \sqrt{nT}/\log (nT)$, $L^{2p/d}\gg nT\{1+\|\int_x\bm{U}(x) \mathbb{G}(dx)\|_2^{-2}\}$, we obtain
\begin{eqnarray}\label{proofthm1eq3}
\frac{\sqrt{nT}\{\widehat{V}(\pi;\mathbb{G})-V(\pi;\mathbb{G})\} }{\sigma(\pi;\mathbb{G})}=\frac{\sqrt{nT} \{\int_x \bm{U}(x)\mathbb{G}(dx)\}^{\top} \zeta_1 }{\sigma(\pi;x)}+o_p(1).
\end{eqnarray}
This completes the second step of the proof.

\textit{Step 3:} In the following, we show $\sqrt{nT}\sigma^{-1}(\pi;\mathbb{G})\{\int_x \bm{U}(x)\mathbb{G}(dx)\}^{\top}\zeta_1\stackrel{d}{\to}N(0,1)$. For any integer $1\le g\le nT$, let $i(g)$ and $t(g)$ be the quotient and the remainder of $g+T-1$ divided by $T$ that satisfy
\begin{eqnarray*}
	g=\{i(g)-1\}T+t(g)+1\,\,\,\,\hbox{and}\,\,\,\,0\le t(g)<T.
\end{eqnarray*}
Let $\mathcal{F}^{(0)}=\{X_{1,0},A_{1,0}\}$. Then we iteratively define $\{\mathcal{F}^{(g)}\}_{1\le g\le nT}$ as follows:
\begin{eqnarray*}
	&&\mathcal{F}^{(g)}=\mathcal{F}^{(g-1)}\cup \{Y_{i(g), t(g)},X_{i(g),t(g)+1},A_{i(g),t(g)+1}\},\,\,\,\,\hbox{if}\,\,\,\,t(g)< T-1,\\
	&&\mathcal{F}^{(g)}=\mathcal{F}^{(g-1)}\cup \{Y_{i(g),T-1},X_{i(g),T},X_{i(g)+1,0},A_{i(g)+1,0} \},\,\,\,\,\hbox{otherwise}. 
\end{eqnarray*}
Let $\bm{\xi}^{(g)}=\bm{\xi}_{i(g),t(g)}$ and $\varepsilon^{(g)}=\varepsilon_{i(g),t(g)}$. It follows that
\begin{eqnarray}\label{proofthm1eq4}
\sqrt{nT}\frac{\{\int_x \bm{U}(x)\mathbb{G}(dx)\}^{\top}\zeta_1}{\sigma(\pi;x)}=\sum_{g=1}^{nT} \frac{\{\int_x \bm{U}(x)\mathbb{G}(dx)\}^{\top} \bm{\Sigma}^{-1} \bm{\xi}^{(g)}\varepsilon^{(g)}}{\sqrt{nT} \sigma(\pi;x)}.
\end{eqnarray}
By MA, CMIA and the Bellman equation in \eqref{QBell2}, we obtain that $$\Mean \{\varepsilon^{(g)}|\mathcal{F}^{(g-1)}\}=\Mean \{\varepsilon^{(g)}|X_{i(g),t(g)},A_{i(g),t(g)}\}=0.$$ Hence, the RHS of \eqref{proofthm1eq4} forms a martingale with respect to the filtration $\{\sigma(\mathcal{F}^{(g)})\}_{g\ge 0}$, where $\sigma(\mathcal{F}^{(g)})$ stands for the $\sigma$-algebra generated by $\mathcal{F}^{(g)}$. To show the asymptotic normality, we use a martingale central limit theorem for triangular arrays \citep[Corollary 2.8 of][]{McLeish1974}. This requires to verify the following two conditions: 
\begin{itemize}
	\item[(a)] $\max_{1\le g\le nT} |\{\int_x \bm{U}(x)\mathbb{G}(dx)\}^{\top}\bm{\Sigma}^{-1} \bm{\xi}^{(g)}\varepsilon^{(g)}|/\{\sqrt{nT}\sigma(\pi;x)\}\stackrel{p}{\to} 0$;
	\item[(b)] $(nT)^{-1}\sum_{g=1}^{nT} |\{\int_x \bm{U}(x)\mathbb{G}(dx)\}^{\top}\bm{\Sigma} \bm{\xi}^{(g)}\varepsilon^{(g)}|^2/\{\sigma^2(\pi;x)\}\stackrel{p}{\to} 1$.
\end{itemize}
Notice that
\begin{eqnarray*}
	&&\left|\frac{\{\int_x \bm{U}(x)\mathbb{G}(dx)\}^{\top}\bm{\Sigma}^{-1} \bm{\xi}^{(g)}\varepsilon^{(g)}}{\sqrt{nT}\sigma(\pi;x)}\right|\le \frac{\|\{\int_x \bm{U}(x)\mathbb{G}(dx)\}^{\top}\bm{\Sigma}^{-1}\|_2 \|\bm{\xi}^{(g)}\|_2 |\varepsilon^{(g)}| }{\sqrt{nT}\sigma(\pi;x)}\\&\le& (c_0+2c')  \frac{\|\{\int_x \bm{U}(x)\mathbb{G}(dx)\}^{\top}\bm{\Sigma}^{-1}\|_2 \|\bm{\xi}^{(g)}\|_2 }{\sqrt{nT}\sigma(\pi;x)}
	\le (c_0+2c') c^* \sqrt{L} \frac{\|\{\int_x \bm{U}(x)\mathbb{G}(dx)\}^{\top}\bm{\Sigma}^{-1}\|_2 }{\sqrt{nT}\sigma(\pi;x)}\\&\le& \frac{\sqrt{3c_0}(c_0+2c')c^*}{\sqrt{\bar{c}}} \frac{\sqrt{L}}{\sqrt{nT}},
\end{eqnarray*}
where the first inequality follows from Cauchy-Schwarz inequality, the second inequality is due to \eqref{maxresponse}, the third inequality is due to Lemma \ref{lemma2} and the fact that $\|\bm{\xi}^{(g)}\|_2\le \sup_{x} \|\Phi_L(x)\|_2$, and the last inequality follows from \eqref{lowerbound1}. Since $L\ll \sqrt{nT}/\log (nT)$, (a) is proven. To verify (b), notice that
\begin{eqnarray*}
	&&\left|\frac{1}{nT}\sum_{g=1}^{nT} \frac{|\{\int_x \bm{U}(x)\mathbb{G}(dx)\}^{\top}\bm{\Sigma}^{-1} \bm{\xi}^{(g)}\varepsilon^{(g)}|^2}{\sigma^2(\pi;x)}-1\right|=\frac{1}{\sigma^2(\pi;x)}\\
	&\times & \left|\left\{\int_x \bm{U}(x)\mathbb{G}(dx)\right\}^{\top}\bm{\Sigma}^{-1} \left\{\frac{1}{nT}\sum_{g=1}^{nT}(\varepsilon^{(g)})^2 \bm{\xi}^{(g)}(\bm{\xi}^{(g)})^{\top}-\bm{\Omega} \right\} (\bm{\Sigma}^{\top})^{-1} \left\{\int_x \bm{U}(x)\mathbb{G}(dx)\right\}\right|\\
	&\le& \frac{\|\{\int_x \bm{U}(x)\mathbb{G}(dx)\}^{\top} \bm{\Sigma}^{-1}\|_2^2}{\sigma^2(\pi;x)}\left\|\frac{1}{nT}\sum_{g=1}^{nT}(\varepsilon^{(g)})^2 \bm{\xi}^{(g)}(\bm{\xi}^{(g)})^{\top}-\bm{\Omega}\right\|_2.
\end{eqnarray*}
In view of \eqref{lowerbound1}, it suffices to show
\begin{eqnarray}\label{proofthm1eq5}
\left\|\frac{1}{nT}\sum_{g=1}^{nT}(\varepsilon^{(g)})^2 \bm{\xi}^{(g)}(\bm{\xi}^{(g)})^{\top}-\bm{\Omega}\right\|_2=o_p(1).
\end{eqnarray}
This can be proven using similar arguments in bounding $\|\widehat{\bm{\Sigma}}-\bm{\Sigma}\|_2$ in the proof of Lemma \ref{lemma3}. In view of \eqref{proofthm1eq3} and \eqref{proofthm1eq4}, we have by Slutsky's theorem that
\begin{eqnarray*}
	\frac{\sqrt{nT}\{\widehat{V}(\pi;\mathbb{G})-V(\pi;\mathbb{G})\} }{\sigma(\pi;\mathbb{G})}\stackrel{d}{\to} N(0,1). 
\end{eqnarray*}
To complete the proof, it remains to show $\widehat{\sigma}(\pi;\mathbb{G})/\sigma(\pi;\mathbb{G})\stackrel{p}{\to} 1$. Using similar arguments in verifying (b), it suffices to show $\|\widehat{\bm{\Sigma}}^{-1} \widehat{\bm{\Omega}} (\widehat{\bm{\Sigma}}^{\top})^{-1}-\bm{\Sigma}^{-1}\bm{\Omega}(\bm{\Sigma}^{\top})^{-1} \|_2=o_p(1)$. By \eqref{maxresponse} and Lemma \ref{lemma4}, we have 
\begin{eqnarray*}
	\lambda_{\max}(\bm{\Omega})\le (c_0+2c')^2 \lambda_{\max}\left(\frac{1}{T} \sum_{t=0}^{T-1}\Mean \bm{\xi}_{0,t}\bm{\xi}_{0,t}^{\top} \right)=O(1),
\end{eqnarray*}
and hence $\|\bm{\Omega}\|_2=O(1)$. This together with Lemma \ref{lemma3} and the condition $L\ll \sqrt{nT}/\log (nT)$ yields that
\begin{eqnarray*}
	\|\widehat{\bm{\Sigma}}^{-1} \bm{\Omega} (\widehat{\bm{\Sigma}}^{\top})^{-1}-{\bm{\Sigma}}^{-1}\bm{\Omega}({\bm{\Sigma}}^{\top})^{-1}\|_2\le \|\widehat{\bm{\Sigma}}^{-1}-\bm{\Sigma}\|_2 \|\bm{\Omega}\|_2 \|(\widehat{\bm{\Sigma}}^{\top})^{-1}\|_2+\|\bm{\Sigma}^{-1}\|_2 \|\bm{\Omega}\|_2 \|\widehat{\bm{\Sigma}}^{-1}-\bm{\Sigma}\|_2\\
	=O_p\{ L^{1/2} (nT)^{-1/2}\log (nT) \}=o_p(1).
\end{eqnarray*}
Thus, it remains to show $\|\widehat{\bm{\Sigma}}^{-1} \widehat{\bm{\Omega}} (\widehat{\bm{\Sigma}}^{\top})^{-1}-\widehat{\bm{\Sigma}}^{-1}\bm{\Omega}(\widehat{\bm{\Sigma}}^{\top})^{-1} \|_2=o_p(1)$, or $\|\widehat{\bm{\Omega}}-\bm{\Omega}\|_2=o_p(1)$, 
by Lemma \ref{lemma3}. In view of \eqref{proofthm1eq5}, it suffices to show $\|(nT)^{-1}\sum_{g=1}^{nT} (\varepsilon^{(g)})^2\bm{\xi}^{(g)}(\bm{\xi}^{(g)})^{\top}-\widehat{\bm{\Omega}}\|_2=o_p(1)$, or equivalently, 
\begin{eqnarray*}
	\sup_{a\in \mathbb{S}^{mL-1}} \left| \frac{1}{nT}\sum_{g=1}^{nT}\bm{a}^{\top}\bm{\xi}^{(g)}(\bm{\xi}^{(g)})^{\top}\bm{a} \{(\varepsilon^{(g)})^2- (\widehat{\varepsilon}^{(g)})^2 \}  \right|=o_p(1),
\end{eqnarray*}
where 
\begin{eqnarray*}
	\widehat{\varepsilon}^{(g)}=Y_{i(g),t(g)}+\gamma \sum_{a\in \mathcal{A}} \Phi_L^{\top}(X_{i(g),t(g)+1})\widehat{\beta}_a\pi(a|X_{i(g),t(g)+1})-\Phi_L^{\top}(X_{i(g),t(g)})\widehat{\beta}_{A_{i(g),t(g)}}.
\end{eqnarray*} 
By Lemma \ref{lemma4}, we have $\sup_{a\in \mathbb{S}^{mL-1}} (nT)^{-1}\sum_{g=1}^{nT}\bm{a}^{\top}\bm{\xi}^{(g)}(\bm{\xi}^{(g)})^{\top}\bm{a}=O_p(1)$. Hence, it suffices to show $\max_{1\le g\le nT} |(\varepsilon^{(g)})^2-(\widehat{\varepsilon}^{(g)})^2|=o_p(1)$. Suppose we have shown that $\max_{1\le g\le nT} |\varepsilon^{(g)}-\widehat{\varepsilon}^{(g)}|=o_p(1)$. By \eqref{maxresponse}, $\varepsilon^{(g)}$s are uniformly bounded with probability $1$ and thus we have $\max_{1\le g\le nT} |\varepsilon^{(g)}+\widehat{\varepsilon}^{(g)}|=O_p(1)$. It follows that
\begin{eqnarray*}
	\max_{1\le g\le nT} |(\varepsilon^{(g)})^2-(\widehat{\varepsilon}^{(g)})^2|\le \max_{1\le g\le nT} |\varepsilon^{(g)}-\widehat{\varepsilon}^{(g)}|\max_{1\le g\le nT} |\varepsilon^{(g)}+\widehat{\varepsilon}^{(g)}|=o_p(1). 
\end{eqnarray*}
Therefore, it remains to show $\max_{1\le g\le nT} |\varepsilon^{(g)}-\widehat{\varepsilon}^{(g)}|=o_p(1)$, or equivalently, 
\begin{eqnarray}\label{proofthm1eq6}
\begin{split}
\max_{1\le i\le n,0\le t\le T-1} \left|\gamma \sum_{a\in \mathcal{A}} Q(\pi;X_{i,t+1},a)\pi(a|X_{i,t+1})-Q(\pi;X_{i,t},A_{i,t})\right.\\
\left.-\gamma \sum_{a\in \mathcal{A}} \Phi_L^{\top}(X_{i,t+1})\widehat{\beta}_a\pi(a|X_{i,t+1})-\Phi_L^{\top}(X_{i,t})\widehat{\beta}_{A_{i,t}} \right|=o_p(1).
\end{split}
\end{eqnarray}
The LHS of \eqref{proofthm1eq6} is upper bound by
\begin{eqnarray*}
	(1+\gamma)\sup_{x\in \mathbb{X},a\in \mathcal{A}}  |Q(\pi;x,a)-\Phi_L^{\top}(x)\widehat{\beta}_a|. 
\end{eqnarray*}
By \eqref{biasbound}, \eqref{betaerrorbound1} and Lemma 2, we have
\begin{eqnarray*}
	\sup_{x\in \mathbb{X},a\in \mathcal{A}}  |Q(\pi;x,a)-\Phi_L^{\top}(x)\widehat{\beta}_a|\le \sup_{x\in \mathbb{X},a\in \mathcal{A}}  |Q(\pi;x,a)-\Phi_L^{\top}(x){\beta}^*_a|+\sup_{x\in \mathbb{X}} |\Phi_L^{\top}(x)\widehat{\beta}_a-\Phi_L^{\top}(x){\beta}^*_a|\\
	\le CL^{-p/d}+\sup_{x\in \mathbb{X}}\|\Phi_L(x)\|_2 \sup_{a\in \mathcal{A}}\|\widehat{\beta}_a-\beta_a^*\|_2=O(L^{-p/d})+O(L^{1/2}) \sup_{a\in \mathcal{A}}\|\widehat{\beta}_a-\beta_a^*\|_2\\
	=O_p(L^{1/2-p/d})+O_p(Ln^{-1/2}T^{-1/2}).
\end{eqnarray*}
Under the given conditions, we have $L^{p/d}\gg \sqrt{nT}$ and $L\ll \sqrt{nT}/\log (nT)$. This implies $O_p(L^{1/2-p/d})=o_p(1)$, and $O_p(Ln^{-1/2}T^{-1/2})=o_p(1)$. Therefore, we have
\begin{eqnarray*}
	\sup_{x\in \mathbb{X},a\in \mathcal{A}}  |Q(\pi;x,a)-\Phi_L^{\top}(x)\widehat{\beta}_a|=o_p(1). 
\end{eqnarray*}
The proof is hence completed.


\subsection{Proof of Lemma \ref{lemma2}}
For B-spline basis, the assertion in \eqref{lemma2eq1} follows from the arguments used in the proof of Theorem 3.3 of \cite{Burman1989}. For wavelet basis, the assertion in \eqref{lemma2eq1} follows from the arguments used in the proof of Theorem 5.1 of \cite{Chen2015}. 

For either B-spline or wavelet sieve and any $L\ge 1$, $x\in \mathbb{X}$, the number of nonzero elements in the vector $\Phi_L(x)$ is bounded by some constant. Moreover, each of the basis function is uniformly bounded by $O(\sqrt{L})$. This proves that the second assertion. 

\subsection{Proof of Lemma \ref{lemma3}}
We consider two scenarios: (i) $T$ grows to infinity; (ii) $T$ is bounded. 
The proof is divided into four parts. In the first part, we show $\bm{a}^{\top} \bm{\Sigma} \bm{a}\ge \bar{c} \|\bm{a}\|_2^2/2$ for any $\bm{a}\in \mathbb{R}^{mL}$ and $\|\bm{\Sigma}^{-1}\|_2\le 2/\bar{c}$, as either $n\to \infty$, or $T\to \infty$. In the second part, we bound $\|\widehat{\bm{\Sigma}}-\bm{\Sigma}\|_2$. In the third part, we bound $\|\widehat{\bm{\Sigma}}^{-1}-\bm{\Sigma}^{-1}\|_2$. Finally, we show $\|\bm{\Sigma}\|_2=O(1)$. 

\noindent \textit{Part 1:} 
It follows from Cauchy-Schwarz inequality that
\begin{eqnarray*}
	\int_{x\in \mathbb{X}}\sum_{a\in \mathcal{A}} \{\bm{a}^{\top} \bm{\xi}(x,a)\} [\Mean\{\bm{U}(X_{0,1})|X_{0,0}=x,A_{0,0}=a\}^{\top} \bm{a}] b(a|x)\bar{\mu}(x)dx\le \eta_1^{1/2}(\bm{a}) \eta_2^{1/2}(\bm{a}),
\end{eqnarray*}
where $\bar{\mu}=T^{-1}\sum_{t=0}^{T-1}\mu_t$, $\mu_t$ is the marginal density of $X_{0,t}$ and
\begin{eqnarray*}
	\eta_1(\bm{a})=\int_{x\in \mathbb{X}}\sum_{a\in \mathcal{A}} \{\bm{a}^{\top} \bm{u}(x,a)\}^2 b(a|x)\bar{\mu}(x)dx,\,\,\,\,
	\eta_2(\bm{a})=\int_{x\in \mathbb{X}}\sum_{a\in \mathcal{A}} \{\bm{a}^{\top} \bm{\xi}(x,a)\}^2 b(a|x)\bar{\mu}(x)dx.
\end{eqnarray*}
Therefore,
\begin{eqnarray*}
	\bm{a}^{\top} \bm{\Sigma} \bm{a}\ge \eta_2(\bm{a})-\sqrt{\eta_2(\bm{a})} \sqrt{\gamma^2 \eta_1(\bm{a})}=\sqrt{\eta_2(\bm{a})} \frac{\eta_2(\bm{a})-\gamma^2 \eta_1(\bm{a})}{\sqrt{\eta_2(\bm{a})}+\gamma\sqrt{\eta_1(\bm{a})}}.
\end{eqnarray*}
Under A3(i), we obtain $\eta_2(\bm{a})-\gamma^2 \eta_1(\bm{a})\ge \bar{c} \|\bm{a}\|_2^2$ for $\bm{a}\in \mathbb{R}^{mL}$. It follows that $\gamma\sqrt{\eta_1(\bm{a})}\le \sqrt{\eta_2(\bm{a})}$ and
\begin{eqnarray}\label{prooflemma3eq2}
\bm{a}^{\top} \bm{\Sigma} \bm{a}\ge \frac{\bar{c} \|\bm{a}\|_2^2 \sqrt{\eta_2(\bm{a})}}{\sqrt{\eta_2(\bm{a})}+\gamma\sqrt{\eta_1(\bm{a})}}\ge \frac{\bar{c}}{2}\|\bm{a}\|_2^2,\,\,\,\,\,\,\,\,\forall \bm{a}\in \mathbb{R}^{mL}. 
\end{eqnarray}
We now show 
\begin{eqnarray}\label{prooflemma3eq3}
\|\bm{\Sigma} \bm{a}\|_2\ge \frac{\bar{c}}{2}\|\bm{a}\|_2,\,\,\,\,\,\,\,\,\forall \bm{a}\in \mathbb{R}^{mL}.
\end{eqnarray}
Otherwise, there exists some $\bm{a}_0\in \mathbb{R}^{mL}$ such that $\|\bm{\Sigma} \bm{a}_0\|_2< 2^{-1}\bar{c}\|\bm{a}_0\|_2$. By Cauchy-Schwarz inequality, we obtain $\bm{a}_0^{\top} \bm{\Sigma} \bm{a}_0\le \|\bm{a}_0\|_2 \|\bm{\Sigma}\bm{a}_0\|_2<2^{-1}\bar{c}\|\bm{a}_0\|_2^2$. However, this violates the assertion in \eqref{prooflemma3eq2}. \eqref{prooflemma3eq3} is thus proven. 

According to the singular value decomposition, we have $\bm{\Sigma}=\bm{V}_1^{\top} \bm{\Lambda} \bm{V}_2$ for some orthogonal matrices $\bm{V}_1$, $\bm{V}_2$ and some diagonal matrix $\bm{\Lambda}$. By orthogonality, we obtain $\|\bm{\Sigma}\bm{a}\|_2=\|\bm{\Lambda}\bm{V}_2\bm{a}\|_2$ and $\|\bm{a}\|_2=\|\bm{V}_2\bm{a}\|_2$. In view of \eqref{prooflemma3eq3}, we have $\|\bm{\Lambda}\bm{a}\|_2\ge 2^{-1}\bar{c}\|\bm{a}\|_2,\forall \bm{a}\in \mathbb{R}^{mL}$. This implies that the absolute value of each diagonal element in $\bm{\Lambda}$ is at least $\bar{c}$. Thus, we obtain $\|\bm{\Lambda}^{-1}\|_2\le 2\bar{c}^{-1}$ and hence $\|\bm{\Sigma}^{-1}\|_2\le 2\bar{c}^{-1}$. 


\noindent \textit{Part 2: }We first consider Scenario (ii). Define the random matrix 
\begin{eqnarray*}
	\bm{R}_i=\frac{1}{T}\sum_{t=0}^{T-1}\bm{\xi}(X_{i,t},A_{i,t})\{\bm{\xi}(X_{i,t},A_{i,t})-\gamma \bm{U}(X_{i,t+1}) \}^{\top}.
\end{eqnarray*}
By Lemma \ref{lemma2}, we have $\max_{1\le i\le n,0\le t\le T-1}\|\bm{\xi}(X_{i,t},A_{i,t})\|_2\le \sup_{x} \|\Phi_L(x)\|_2\le c^*\sqrt{L}$ and $\max_{1\le i\le n,1\le t\le T} \|\bm{U}(X_{i,t+1})\|_2\le \sup_{x} \|\Phi_L(x)\|_2\le c^*\sqrt{L}$. It follows that
\begin{eqnarray}\label{maxRi}
\max_{1\le i\le n}\|\bm{R}_i-\Mean \bm{R}_i\|_2\le \frac{2}{T}\sum_{t=0}^{T-1} c^*\sqrt{L}(c^*\sqrt{L}+\gamma c^*\sqrt{L})\le 4L(c^*)^2. 
\end{eqnarray}
Let
\begin{eqnarray*}
	\sigma_n^2&=&\max\left\{ \left\|\sum_{i=1}^n \Mean (\bm{R}_i-\Mean \bm{R}_i) (\bm{R}_i-\Mean \bm{R}_i)^{\top}\right\|_2, \left\|\sum_{i=1}^n \Mean(\bm{R}_i-\Mean \bm{R}_i)^{\top} (\bm{R}_i-\Mean \bm{R}_i)\right\|_2 \right\}\\
	&=&n\max\left\{\left\|\Mean (\bm{R}_0-\Mean \bm{R}_0) (\bm{R}_0-\Mean \bm{R}_0)^{\top}\right\|_2,\left\|\Mean (\bm{R}_0-\Mean \bm{R}_0)^{\top} (\bm{R}_0-\Mean \bm{R}_0)\right\|_2 \right\}.
\end{eqnarray*}
For any $\bm{v}\in \mathbb{R}^{mL}$, we have
\begin{eqnarray*}
	\bm{v}^{\top} \Mean (\bm{R}_0-\Mean \bm{R}_0) (\bm{R}_0-\Mean \bm{R}_0)^{\top} \bm{v}=\bm{v}^{\top} \Mean \bm{R}_0\bm{R}_0^{\top} \bm{v}-\bm{v}^{\top} (\Mean \bm{R}_0)(\Mean \bm{R}_0)^{\top} \bm{v}\le \bm{v}^{\top} \Mean \bm{R}_0\bm{R}_0^{\top} \bm{v}.
\end{eqnarray*}
Moreover, using similar arguments in proving \eqref{maxRi}, we can show
\begin{eqnarray*}
	\bm{v}^{\top} \Mean \bm{R}_0\bm{R}_0^{\top} \bm{v}\le \Mean \left\|\frac{1}{T}\sum_{t=0}^{T-1}\bm{v}^{\top}\bm{\xi}(X_{0,t},A_{0,t})\{\bm{\xi}(X_{0,t},A_{0,t})-\gamma \bm{U}(X_{0,t+1}) \}^{\top}\right\|_2^2\\
	\le 4L(c^*)^2 \Mean \left\{\frac{1}{T}\sum_{t=0}^{T-1}\bm{v}^{\top}\bm{\xi}(X_{0,t},A_{0,t})\right\}^2.
\end{eqnarray*}
By Cauchy-Schwarz inequality, we obtain
\begin{eqnarray*}
	&&\bm{v}^{\top} \Mean \bm{R}_0\bm{R}_0^{\top} \bm{v}\le \frac{4 L (c^*)^2}{T} \sum_{t=0}^{T-1} \Mean \{\bm{v}^{\top}\bm{\xi}(X_{0,t},A_{0,t})\}^2\\
	&\le& \frac{4 L (c^*)^2 \|\bm{v}^2\|_2}{T} \sum_{t=0}^{T-1}\lambda_{\max}\left\{\Mean \bm{\xi}(X_{0,t},A_{0,t})\bm{\xi}(X_{0,t},A_{0,t})^{\top}\right\}.
\end{eqnarray*}
Similarly, we can show
\begin{eqnarray*}
	\bm{v}^{\top} \Mean \bm{R}_0^{\top}\bm{R}_0 \bm{v}\le \frac{4 L (c^*)^2 \|\bm{v}^2\|_2}{T} \sum_{t=0}^{T-1}\lambda_{\max}\{\Mean \bm{\xi}(X_{0,t},A_{0,t})\bm{\xi}(X_{0,t},A_{0,t})^{\top}\},
\end{eqnarray*}
and hence
\begin{eqnarray}\label{prooflemma3eq4}
\sigma_n^2\le \frac{4Ln(c^*)^2}{T}\sum_{t=0}^{T-1}\lambda_{\max}\{\Mean \bm{\xi}(X_{0,t},A_{0,t})\bm{\xi}(X_{0,t},A_{0,t})^{\top}\}.
\end{eqnarray}
Consider $\lambda_{\max}\{ \Mean \bm{\xi}(X_{0,0},A_{0,0})\bm{\xi}(X_{0,0},A_{0,0})^{\top} \}$ first. Notice that $\Mean \bm{\xi}(X_{0,0},A_{0,0})\bm{\xi}(X_{0,0},A_{0,0})^{\top}$ is a block diagonal matrix. For any $\bm{v}\in \mathbb{R}^{mL}$, let $\bm{v}=(a_1^{\top},\dots,a_m^{\top})^{\top}$ where all the sub-vectors $a_j$s have the same length. With some calculations, we have
\begin{eqnarray*}
	&&\bm{v}^{\top} \Mean \bm{\xi}(X_{0,0},A_{0,0})\bm{\xi}(X_{0,0},A_{0,0})^{\top} \bm{v}=\sum_{j=1}^m \Mean \{a_j^{\top} \Phi_L(X_{0,0})\}^2 b(j|X_{0,0})\\&\le& \lambda_{\max}\{\Mean \Phi_L(X_{0,0}) \Phi_L^{\top}(X_{0,0})\} \|\bm{v}\|_2^2
	\le \lambda_{\max}\left\{ \int_{x\in \mathbb{X}} \Phi_L(x) \Phi_L^{\top}(x)\nu_0(x)dx \right\} \|\bm{v}\|_2^2.
\end{eqnarray*} 
By Condition A2 and Lemma \ref{lemma2}, we obtain
\begin{eqnarray*}
	\lambda_{\max}\left\{ \int_{x\in \mathbb{X}} \Phi_L(x) \Phi_L^{\top}(x)\nu_0(x)dx \right\}\le \sup_{x\in \mathbb{X}} \nu_0(x) \lambda_{\max}\left\{ \int_{x\in \mathbb{X}} \Phi_L(x) \Phi_L^{\top}(x)dx \right\}\preceq 1.
\end{eqnarray*}
This yields
\begin{eqnarray}\label{eigenmax0}
\lambda_{\max}\{\Mean \bm{\xi}(X_{0,0},A_{0,0})\bm{\xi}(X_{0,0},A_{0,0})^{\top}\}\preceq 1.
\end{eqnarray}
For any $t>0$, the marginal density function of $X_{0,t}$ is given by
\begin{eqnarray}\label{marginaldensityt}
\mu_t(x)=\int_{x_0,\dots,x_{t-1}\in \mathbb{X}} \nu_0(x_0)q_X(x_1|x_0)\cdots q_X(x_{t-1}|x_{t-2})q_X(x|x_{t-1})dx_0dx_1\dots d_{x_{t-1}}. 
\end{eqnarray} 
Thus, we have $\mu_t(x)\le \sup_{x',x''}q_X(x'|x'')$ for any $t\ge 1$ and $x\in \mathbb{X}$.
Under Condition A1, we can show the density function $q_X(x|x')$ is uniformly bounded for any $x$ and $x'$. It follows that $\mu_t(x)$ is uniformly bounded for any $t\ge 1$ and $x$. Using similar arguments in proving \eqref{eigenmax0}, we can show
\begin{eqnarray}\label{eigenmaxt}
\lambda_{\max}\{\Mean \bm{\xi}(X_{0,t},A_{0,t})\bm{\xi}(X_{0,t},A_{0,t})^{\top}\}\preceq 1,\,\,\,\,\,\,\,\,\forall 1\le t\le T.
\end{eqnarray}
This together with \eqref{prooflemma3eq4} and \eqref{eigenmax0} yields 
\begin{eqnarray}\label{prooflemma3eq4.5}
\sigma_n^2 \le CnL,
\end{eqnarray}
for some constant $C>0$. Combining this together with \eqref{maxRi}, an application of the matrix concentration inequality \citep[see Theorem 1.6 in][]{Tropp2012} yields that
\begin{eqnarray*}
	\prob\left( \left\|\sum_{i=1}^n \bm{R}_i-n\Mean \bm{R}_0\right\|_2\ge \tau \right)\le 2mL \exp\left( -\frac{\tau^2}{CnL+ 8L(c^*)^2\tau/3 } \right),\,\,\,\,\,\,\,\,\forall \tau>0.
\end{eqnarray*}
Set $\tau=3 \sqrt{C n L\log n}$. Since $T$ is bounded, under the given conditions, $n$ will grow to infinity. For sufficiently large $n$, we have $8L(c^*)^2\tau/3\ll \tau^2$ and hence 
\begin{eqnarray*}
	\prob\left( \left\|\sum_{i=1}^n \bm{R}_i-n\Mean \bm{R}_0\right\|_2\ge 3 \sqrt{C n L\log n} \right)\le \frac{2mL}{n^4}.
\end{eqnarray*}
Since $L\ll n$ and $T$ is bounded, we obtain $2mL/n^4\ll 1/(n^2 T^2)$. 
Thus, we can show that the following event occurs with probability at least $1-O(n^{-2} T^{-2})$,  
\begin{eqnarray}\label{prooflemma3eq5}
\left\|\widehat{\bm{\Sigma}}-\bm{\Sigma}\right\|_2=\frac{1}{n} \left\|\sum_{i=1}^n \bm{R}_i-n\Mean \bm{R}_0\right\|_2\preceq \sqrt{(nT)^{-1} L \log (nT) },
\end{eqnarray}
since $T$ is bounded.

Now let's consider Scenario (i). 
Let
\begin{eqnarray*}
	\bm{R}_{i,t}=\bm{\xi}(X_{i,t},A_{i,t})\{\bm{\xi}(X_{i,t},A_{i,t})-\gamma \bm{U}(X_{i,t+1})\}^{\top},\,\,\,\,\,\,\,\,\forall i,t.
\end{eqnarray*}
We aim to apply the matrix concentration inequality to the sum of independent random matrix (regardless of whether $n$ is bounded or not),
\begin{eqnarray*}
	\frac{1}{n}\sum_{i=1}^n \left\{ \frac{1}{T}\sum_{t=0}^{T-1} (\bm{R}_{i,t}-\bm{\Sigma}) \right\}.
\end{eqnarray*}
We begin by providing an upper error bound for $\max_{1\le i\le n}\|T^{-1} \sum_{t=0}^{T-1}(\bm{R}_{i,t}-\bm{\Sigma})\|_2$. 
Let $\mathcal{F}_{t-1}=\{ (X_{0,j},A_{0,j}) \}_{0\le j\le t}$, for all $t\ge 0$, and $\sigma(\mathcal{F}_t)$ be the $\sigma$-algebra generated by $\mathcal{F}_t$. Define
\begin{eqnarray*}
	\bm{R}_{0,t}^{*}=\bm{\xi}(X_{0,t},A_{0,t})\{\bm{\xi}(X_{0,t},A_{0,t})-\gamma \bm{u}(X_{0,t},A_{0,t})\}^{\top}.
\end{eqnarray*}
The sum $\sum_{t=0}^{T-1} (\bm{R}_{0,t}^*-\bm{R}_{0,t})$ forms a mean zero matrix martingale with respect to the filtration $\{\sigma(\mathcal{F}_t):t\ge -1\}$. Similar to \eqref{maxRi} and \eqref{prooflemma3eq4.5}, we can show
\begin{eqnarray*}
	&&\max_{0\le t\le T-1} \|\bm{R}_{0,t}-\bm{R}_{0,t}^*\|_2\le 4L (c^*)^2,\\
	&&\max\left\{\left\| \sum_{t=0}^{T-1} \Mean \{(\bm{R}_{0,t}-\bm{R}_{0,t}^*)^{\top} (\bm{R}_{0,t}-\bm{R}_{0,t}^*)|\mathcal{F}_{t-1} \} \right\|_2,\right.\\
	&&\left.\left\| \sum_{t=0}^{T-1} \Mean \{(\bm{R}_{0,t}-\bm{R}_{0,t}^*) (\bm{R}_{0,t}-\bm{R}_{0,t}^*)^{\top}|\mathcal{F}_{t-1} \} \right\|_2 \right\}\preceq LT. 
\end{eqnarray*}
By the matrix martingale concentration inequality \citep[Corollary 1.3,][]{Tropp2011}, we obtain the following occurs with probability at least $1-O(n^{-3} T^{-2})$,
\begin{eqnarray}\label{firstbound}
\left\|\sum_{t=0}^{T-1} (\bm{R}_{0,t}-\bm{R}_{0,t}^*)\right\|_2\preceq \sqrt{LT \log (nT)}. 
\end{eqnarray}
Define 
\begin{eqnarray}\label{R0t3star}
\bm{R}_{0,t}^{**}=\sum_{a\in \mathcal{A}}\bm{\xi}(X_{0,t}^*,a)\{\bm{\xi}(X_{0,t}^*,a)-\gamma \bm{u}(X_{0,t}^*,a)\}^{\top} b(a|X_{0,t}^*).
\end{eqnarray}
Conditional on $\{X_{0,t}^*\}_{t\ge 0}$, $\{\bm{R}_{0,t}^*-\bm{R}_{0,t}^{**}\}_{t\ge 0}$ are independent mean zero random variables. Using similar arguments in proving \eqref{prooflemma3eq5}, we can show that
\begin{eqnarray*}
	\hbox{Pr}\left( \left. \left\|\sum_{t=0}^{T-1} (\bm{R}_{0,t}^*-\bm{R}_{0,t}^{**})\right\|_2\ge C\sqrt{LT\log (nT)} \right| \{X_{0,t}^*\}_{t\ge 0} \right)=O(n^{-3} T^{-2}),
\end{eqnarray*}
for some constant $C>0$, where the big-$O$ term is independent of $\{X_{0,t}^*\}_{t\ge 0}$. Thus, we obtain
\begin{eqnarray*}
	\hbox{Pr}\left( \left\|\sum_{t=0}^{T-1} (\bm{R}_{0,t}^*-\bm{R}_{0,t}^{**})\right\|_2\ge C\sqrt{LT\log (nT)}  \right)=O(n^{-3} T^{-2}),
\end{eqnarray*}
This together with \eqref{firstbound} implies that the following event occurs with probability at least $1-O(n^{-3} T^{-2})$,
\begin{eqnarray}\label{secondbound}
\left\|\sum_{t=0}^{T-1} (\bm{R}_{0,t}-\bm{R}_{0,t}^{**})\right\|_2\preceq \sqrt{LT \log (nT)}.
\end{eqnarray} 
Notice that each $\bm{R}_{0,t}^{**}$ is a function of $X_{0,t}$ only, with mean $\bm{\Sigma}$. 
Following \cite{davydov1973}, define the $\beta$-mixing coefficient of the stationary Markov chain $\{X_{0,t}\}_{t\ge 0}$ as 
\begin{eqnarray*}
	\beta(q)=\int_{x\in \mathbb{X}} \sup_{0\le \varphi \le 1} \left|\Mean \{\varphi(X_{0,q})|X_{0,0}=x\}- \Mean \varphi(X_{0,0}) \right|\mu(x)dx.
\end{eqnarray*}
Under the geometric ergodicity assumption in A3(ii) and , it follows from Lemma 1 of \cite{meitz2019subgeometric} that $\{X_{0,t}\}_{t\ge 0}$ is exponentially $\beta$-mixing. That is, $\beta(t)= O(\rho^t)$ for some $\rho< 1$ and any $t\ge 0$. Using similar arguments in proving \eqref{maxRi},  we can show
\begin{eqnarray}\label{prooflemma3eq7}
\max_{0\le t\le T-1} \|\bm{R}_{0,t}^{**}-\bm{\Sigma}\|_2\le 4L(c^*)^2.
\end{eqnarray}
Moreover, for any $0\le t_1\le t_2\le T-1$ and any $\bm{v}_1,\bm{v}_2\in \mathbb{R}^{mL}$, we have by Cauchy-Schwarz inequality that
\begin{eqnarray*}
	&&|\bm{v}_1^{\top} \Mean (\bm{R}_{0,t_1}^{**}-\bm{\Sigma})  (\bm{R}_{0,t_2}^{**}-\bm{\Sigma})^{\top} \bm{v}_2|\le \sqrt{\Mean\|\bm{v}_1^{\top} (\bm{R}_{0,t_1}^{**}-\bm{\Sigma})\|_2^2}\sqrt{\Mean\|\bm{v}_2^{\top} (\bm{R}_{0,t_2}^{**}-\bm{\Sigma})\|_2^2}\\
	&\le& \sqrt{\lambda_{\max} \{\Mean (\bm{R}_{0,t_1}^{**}-\bm{\Sigma}) (\bm{R}_{0,t_1}^{**}-\bm{\Sigma})^{\top} \}}\sqrt{\lambda_{\max} \{\Mean (\bm{R}_{0,t_2}^{**}-\bm{\Sigma}) (\bm{R}_{0,t_2}^{**}-\bm{\Sigma})^{\top} \}} \|\bm{v}_1\|_2 \|\bm{v}_2\|_2.
\end{eqnarray*}
Using similar arguments in proving \eqref{prooflemma3eq4.5}, we can show
\begin{eqnarray*}
	\max_{0\le t\le T-1}\lambda_{\max} \{\Mean (\bm{R}_{0,t}^{**}-\bm{\Sigma}) (\bm{R}_{0,t}^{**}-\bm{\Sigma})^{\top}\}\preceq L.
\end{eqnarray*}
This implies 
\begin{eqnarray*}
	\max_{0\le t_1,t_2\le T-1}\sup_{\bm{v}_1\neq 0,\bm{v}_2\neq 0} \frac{|\bm{v}_1^{\top} \Mean (\bm{R}_{0,t_1}^{**}-\bm{\Sigma}) (\bm{R}_{0,t_2}^{**}-\bm{\Sigma})^{\top} \bm{v}_2|}{\|\bm{v}_1\|_2\|\bm{v}_2\|_2}\preceq L,
\end{eqnarray*}
and hence,
\begin{eqnarray*}
	\max_{0\le t_1,t_2\le T-1}\sup_{\bm{v}\neq 0} \frac{\|\Mean (\bm{R}_{0,t_1}^{**}-\bm{\Sigma}) (\bm{R}_{0,t_2}^{**}-\bm{\Sigma})^{\top} \bm{v}\|_2}{\|\bm{v}\|_2}\preceq L,
\end{eqnarray*}
or equivalently,
\begin{eqnarray*}
	\max_{0\le t_1,t_2\le T-1}\|\Mean (\bm{R}_{0,t_1}^{**}-\bm{\Sigma})  (\bm{R}_{0,t_2}^{**}-\bm{\Sigma})^{\top}\|_2\preceq L.
\end{eqnarray*}
Similarly, we can show 
\begin{eqnarray}\label{prooflemma3eq7.5}
\max_{0\le t_1,t_2\le T-1}\|\Mean (\bm{R}_{0,t_1}^{**}-\bm{\Sigma})^{\top}  (\bm{R}_{0,t_2}^{*** }-\bm{\Sigma})\|_2\preceq L.
\end{eqnarray}
Let $\bm{\Sigma}_t=\Mean \bm{R}_{0,t}$. Notice that $\bm{\Sigma}=T^{-1}\sum_{t=0}^{T-1} \bm{\Sigma}_t$ and we have $\sum_{t=0}^{T-1} (\bm{R}_{0,t}-\bm{\Sigma})=\sum_{t=0}^{T-1} (\bm{R}_{0,t}-\bm{\Sigma}_t)$.  
Similar to Theorem 4.2 of \cite{Chen2015}, we can show there exist some constant $C>0$ such that for any $\tau\ge 0$ and integer $1<q<T$,
\begin{eqnarray}\nonumber
\hbox{Pr}\left( \left\|\sum_{t=0}^{T-1} (\bm{R}_{0,t}^{**}-\bm{\Sigma}_t) \right\|_2\ge 6\tau \right)\le \frac{T}{q}\beta(q)+\prob\left( \left\|\sum_{t\in \mathcal{I}_r} (\bm{R}_{0,t}^{**}-\bm{\Sigma}_t) \right\|_2\ge \tau \right)\\\label{prooflemma3eq8}
+4mL \exp\left( -\frac{\tau^2/2}{CTqL+4qL\tau(c^*)^2/3} \right),
\end{eqnarray}
where $\mathcal{I}_r=\{q\floor{(T+1)/q},q\floor{(T+1)/q} +1, \cdots,T-1\}$. Suppose $\tau\ge 5q L (c^*)^2$. Notice that $|\mathcal{I}_r|\le q$. It follows from \eqref{prooflemma3eq7} that
\begin{eqnarray}\label{prooflemma3eq8.5}
\prob\left( \left\|\sum_{t\in \mathcal{I}_r} (\bm{R}_{0,t}^{**}-\bm{\Sigma}_t) \right\|_2\ge \tau \right)=0.
\end{eqnarray}
Since $\beta(q)=O(\rho^q)$, set $q=-3\log (nT)/\log \rho$, we obtain $T\beta(q)/q=O(n^{-3} T^{-2})$. Set $\tau=\max\{4\sqrt{CTqL\log (Tn)}, 11qL(c^*)^2\log(nT)\}$, we obtain that
\begin{eqnarray*}
	\frac{\tau^2}{4}\ge 4CTqL\log (Tn)\,\,\,\,\hbox{and}\,\,\,\,\frac{\tau^2}{4}\ge 16 qL\tau (c^*)^2/3 \log (Tn)\,\,\,\,\hbox{and}\,\,\,\,\tau \ge 5q L (c^*)^2,
\end{eqnarray*}
as either $n\to \infty$ or $T\to \infty$. It follows from \eqref{prooflemma3eq8}, \eqref{prooflemma3eq8.5} and the condition $L\ll nT$ that the following event occurs with probability at least $1-O(n^{-3} T^{-2})$,
\begin{eqnarray*}
	\left\|\sum_{t=0}^{T-1} (\bm{R}_{0,t}^{**}-\bm{\Sigma}) \right\|_2\preceq \max\{ \sqrt{TL}\log (Tn), L\log^2 (Tn) \}.
\end{eqnarray*}
Combining this together with \eqref{secondbound} yields that the following event occurs with probability at least $1-O(n^{-3}T^{-2})$,
\begin{eqnarray}\label{immediatestep0}
\left\|\sum_{t=0}^{T-1} (\bm{R}_{0,t}-\bm{\Sigma}) \right\|_2\preceq \max\{ \sqrt{TL}\log (Tn), L\log^2 (Tn) \}.
\end{eqnarray}
By Bonferroni's inequality, we obtain with probability at least $1-O(n^{-2} T^{-2})$ that
\begin{eqnarray}\label{prooflemma3eq9}
\max_{i\in \{1,\dots,n\}}\left\|\sum_{t=0}^{T-1} (\bm{R}_{i,t}-\bm{\Sigma}) \right\|_2\le \bar{C} \max\{ \sqrt{TL}\log (Tn), L\log^2 (Tn) \},
\end{eqnarray}
for some constant $\bar{C}>0$. For $i=0,1,\dots,n$, let $\mathcal{A}_i$ denote the event
\begin{eqnarray*}
	\mathcal{A}_i=\left\{\left\|\sum_{t=0}^{T-1} (\bm{R}_{i,t}-\bm{\Sigma}) \right\|_2\le \bar{C} \max\{ \sqrt{TL}\log (Tn), L\log^2 (Tn) \}
	\right\}.
\end{eqnarray*}
It follows from \eqref{prooflemma3eq9} that the following event occurs with probability at least $1-O(n^{-2} T^{-2})$,
\begin{eqnarray}\label{prooflemma3eq9.5}
\sum_{i=1}^n \sum_{t=0}^{T-1}(\bm{R}_{i,t}-\bm{\Sigma})=\sum_{i=1}^n \left\{\sum_{t=0}^{T-1}(\bm{R}_{i,t}-\bm{\Sigma})\right\}\mathbb{I}(\mathcal{A}_i). 
\end{eqnarray}
Now we provide an upper error bound for
\begin{eqnarray*}
	\left\| \Mean \left\{ \sum_{t=0}^{T-1} (\bm{R}_{0,t}-\bm{\Sigma}) \right\}^{\top} \left\{ \sum_{t=0}^{T-1} (\bm{R}_{0,t}-\bm{\Sigma}) \right\} \right\|_2=\sup_{\bm{v}\in \mathbb{S}^{mL-1}} \Mean \left\|\left\{ \sum_{t=0}^{T-1} (\bm{R}_{0,t}-\bm{\Sigma}) \right\} \bm{v} \right\|_2^2.
\end{eqnarray*}
Note that
\begin{eqnarray}\nonumber
&&\sup_{\substack{\bm{v}\in \mathbb{S}^{mL-1}}} \Mean \bm{v}^{\top} \left\{ \sum_{t=0}^{T-1} (\bm{R}_{0,t}-\bm{\Sigma}) \right\}^{\top} \left\{ \sum_{t=0}^{T-1} (\bm{R}_{0,t}-\bm{\Sigma}) \right\} \bm{v}\\ \label{ineqn1}
&\le& \underbrace{\sum_{t=0}^{T-1} \sup_{\substack{\bm{v}\in \mathbb{S}^{mL-1}}} \Mean \bm{v}^{\top} (\bm{R}_{0,t}-\bm{\Sigma}_t)^{\top} (\bm{R}_{0,t}-\bm{\Sigma}_t)\bm{v}}_{\eta_1}\\ \nonumber
&+&\underbrace{\sum_{\substack{0\le t_1,t_2\le T-1\\ 1\le |t_1-t_2|\le 3 }} \sup_{\substack{\bm{v}\in \mathbb{S}^{mL-1}}} \Mean \bm{v}^{\top} (\bm{R}_{0,t_1}-\bm{\Sigma}_{t_1})^{\top} (\bm{R}_{0,t_2}-\bm{\Sigma}_{t_2})\bm{v}}_{\eta_2}\\ \nonumber
&+&\underbrace{\sum_{\substack{0\le t_1,t_2\le T-1\\ |t_1-t_2|\ge 4 }} \sup_{\substack{\bm{v}\in \mathbb{S}^{mL-1}}} \Mean \bm{v}^{\top} (\bm{R}_{0,t_1}-\bm{\Sigma}_{t_1})^{\top} (\bm{R}_{0,t_2}-\bm{\Sigma}_{t_2})\bm{v}}_{\eta_3}.
\end{eqnarray}
By \eqref{prooflemma3eq7.5}, we obtain that
\begin{eqnarray}\label{firstwoterms}
\eta_1\preceq LT\,\,\,\,\hbox{and}\,\,\,\,\eta_2\preceq LT.
\end{eqnarray}
For any $0\le t_1<t_2\le T-1$ with $t_2-t_1\ge 4$, it follows from MA that $(X_{0,t_2}, A_{0,t_2}, X_{0,t_2+1})$ is independent of $(X_{0,t_1},A_{0,t_1},X_{0,t_1+1})$ given $X_{0,t_2-1}$. Thus, we have
\begin{eqnarray*}
	\Mean (\bm{R}_{0,t_1}-\bm{\Sigma}_{t_1})^{\top} (\bm{R}_{0,t_2}-\bm{\Sigma}_{t_2})=\Mean (\bm{R}_{0,t_1}-\bm{\Sigma}_{t_1})^{\top} \Mean \{(\bm{R}_{0,t_2}-\bm{\Sigma}_{t_2})|X_{0,t_2-1}\}.
\end{eqnarray*}
Similarly, conditional on $X_{0,t_1+2}$, $(X_{0,t_1}, A_{0,t_1},X_{0,t_1+1})$ and $X_{0,t_2-1}$ are independent. It follows that
\begin{eqnarray*}
	&&\Mean (\bm{R}_{0,t_1}-\bm{\Sigma}_{t_1})^{\top} \Mean \{(\bm{R}_{0,t_2}-\bm{\Sigma}_{t_2})|X_{0,t_2-1}\}\\&=&\Mean [\Mean\{ (\bm{R}_{0,t_1}-\bm{\Sigma}_{t_1}) |X_{0,t_1+2} \} ]^{\top} [\Mean \{(\bm{R}_{0,t_2}-\bm{\Sigma}_{t_2})|X_{0,t_2-1}\}],
\end{eqnarray*}
and hence
\begin{eqnarray}\label{prooflemma3eq11}
\Mean (\bm{R}_{0,t_1}-\bm{\Sigma}_{t_1})^{\top} (\bm{R}_{0,t_2}-\bm{\Sigma}_{t_2})=\Mean [\Mean\{ (\bm{R}_{0,t_1}-\bm{\Sigma}_{t_1}) |X_{0,t_1+2} \} ]^{\top} [\Mean \{(\bm{R}_{0,t_2}-\bm{\Sigma}_{t_2})|X_{0,t_2-1}\}].
\end{eqnarray}
Define $\bm{\Theta}_{1,t_1}(x)=\Mean (\bm{R}_{0,t_1}|X_{0,t_1+2}=x)$ and $\bm{\Theta}_{2,t_2}(x)=\Mean (\bm{R}_{0,t_2}|X_{0,t_2-1}=x)$. Let $\Mean^{X_{0,t}}$ denote the conditional expectation given $X_{0,t}$. With some calculations, we can show
\begin{eqnarray*}
	\bm{\Theta}_{1,t_1}(x)=\Mean \left[\left.\Mean^{X_{0,t_1+1}}  \bm{\xi}(X_{0,t_1},A_{0,t_1}) \{\bm{\xi}(X_{0,t_1},A_{0,t_1})-\gamma \bm{U}(X_{0,t_1+1}) \}^{\top}  \right|X_{0,t_1+2}=x\right]\\
	=\Mean\left[\left.\int_{y\in \mathbb{X}} \sum_{a\in \mathcal{A}} \bm{\xi}(y,a)\{\bm{\xi}(y,a)-\gamma \bm{U}(X_{0,t_1+1}) \}^{\top} \frac{b(a|y)q(X_{0,t_1+1}|y,a)\mu_{t_1}(y)}{\mu_{t_1+1}(X_{0,t_1+1})} dy \right|X_{0,t_1+2}=x\right]\\
	= \int_{y_1,y_2\in \mathbb{X}} \sum_{a\in \mathcal{A}} \bm{\xi}(y_1,a)\{\bm{\xi}(y_1,a)-\gamma \bm{U}(y_2) \}^{\top} b(a|y_1)q(y_2|y_1,a)\frac{\mu_{t_1}(y_1)q_X(x|y_2)}{\mu_{t_1+2}(x)} dy_1dy_2,
\end{eqnarray*}
and
\begin{eqnarray*}
	\bm{\Theta}_{2,t_2}(x)&=&\Mean \left[\left.\sum_{a\in \mathcal{A}} \bm{\xi}(X_{0,t_2},a)\{\bm{\xi}(X_{0,t_2},a)-\gamma  \bm{u}(X_{0,t_2},a)\}^{\top}b(a|X_{0,t_2})\right|X_{0,t_2-1}=x\right]\\
	&=&\int_{y\in \mathbb{X}} \sum_{a\in \mathcal{A}} \bm{\xi}(y,a)\{\bm{\xi}(y,a)-\gamma \bm{u}(y,a) \}^{\top} b(a|y)q_X(y|x)dy.
\end{eqnarray*}
It follows from \eqref{prooflemma3eq11} that
\begin{eqnarray*}
	\bm{v}^{\top} \Mean (\bm{R}_{0,t_1}-\bm{\Sigma})^{\top} (\bm{R}_{0,t_2}-\bm{\Sigma}) \bm{v}=\bm{v}^{\top} \Mean \bm{\Theta}_1^{\top}(X_{0,t_1+2}) \bm{\Theta}_2(X_{0,t_2-1})\bm{v}-\bm{v}^{\top}(\bm{\Sigma})^{\top} \bm{\Sigma} \bm{v}\\
	= \bm{v}^{\top} \Mean \bm{\Theta}_{1,t_1}^{\top}(X_{0,t_1+2}) \bm{\Theta}_{2,t_2}(X_{0,t_2-1})\bm{v}-\bm{v}^{\top} \{\Mean \bm{\Theta}_{1,t_1}^{\top}(X_{0,t_1+2})\}^{\top} \Mean \bm{\Theta}_{2,t_2}(X_{0,t_2-1}) \bm{v}\\
	=\sum_{j=1}^{mL} \Cov\{\bm{v}^{\top} \bm{\Theta}_{1,t_1,\cdot,j}(X_{0,t_1+2}), \bm{\Theta}_{2,t_2,j,\cdot}(X_{0,t_2-1}) \bm{v}\}
\end{eqnarray*}
where $\bm{\Theta}_{l,t,j,\cdot}(x)$ and $\bm{\Theta}_{l,t,\cdot,j}(x)$ denote the $j$-th row and $j$-column of $\bm{\Theta}_{l,t}(x)$, respectively. Let $\xi_j(\cdot,\cdot)$ be the $j$-th element of $\bm{\xi}(\cdot,\cdot)$. By Lemma \ref{lemma2} and the definitions of $\bm{\xi}$ and $\bm{U}$, 
\begin{eqnarray*}
	\sup_{\substack{  \bm{v}\in \mathbb{S}^{mL-1},x\in \mathbb{X} \\ a\in \mathcal{A} }} |\bm{\xi}^{\top}(x,a) \bm{v}|\preceq L^{1/2}\,\,\,\,\hbox{and}\,\,\,\,\sup_{\bm{v}\in \mathbb{S}^{mL-1},x\in \mathbb{X}} |\bm{U}^{\top}(x) \bm{v}|\preceq L^{1/2}. 
\end{eqnarray*}
It follows that
\begin{eqnarray*}
	&&\max_t\max_{j\in \{1,\dots,mL\}}\sup_{\substack{  \bm{v}\in \mathbb{S}^{mL-1},x\in \mathbb{X}}} |\bm{v}^{\top} \bm{\Theta}_{1,t,\cdot,j}(x)|\\
	&\preceq& \int_{y_1,y_2\in \mathbb{X}} \sum_{a\in \mathcal{A}} |\xi_{j}(y_1,a)| |\bm{v}^{\top} \{\bm{\xi}(y_1,a)-\gamma \bm{U}(y_2)\}|b(a|y_1) dy_1dy_2\\
	&\preceq& L^{1/2} \int_{y_1\in \mathbb{X}} \sum_{a\in \mathcal{A}} |\xi_{j}(y_1,a)|b(a|y_1)dy_1\le L^{1/2} \max_{j\in \{1,\dots,L\}} \int_{y\in \mathbb{X}} |\phi_{L,j}(y)|dy.
\end{eqnarray*}
Similarly to Lemma \ref{lemma2}, we can show $\max_{j\in \{1,\dots,L\}} \int_{y\in \mathbb{X}} |\phi_{L,j}(y)|dy\preceq L^{-1/2}$, and hence
\begin{eqnarray}\label{Theta1a}
\max_{j\in \{1,\dots,mL\}}\sup_{\substack{  \bm{v}\in \mathbb{S}^{mL-1},x\in \mathbb{X}}} |\bm{v}^{\top} \bm{\Theta}_{1,t,\cdot,j}(x)|\preceq 1.
\end{eqnarray}
It follows that
\begin{eqnarray*}
	&&\sup_{\substack{\bm{v}\in \mathbb{S}^{mL-1} }}|\bm{v}^{\top} \Mean (\bm{R}_{0,t_1}-\bm{\Sigma}_{t_1})^{\top} (\bm{R}_{0,t_2}-\bm{\Sigma}_{t_2}) \bm{v}|\\
	&\le& \sup_{\substack{\bm{v}\in \mathbb{S}^{mL-1} }} \sum_{j=1}^{mL} \int_{x\in \mathbb{X}}|\Mean \{\bm{\Theta}_{2,t,j,\cdot}(X_{0,t_2-1}) \bm{v}|X_{0,t_1+2}\}-\Mean \{\bm{\Theta}_{2,t,j,\cdot}(X_{0,t_2-1}) \bm{v}||\bm{v}^{\top} \bm{\Theta}_{1,\cdot,j}(x)|\mu(x)dx\\
	&\preceq&  \sup_{\substack{\bm{v}\in \mathbb{S}^{mL-1} }} \sum_{j=1}^{mL} \int_{x\in \mathbb{X}}|\Mean \{\bm{\Theta}_{2,t,j,\cdot}(X_{0,t_2-1}) \bm{v}|X_{0,t_1+2}\}-\Mean \{\bm{\Theta}_{2,t,j,\cdot}(X_{0,t_2-1}) \bm{v}|\mu(x)dx.
\end{eqnarray*}
Similar to \eqref{Theta1a}, we can show $\max_t\max_{j\in \{1,\dots,mL\}}\sup_{\substack{  \bm{v}\in \mathbb{S}^{mL-1},x\in \mathbb{X}}} | \bm{\Theta}_{2,j,\cdot}(x) \bm{v}|\preceq 1$. 
It follows from the definition of $\beta$-mixing coefficients and the geometric ergodicity that
\begin{eqnarray*}
	\sup_{\substack{\bm{v}\in \mathbb{S}^{mL-1} }}|\bm{v}^{\top} \Mean (\bm{R}_{0,t_1}-\bm{\Sigma}_{t_1})^{\top} (\bm{R}_{0,t_2}-\bm{\Sigma}_{t_2}) \bm{v}|\preceq \sum_{j=1}^{mL} \beta(t_2-t_1-3)\preceq L\rho^{t_2-t_1-3},
\end{eqnarray*}
where the above bound is uniform for any pair $(t_1,t_2)$ that satisfies $0\le t_1\le t_2-4$. Therefore, we have
\begin{eqnarray*}
	\sum_{\substack{0\le t_1<t_2\le T-1\\ |t_2-t_1|\ge 4 }}\sup_{\substack{\bm{v}\in \mathbb{S}^{mL-1} }}|\bm{v}^{\top} \Mean (\bm{R}_{0,t_1}-\bm{\Sigma}_{t_1})^{\top} (\bm{R}_{0,t_2}-\bm{\Sigma}_{t_2}) \bm{v}|\preceq L\sum_{\substack{0\le t_1<t_2\le T-1\\ |t_2-t_1|\ge 4 }} \rho^{t_2-t_1-3}\preceq LT. 
\end{eqnarray*}
Similarly, we can show
\begin{eqnarray*}
	\sum_{\substack{0\le t_1<t_2\le T-1\\ |t_2-t_1|\ge 4 }}\sup_{\substack{\bm{v}\in \mathbb{S}^{mL-1} }}|\bm{v}^{\top} \Mean (\bm{R}_{0,t_1}-\bm{\Sigma}_{t_1})^{\top} (\bm{R}_{0,t_2}-\bm{\Sigma}_{t_2}) \bm{v}|\preceq LT,
\end{eqnarray*}
and hence $\eta_3\preceq LT$. This together with \eqref{ineqn1} and \eqref{firstwoterms} yields that 
\begin{eqnarray}\label{immediatestep}
\sup_{\substack{\bm{v}\in \mathbb{S}^{mL-1}}} \Mean \bm{v}^{\top} \left\{ \sum_{t=0}^{T-1} (\bm{R}_{0,t}-\bm{\Sigma}) \right\}^{\top} \left\{ \sum_{t=0}^{T-1} (\bm{R}_{0,t}-\bm{\Sigma}) \right\} \bm{v}\preceq LT.
\end{eqnarray}
By Cauchy-Schwarz inequality, we obtain
\begin{eqnarray*}
	\sup_{\substack{\bm{v}\in \mathbb{S}^{mL-1}}} \Mean \left\| \left\{ \sum_{t=0}^{T-1} (\bm{R}_{0,t}-\bm{\Sigma})\right\}\mathbb{I}(\mathcal{A}_0)\bm{v}-\Mean \left\{\sum_{t=0}^{T-1} (\bm{R}_{0,t}-\bm{\Sigma})\mathbb{I}(\mathcal{A}_0) \right\} \bm{v} \right\|_2^2
	\preceq LT.
\end{eqnarray*}
Combining this with \eqref{prooflemma3eq9}, an application of the matrix Bernstein inequality \citep[Theorem 1.6 in][]{Tropp2012} yields that 
\begin{eqnarray*}
	\left\|\sum_{i=1}^n \left\{\sum_{t=0}^{T-1} (\bm{R}_{i,t}-\bm{\Sigma})\right\}\mathbb{I}(\mathcal{A}_i)-n \Mean \left\{\sum_{t=0}^{T-1} (\bm{R}_{0,t}-\bm{\Sigma})\right\}\mathbb{I}(\mathcal{A}_0)\right\|_2=O_p\{\sqrt{nTL}\log (nT)\},
\end{eqnarray*}
under the assumption that $L=o\{nT/\log^2 (nT)\}$. This together with \eqref{prooflemma3eq9} yields that
\begin{eqnarray}\label{matrixinequality2}
\left\|\sum_{i=1}^n \left\{\sum_{t=0}^{T-1} (\bm{R}_{i,t}-\bm{\Sigma})\right\}-n \Mean \left\{\sum_{t=0}^{T-1} (\bm{R}_{0,t}-\bm{\Sigma})\right\}\mathbb{I}(\mathcal{A}_0)\right\|_2=O_p\{\sqrt{nTL}\log (nT)\}.
\end{eqnarray}
By Cauchy-Schwarz inequality, we have for any $\bm{v}_1,\bm{v}_2\in \mathbb{R}^{mL}$ with $\|\bm{v}_1\|_2=\|\bm{v}_2\|_2=1$ that
\begin{eqnarray*}
	\left|\bm{v}_1^{\top} \Mean \left\{\sum_{t=0}^{T-1} (\bm{R}_{0,t}-\bm{\Sigma})\right\}\mathbb{I}(\mathcal{A}_0^c) \bm{v}_2\right|\le \sqrt{\Mean \left[ \bm{v}_1^{\top} \left\{\sum_{t=0}^{T-1} (\bm{R}_{0,t}-\bm{\Sigma})\right\} \bm{v}_2 \right]^2}\sqrt{\prob(\mathcal{A}_0^c)}\\
	\le \sqrt{\Mean \left\| \left\{\sum_{t=0}^{T-1} (\bm{R}_{0,t}-\bm{\Sigma})\right\} \bm{v}_2 \right\|_2^2}\sqrt{\prob(\mathcal{A}_0^c)}\preceq \sqrt{LT} n^{-3/2} T^{-1}=O(n^{-1}),
\end{eqnarray*}
by \eqref{immediatestep0}, \eqref{immediatestep} and the condition that $L\ll Tn/\log^2 (Tn)$. This together with \eqref{matrixinequality2} yields that
\begin{eqnarray*}
	\left\|\sum_{i=1}^n \left\{\sum_{t=0}^{T-1} (\bm{R}_{i,t}-\Mean \bm{R}_{0,t})\right\}\right\|_2=O_p\{\sqrt{nTL}\log (nT)\},
\end{eqnarray*}
and hence $\|\widehat{\bm{\Sigma}}-\bm{\Sigma}\|_2=O_p\{L^{1/2} (nT)^{-1/2} \log (nT) \}$. 

\noindent \textit{Part 3}: In either scenario, we have shown $\|\widehat{\bm{\Sigma}}-\bm{\Sigma}\|_2=O_p\{L^{1/2} (nT)^{-1/2} \log (nT) \}$. Under the condition that $L=o\{nT/\log (nT)\}$, we have $\|\widehat{\bm{\Sigma}}-\bm{\Sigma}\|_2=o_p(1)$. By definition, this implies that $\|\widehat{\bm{\Sigma}}-\bm{\Sigma}\|_2\le \bar{c}/6$, with probability approaching $1$. In \textit{Part 1}, we have shown that $\bm{v}^{\top} \bm{\Sigma} \bm{v}\ge \bar{c} \|\bm{v}\|_2/2$, for any $\bm{v}\in \mathbb{R}^{mL}$. By Cauchy-Schwarz inequality, the following event occurs with probability approaching $1$,
\begin{eqnarray*}
	\bm{v}^{\top} \widehat{\bm{\Sigma}} \bm{v}\ge \bar{c} \|\bm{v}\|_2^2/3,\,\,\,\,\,\,\,\,\forall \bm{v}\in \mathbb{R}^{mL}. 
\end{eqnarray*}
Using similar arguments in \textit{Part 1}, this implies $\widehat{\bm{\Sigma}}$ is invertible and satisfies $\|\widehat{\bm{\Sigma}}^{-1}\|_2\le 3 \bar{c}^{-1}$, with probability tending to $1$. Therefore
\begin{eqnarray*}
	\|\widehat{\bm{\Sigma}}^{-1}-\bm{\Sigma}^{-1}\|_2=\|\widehat{\bm{\Sigma}}^{-1}(\widehat{\bm{\Sigma}}-\bm{\Sigma})\bm{\Sigma}^{-1}\|_2\le \|\widehat{\bm{\Sigma}}^{-1}\|_2 \|\widehat{\bm{\Sigma}}-\bm{\Sigma}\|_2 \|\bm{\Sigma}^{-1}\|_2\le 6 \bar{c}^{-2} \|\widehat{\bm{\Sigma}}-\bm{\Sigma}\|_2,
\end{eqnarray*}
with probability tending to $1$. Since $\|\widehat{\bm{\Sigma}}-\bm{\Sigma}\|_2=O_p\{L^{1/2} (nT)^{-1/2} \log (nT) \}$, we obtain $\|\widehat{\bm{\Sigma}}^{-1}-\bm{\Sigma}^{-1}\|_2=O_p\{L^{1/2} (nT)^{-1/2} \log (nT) \}$. The proof is hence completed. 

\noindent \textit{Part 4: }
It suffices to show
\begin{eqnarray*}
	\eta_4\equiv \max_t \sup_{\substack{\bm{v}_1, \bm{v}_2 \in \mathbb{S}^{mL-1}}} \left|\bm{v}_1^{\top} (\Mean \bm{R}_{0,t}^{**}) \bm{v}_2 \right|\preceq 1.
\end{eqnarray*}
With some calculations, we have
\begin{eqnarray*}
	&&\eta_4\le \max_t\sup_{\substack{\bm{v}_1, \bm{v}_2 \in \mathbb{S}^{mL-1}}} \sum_{a\in \mathcal{A}} \Mean |\bm{v}_1^{\top} \bm{\xi}(X_{0,t},a)| |\bm{v}_2^{\top} \bm{\xi}(X_{0,t},a)| b(a|X_{0,t})\\
	&+&\max_t\sup_{\substack{\bm{v}_1, \bm{v}_2 \in \mathbb{S}^{mL-1}}} \sum_{a\in \mathcal{A}} \Mean |\bm{v}_1^{\top} \bm{\xi}(X_{0,t},a)| |\bm{v}_2^{\top} \bm{u}(X_{0,t},a)| b(a|X_{0,t})\le \frac{3}{2}\eta_4^{(1)}+\frac{1}{2}\eta_4^{(2)},
\end{eqnarray*}
by Cauchy-Schwarz inequality, where
\begin{eqnarray*}
	\eta_4^{(1)}=\max_t\sup_{\bm{v}\in \mathbb{S}^{mL-1}} \sum_{a\in \mathcal{A}} \Mean |\bm{v}^{\top} \bm{\xi}(X_{0,t},a)|^2b(a|X_{0,t}),\\
	\eta_4^{(2)}=\max_t\sup_{\bm{v}\in \mathbb{S}^{mL-1}} \sum_{a\in \mathcal{A}} \Mean |\bm{v}^{\top} \bm{u}(X_{0,t},a)|^2b(a|X_{0,t}).
\end{eqnarray*}
By definition, we have
\begin{eqnarray*}
	\eta_4^{(1)}=\max_t\lambda_{\max}\left\{\sum_{a\in \mathcal{A}}\Mean \bm{\xi}(X_{0,t},a)\bm{\xi}^{\top}(X_{0,t},a)b(a|X_{0,t}) \right\}.
\end{eqnarray*}
Since the matrix $\sum_{a\in \mathcal{A}} \bm{\xi}(X_{0,t},a)\bm{\xi}^{\top}(X_{0,t},a)b(a|X_{0,t})$ is block diagonal with the main-diagonal blocks $\{\Phi_L(X_{0,t})\Phi_L(X_{0,t})^{\top} b(j|X_{0,t}) \}_{j=1,\dots,m}$. By Lemma \ref{lemma2} and Condition A2, we can show $\eta_4^{(1)}\preceq 1$. As for $\eta_4^{(2)}$,  we have
\begin{eqnarray*}
	\eta_4^{(2)}\le \max_t\sup_{\bm{v}\in \mathbb{S}^{mL-1}} \Mean |\bm{v}^{\top} \bm{U}(X_{0,t+1})|^2=\lambda_{\max}\left\{\Mean \bm{U}(X_{0,t+1}) \bm{U}^{\top}(X_{0,t+1}) \right\}\\\le \sup_{\substack{v_1,\dots,v_m\in \mathbb{R}^L\\ \|v_1\|_2^2+\cdots+\|v_m\|_2^2=1 }} \sum_{1\le j_1,j_2\le m} v_{j_1}^{\top} \Mean \Phi_L(X_{0,t+1})\Phi_L(X_{0,t+1})^{\top} v_{j_2} \pi(j_1|X_{0,t+1})\pi(j_2|X_{0,t+1})\\
	\le \lambda_{\max} \left\{\Mean \Phi_L(X_{0,t+1})\Phi_L(X_{0,t+1}^*)^{\top}\right\},
\end{eqnarray*} 
where the first inequality follows from Jensen's inequality. By Lemma \ref{lemma2}, we can similarly show that $\eta_4^{(2)}\preceq 1$. Thus, we obtain $\|\bm{\Sigma}\|_2=\eta_4\preceq 1$. 
The proof is hence completed.

\subsection{Proof of Lemma \ref{lemma4}}
The proof of Lemma \ref{lemma4} is very similar to that of Lemma \ref{lemma3}. 
By Condition A3(i), we obtain $\lambda_{\min}(T^{-1}\sum_{t=0}^{T-1} \Mean \bm{\xi}_{0,t}\bm{\xi}_{0,t}^{\top})\ge \bar{c}$, when $T$ is bounded. A4(iii) further implies that $$\lambda_{\min}\left[\int_{x\in \mathbb{X}}\left\{\sum_{a\in \mathcal{A}}\bm{\xi}(x,a)\bm{\xi}^{\top}(x,a)b(a|x)\right\}\right]\ge \bar{c},$$ as $T\to \infty$. Using similar arguments as in the first part of the proof of Lemma \ref{lemma3}, we can show that
\begin{eqnarray}\label{prooflemma3eq16}
\lambda_{\min}\left(\frac{1}{T} \sum_{t=0}^{T-1} \Mean \bm{\xi}_{0,t}\bm{\xi}_{0,t}^{\top}\right)\ge \frac{\bar{c}}{2},
\end{eqnarray} 
as either $n\to \infty$, or $T\to \infty$. 
Using similar arguments in the third part of the proof of Lemma \ref{lemma3}, we can show that
\begin{eqnarray}\label{prooflemma3eq15}
\left\| \frac{1}{nT}\sum_{i=1}^n \sum_{t=0}^{T-1}\bm{\xi}_{i,t}\bm{\xi}_{i,t}^{\top}- \frac{1}{T}\sum_{t=0}^{T-1}\Mean\bm{\xi}_{0,t}\bm{\xi}_{0,t}^{\top}\right\|_2=O_p\{L^{1/2}(nT)^{-1/2}\log (nT) \}.
\end{eqnarray}
Since $L=o\{\sqrt{nT}/\log (nT) \}$, it follows from \eqref{prooflemma3eq16} and \eqref{prooflemma3eq15} that the following event occurs with probability tending to $1$,
\begin{eqnarray*}
	\lambda_{\min}\left(\frac{1}{nT} \sum_{i=1}^n\sum_{t=0}^{T-1}  \bm{\xi}_{i,t}\bm{\xi}_{i,t}^{\top}\right)\ge \lambda_{\min}\left(\frac{1}{T} \sum_{t=0}^{T-1} \Mean \bm{\xi}_{0,t}\bm{\xi}_{0,t}^{\top}\right)\\
	-\left\| \frac{1}{nT}\sum_{i=1}^n \sum_{t=0}^{T-1}\bm{\xi}_{i,t}\bm{\xi}_{i,t}^{\top}- \frac{1}{T}\sum_{t=0}^{T-1}\Mean\bm{\xi}_{0,t}\bm{\xi}_{0,t}^{\top}\right\|_2
	\ge \frac{\bar{c}}{2}-\frac{\bar{c}}{6}=\frac{\bar{c}}{3}. 
\end{eqnarray*}
It remains to show $\lambda_{\max}\{(nT)^{-1} \sum_{i=1}^n \sum_{t=0}^{T-1} \bm{\xi}_{i,t} \bm{\xi}_{i,t}^{\top}\}=O_p(1)$ and
\begin{eqnarray}\label{prooflemma3eq17}
\lambda_{\max}\left(\frac{1}{T}\sum_{t=0}^{T-1}\Mean \bm{\xi}_{0,t}\bm{\xi}_{0,t}^{\top}\right)=O(1),
\end{eqnarray}
Suppose \eqref{prooflemma3eq17} holds. 
By \eqref{prooflemma3eq15} and the condition that $L=o\{\sqrt{nT}/\log (nT) \}$, we have $\lambda_{\max}\{(nT)^{-1} \sum_{i=1}^n \sum_{t=0}^{T-1} \bm{\xi}_{i,t} \bm{\xi}_{i,t}^{\top}\}=O_p(1)$. Thus, it suffices to show \eqref{prooflemma3eq17}. This can be proven using similar arguments in Part 2 of the proof of Lemma \ref{lemma3}. We omit the details for brevity. 
The proof is hence completed.  

\subsection{Proof of Lemma \ref{lemma5}}
It follows from Cauchy-Schwarz inequality and triangle inequality that
\begin{eqnarray*}
	&&\left\| \int_{x\in \mathbb{X}} \bm{U}(x)\mathbb{G}(dx) \right\|_2^2=\sum_{a=1}^m \left\| \int_{x\in \mathbb{X}} \Phi_L(x)\pi(a|x)\mathbb{G}(dx) \right\|_2^2\\
	&\ge& m^{-1} \left\{ \sum_{a=1}^m \left\| \int_{x\in \mathbb{X}} \Phi_L(x)\pi(a|x)\mathbb{G}(dx) \right\|_2\right\}^2\ge m^{-1} \left\| \int_{x\in \mathbb{X}} \sum_{a=1}^m \Phi_L(x)\pi(a|x)\mathbb{G}(dx) \right\|_2^2\\
	&=&m^{-1} \left\| \int_{x\in \mathbb{X}} \Phi_L(x)\mathbb{G}(dx)\right\|_2^2.
\end{eqnarray*}
The proof is hence completed. 

\subsection{Proof of Theorem \ref{thm2}}
Without loss of generality, we assume $n=K_n n_{\min}$ and $T=K_T T_{\min}$ such that $|\mathcal{I}_k|=n_{\min}T_{\min}$ for any $k$. Under the given conditions, $K$ is bounded. 
Similar to Lemma \ref{lemma3}, we can show under A4* that
\begin{eqnarray}\label{maxSigmak}
\max_{2\le k\le K}\|\bm{\Sigma}_{\widehat{\pi}_{\bar{\mathcal{I}}_{k-1}}}^{-1}\|_2\le 2\bar{c}^{-1}.
\end{eqnarray}

We next bound the difference between $\bm{\Sigma}_{\widehat{\pi}_{\bar{\mathcal{I}}_{k-1}}}$ and $\widehat{\bm{\Sigma}}_{\mathcal{I}_k,\widehat{\pi}_{\bar{\mathcal{I}}_{k-1}}}$. 
Consider the scenario where $T$ is bounded first. Since $T_{\min}=T$, the data are divided according to the trajectories they belong to. Thus, for any $k=2,\dots,K$, variables $\{(X_{i,t},A_{i,t},Y_{i,t},X_{i,t+1})\}_{(i,t)\in \mathcal{I}_k}$ are independent of $\{(X_{i,t},A_{i,t},Y_{i,t},X_{i,t+1})\}_{(i,t)\in \bar{\mathcal{I}}_{k-1}}$. Let
\begin{eqnarray*}
	\bm{\Sigma}_{\widehat{\pi}_{\bar{\mathcal{I}}_{k-1}}}=\Mean [\widehat{\bm{\Sigma}}_{\mathcal{I}_k,\widehat{\pi}_{\bar{\mathcal{I}}_{k-1}}}|\{(X_{i,t},A_{i,t},Y_{i,t},X_{i,t+1})\}_{(i,t)\in \bar{\mathcal{I}}_{k-1}}].
\end{eqnarray*}
Using similar arguments in Part 2 of the proof of Lemma \ref{lemma3}, we can show
\begin{eqnarray}\label{maxSigmakdiff}
\max_{2\le k\le K}\|\bm{\Sigma}_{\widehat{\pi}_{\bar{\mathcal{I}}_{k-1}}}-\widehat{\bm{\Sigma}}_{\mathcal{I}_k,\widehat{\pi}_{\bar{\mathcal{I}}_{k-1}}}\|_2\preceq \sqrt{(nT)^{-1} L\log (nT)},
\end{eqnarray}
with probability at least $1-O(n^{-2}T^{-2})=1-o(1)$.

Now let us consider the scenario where $T\to \infty$. 
For $k=1,\dots,K$, define $(i_0(k),t_0(k))$ to be the tuple in $\mathcal{I}_k$ such that $i\ge i_0(k),t\ge t_0(k)$ for any $(i,t)\in \mathcal{I}_k$. Then, we have
\begin{eqnarray*}
	\mathcal{I}_k=\{(i,t):i_0(k)\le i<i_0(k)+n_{\min},t_0(k)\le t<t_0(k)+T_{\min} \}. 
\end{eqnarray*}
Consider any $k\in \{2,\dots,K\}$ with $t_0(k)=0$, $\{(X_{i,t},A_{i,t},Y_{i,t},X_{i,t+1})\}_{(i,t)\in \mathcal{I}_k}$ are independent of  $\{(X_{i,t},A_{i,t},Y_{i,t},X_{i,t+1})\}_{(i,t)\in \bar{\mathcal{I}}_{k-1}}$. Using similar arguments in Part 2 of the proof of Lemma \ref{lemma3}, we can show wpa1 that,
\begin{eqnarray}\label{maxSigmakdiff2}
\max_{\substack{k\in \{2,\dots,K\}\\ t_0(k)=0 }} \|\widehat{\bm{\Sigma}}_{\mathcal{I}_k,\widehat{\pi}_{\bar{\mathcal{I}}_{k-1}}}-\bm{\Sigma}_{\widehat{\pi}_{\bar{\mathcal{I}}_{k-1}}}\|_2\preceq \sqrt{\frac{L}{nT}}\log (nT).
\end{eqnarray}
Consider $k\in \{2,\dots,K\}$ with $t_0(k)>0$. We decompose $\widehat{\bm{\Sigma}}_{\mathcal{I}_k,\widehat{\pi}_{\bar{\mathcal{I}}_{k-1}}}$ as
\begin{eqnarray*}
	\underbrace{\frac{1}{n_{\min}T_{\min}}\sum_{i=i_0(k)}^{i_0(k)+n_{\min}-1} \bm{\xi}_{i,t_0(k)}\{\bm{\xi}_{i,t_0(k)}-\gamma\bm{U}_{\widehat{\pi}_{\bar{\mathcal{I}}_{k-1},i,t_0(k)+1}}\}^{\top} }_{\widehat{\bm{\Sigma}}_{\mathcal{I}_k,\widehat{\pi}_{\bar{\mathcal{I}}_{k-1}}}^{(1)} }\\
	+\underbrace{\frac{1}{n_{\min}T_{\min}} \sum_{\substack{(i,t)\in \mathcal{I}_k\\ t>t_0(k) }} [\bm{\xi}_{i,t}\{\bm{\xi}_{i,t}-\gamma\bm{U}_{\widehat{\pi}_{\bar{\mathcal{I}}_{k-1}},i,t} \}^{\top}}_{\widehat{\bm{\Sigma}}_{\mathcal{I}_k,\widehat{\pi}_{\bar{\mathcal{I}}_{k-1}}}^{(2)}}.
\end{eqnarray*}
It follows that
\begin{eqnarray*}
	\max_{\substack{k\in \{2,\dots,K\}\\ t_0(k)>0 }} \|\widehat{\bm{\Sigma}}_{\mathcal{I}_k,\widehat{\pi}_{\bar{\mathcal{I}}_{k-1}}}-\bm{\Sigma}_{\widehat{\pi}_{\bar{\mathcal{I}}_{k-1}}}\|_2\le \max_{\substack{k\in \{2,\dots,K\}\\ t_0(k)>0 }} \|\widehat{\bm{\Sigma}}^{(1)}_{\mathcal{I}_k,\widehat{\pi}_{\bar{\mathcal{I}}_{k-1}}}-\bm{\Sigma}^{(1)}_{\widehat{\pi}_{\bar{\mathcal{I}}_{k-1}}}\|_2\\
	+\max_{\substack{k\in \{2,\dots,K\}\\ t_0(k)>0 }} \|\widehat{\bm{\Sigma}}^{(2)}_{\mathcal{I}_k,\widehat{\pi}_{\bar{\mathcal{I}}_{k-1}}}-\bm{\Sigma}^{(2)}_{\widehat{\pi}_{\bar{\mathcal{I}}_{k-1}}}\|_2.
\end{eqnarray*}

\noindent \textit{Error bound for $\max_k \|\widehat{\bm{\Sigma}}^{(1)}_{\mathcal{I}_k,\widehat{\pi}_{\bar{\mathcal{I}}_{k-1}}}-\bm{\Sigma}^{(1)}_{\widehat{\pi}_{\bar{\mathcal{I}}_{k-1}}}\|_2$: }Given $\{(X_{j,t},A_{j,t},Y_{j,t},X_{j,t+1})\}_{(j,t)\in \bar{\mathcal{I}}_{k-1}}$, \\$(A_{i_0(k),t_0(k)},X_{i_0(k),t_0(k)+1}),\cdots,(A_{i_0(k)+n_{\min}-1,t_0(k)},X_{i_0(k)+n_{\min}-1,t_0(k)+1})$ conditionally independent. Using the matrix concentration inequality, we can show wpa1 that
\begin{eqnarray}\label{maxSigmakdiff2.3}
\max_{\substack{k\in \{2,\dots,K\}\\ t_0(k)>0 }} \|\widehat{\bm{\Sigma}}^{(1)}_{\mathcal{I}_k,\widehat{\pi}_{\bar{\mathcal{I}}_{k-1}}}-\bm{\Sigma}^{(1)}_{\widehat{\pi}_{\bar{\mathcal{I}}_{k-1}}}\|_2\preceq \sqrt{\frac{L}{nT}\log(nT)}.
\end{eqnarray}

\noindent \textit{Error bound for $\max_k \|\widehat{\bm{\Sigma}}^{(2)}_{\mathcal{I}_k,\widehat{\pi}_{\bar{\mathcal{I}}_{k-1}}}-\bm{\Sigma}^{(2)}_{\widehat{\pi}_{\bar{\mathcal{I}}_{k-1}}}\|_2$: }Given $\{(X_{j,t},A_{j,t},Y_{j,t},X_{j,t+1})\}_{(j,t)\in \bar{\mathcal{I}}_{k-1}}$, \\ the sets of variables $\{(X_{i_0(k),t},A_{i_0(k),t},Y_{i_0(k),t},X_{i_0(k),t+1}): t_0(k)+1\le t<t_0(k)+T_{\min}\},\cdots,\\\{(X_{i_0(k)+n_{\min}-1,t},A_{i_0(k)+n_{\min}-1,t},Y_{i_0(k)+n_{\min}-1,t},X_{i_0(k)+n_{\min}-1,t+1}): t_0(k)+1\le t<t_0(k)+T_{\min}\}$ are conditionally independent. Moreover, for any $i$ such that $(i,t_0(k))\in \mathcal{I}_k$, the density function of $X_{i,t_0(k)+1}$ conditional on $\{(X_{j,t},A_{j,t},Y_{j,t},X_{j,t+1})\}_{(j,t)\in \bar{\mathcal{I}}_{k-1}}$ is uniformly bounded under A3. Using similar arguments in bounding $\|\widehat{\bm{\Sigma}}-\bm{\Sigma}\|_2$ in Part 2 of the proof of Lemma \ref{lemma3}, we can show wpa1 that
\begin{eqnarray}\label{maxSigmakdiff2.5}
\max_{\substack{k\in \{2,\dots,K\}\\ t_0(k)>0 }} \|\widehat{\bm{\Sigma}}^{(2)}_{\mathcal{I}_k,\widehat{\pi}_{\bar{\mathcal{I}}_{k-1}}}-\bm{\Sigma}^{(2)}_{\widehat{\pi}_{\bar{\mathcal{I}}_{k-1}}}\|_2\preceq \sqrt{\frac{L}{nT}}\log(nT).
\end{eqnarray}
Combining \eqref{maxSigmakdiff2.3} with \eqref{maxSigmakdiff2.5} and \eqref{maxSigmakdiff2}, we obtain wpa1 that
\begin{eqnarray}\label{maxSigmak2.5}
\max_{\substack{k\in \{2,\dots,K\} }} \|\widehat{\bm{\Sigma}}_{\mathcal{I}_k,\widehat{\pi}_{\bar{\mathcal{I}}_{k-1}}}-\bm{\Sigma}_{\widehat{\pi}_{\bar{\mathcal{I}}_{k-1}}}\|_2\preceq \sqrt{\frac{L}{nT}}\log(nT).
\end{eqnarray}

To summarize, we have shown that
\begin{eqnarray*}
	\max_{k\in \{2,\dots,K\}} \|\bm{\Sigma}_{\widehat{\pi}_{\bar{\mathcal{I}}_{k-1}}}^{-1}\|_2\le 2\bar{c}^{-1}\,\,\,\,\hbox{and}\,\,\,\,\max_{\substack{k\in \{2,\dots,K\} }} \|\widehat{\bm{\Sigma}}_{\mathcal{I}_k,\widehat{\pi}_{\bar{\mathcal{I}}_{k-1}}}-\bm{\Sigma}_{\widehat{\pi}_{\bar{\mathcal{I}}_{k-1}}}\|_2\preceq \sqrt{\frac{L}{nT}}\log (nT),
\end{eqnarray*}
wpa1, regardless of whether $T$ is bounded or not. Under the given conditions, we have $\sqrt{L/(nT)}\log (nT)=o(1)$. Using similar arguments in the proof of Lemma \ref{lemma3}, we can show wpa1 that
\begin{eqnarray}\label{maxSigmak3}
\max_{k\in \{2,\dots,K\}} \|\widehat{\bm{\Sigma}}_{\mathcal{I}_k,\widehat{\pi}_{\bar{\mathcal{I}}_{k-1}}}^{-1}\|_2\le \frac{3}{\bar{c}},\max_{\substack{k\in \{2,\dots,K\} }} \|\widehat{\bm{\Sigma}}_{\mathcal{I}_k,\widehat{\pi}_{\bar{\mathcal{I}}_{k-1}}}^{-1}-\bm{\Sigma}_{\widehat{\pi}_{\bar{\mathcal{I}}_{k-1}}}^{-1}\|_2\preceq \sqrt{\frac{L}{nT}}\log (nT).
\end{eqnarray}

Notice that $\widehat{\pi}=\widehat{\pi}_{\bar{\mathcal{I}}(K)}$. By Lemma \ref{lemma1}, we have $Q(\widehat{\pi}_{\bar{\mathcal{I}}(k)};\cdot,a)\in \Lambda(p,c')$ for any $k\in \{1,\dots,K\}$. Using similar arguments in the proof of Theorem 12.8 of \cite{Schumaker1981} and proof of Proposition 5 of \cite{Meyer1992}, there exist  some vectors $\{\beta^*_{\widehat{\pi}_{\bar{\mathcal{I}}(k)},a}\}_{a\in \mathcal{A},1\le k\le K}$ that satisfy
\begin{eqnarray}\label{biasbound2}
\max_{1\le k\le K}\sup_{x\in \mathbb{X},a\in \mathcal{A}} |Q(\widehat{\pi}_{\bar{\mathcal{I}}(k)};x,a)-\Phi_L^{\top}(x)\beta_{\widehat{\pi}_{\bar{\mathcal{I}}(k)},a}^*|\le CL^{-p/d},
\end{eqnarray}
for some constant $C>0$. Similar to \eqref{maxresidual}, we have by \eqref{biasbound2} that
\begin{eqnarray}\label{maxresidual1}
\max_{(i,t)\in \mathcal{I}_k} |r_{i,t}|\le 2CL^{-p/d},\,\,\,\,\forall k=2,\dots,K,
\end{eqnarray}
where 
\begin{eqnarray*}
	r_{i,t}&=&\gamma \sum_{a\in \mathcal{A}}\{\Phi_L^{\top}(X_{i,t+1})\beta_{\widehat{\pi}_{\bar{\mathcal{I}}_{k-1}},a}^*-Q(\widehat{\pi}_{\bar{\mathcal{I}}_{k-1}};X_{i,t+1},a)\}\widehat{\pi}_{\bar{\mathcal{I}}_{k-1}}(a|X_{i,t+1})\\
	&-&\{\Phi_L^{\top}(X_{i,t}) \beta_{\widehat{\pi}_{\bar{\mathcal{I}}_{k-1}},A_{i,t}}^*-Q(\widehat{\pi}_{\bar{\mathcal{I}}_{k-1}};X_{i,t},A_{i,t})\}.
\end{eqnarray*}
Similar to the proof of Theorem \ref{thm1}, we have by \eqref{maxSigmak3} and \eqref{maxresidual1} that
\begin{eqnarray*}
	&&\widehat{\bm{\beta}}_{\mathcal{I}_k,\widehat{\pi}_{\bar{\mathcal{I}}_{k-1}}}-\bm{\beta}^*_{\widehat{\pi}_{\bar{\mathcal{I}}_{k-1}}}=\underbrace{\bm{\Sigma}_{\widehat{\pi}_{\bar{\mathcal{I}}_{k-1}}}^{-1}\left( \frac{K}{nT}\sum_{(i,t)\in \mathcal{I}_k} \bm{\xi}_{i,t}\varepsilon_{i,t} \right)}_{\zeta_{1,k}}+O(L^{-p/d})+O_p\{L (nT)^{-1}\log (nT)\},
\end{eqnarray*}
where 
\begin{eqnarray*}
	\varepsilon_{i,t}=Y_{i,t}+\gamma \sum_{a\in \mathcal{A}}Q(\widehat{\pi}_{\bar{\mathcal{I}}_{k-1}};X_{i,t+1},a)\widehat{\pi}_{\bar{\mathcal{I}}_{k-1}}(a|X_{i,t+1})-Q(\widehat{\pi}_{\bar{\mathcal{I}}_{k-1}};X_{i,t},A_{i,t}),
\end{eqnarray*}
for any $(i,t)\in \mathcal{I}_k$. 

To prove the asymptotic normality of $\sqrt{nT(K-1)/K}\widetilde{\sigma}^{-1}(\mathbb{G})\{\widetilde{V}(\mathbb{G})-V(\widehat{\pi};\mathbb{G})\}$, it suffices to show
\begin{eqnarray}\label{thm2CLT}
\sqrt{\frac{nTK}{K-1}} \sum_{k=2}^K \frac{\widehat{V}_{\mathcal{I}_k}(\widehat{\pi}_{\bar{\mathcal{I}}_{k-1}};\mathbb{G})-V(\widehat{\pi};\mathbb{G})}{\widehat{\sigma}_{\mathcal{I}_k}(\widehat{\pi}_{\bar{\mathcal{I}}_{k-1}};\mathbb{G})}\stackrel{d}{\to} N(0,1). 
\end{eqnarray}
Using similar arguments in the proof of Theorem 1, 
we can show
\begin{eqnarray}\label{importanteq1}
&&\left|\sqrt{\frac{nTK}{K-1}} \sum_{k=2}^K \frac{\widehat{V}_{\mathcal{I}_k}(\widehat{\pi}_{\bar{\mathcal{I}}_{k-1}};\mathbb{G})-V(\widehat{\pi};\mathbb{G})}{\widehat{\sigma}_{\mathcal{I}_k}(\widehat{\pi}_{\bar{\mathcal{I}}_{k-1}};\mathbb{G})}\right.\\\nonumber
&-&\left.\sqrt{\frac{nTK}{K-1}} \sum_{k=2}^K \frac{V(\widehat{\pi}_{\bar{\mathcal{I}}_{k-1}};\mathbb{G})+\{\int_x \bm{U}_{\widehat{\pi}_{\bar{\mathcal{I}}_{k-1}}}(x)\mathbb{G}(dx)\}^{\top} \zeta_{1,k}-V(\widehat{\pi};\mathbb{G})}{\widehat{\sigma}_{\mathcal{I}_k}(\widehat{\pi}_{\bar{\mathcal{I}}_{k-1}};\mathbb{G})} \right|=o_p(1).
\end{eqnarray}

Notice that $\widehat{\pi}=\widehat{\pi}_{\bar{\mathcal{I}}_K}$. Under A5, we have
\begin{eqnarray*}
	&&\frac{1}{K-1}\sum_{k=2}^K \Mean |V(\widehat{\pi}_{\bar{\mathcal{I}}_{k-1}};\mathbb{G})-V(\widehat{\pi};\mathbb{G})|\\
	&\le& \frac{1}{K-1}\sum_{k=2}^K \Mean |V(\widehat{\pi}_{\bar{\mathcal{I}}_{k-1}};\mathbb{G})-V(\pi^*;\mathbb{G})|+\Mean |V(\widehat{\pi};\mathbb{G})-V(\pi^*;\mathbb{G})|\\
	&\le& O(1) \left\{ \frac{1}{K-1}\sum_{k=1}^{K-1} (nT)^{-b_0}k^{-b_0}K^{b_0} +(nT)^{-b_0} \right\},
\end{eqnarray*}
where $O(1)$ denotes some positive constant. Since $\sum_{k=1}^{K-1} k^{-b_0}\le 1+\int_1^{K} x^{-b_0}dx\preceq K^{1-b_0}$, we obtain that
\begin{eqnarray*}
	\frac{1}{K-1}\sum_{k=2}^K \Mean |V(\widehat{\pi}_{\bar{\mathcal{I}}_{k-1}};\mathbb{G})-V(\widehat{\pi};\mathbb{G})|=O\{(nT)^{-b_0}\},
\end{eqnarray*}
and hence
\begin{eqnarray*}
	\sqrt{\frac{nTK}{K-1}}\sum_{k=2}^K \Mean |V(\widehat{\pi}_{\bar{\mathcal{I}}_{k-1}};\mathbb{G})-V(\widehat{\pi};\mathbb{G})|=O\{(nT)^{1/2-b_0}\}=o\left\{\left\|\int_x \Phi_L(x)dx \right\|_2\right\},
\end{eqnarray*}
By A6. By Markov's inequality, we obtain that
\begin{eqnarray*}
	\left\|\int_x \Phi_L(x)dx \right\|_2^{-1} \frac{\sqrt{nTK}}{K-1}\sum_{k=2}^K |V(\widehat{\pi}_{\bar{\mathcal{I}}_{k-1}};\mathbb{G})-V(\widehat{\pi};\mathbb{G})|=o_p(1). 
\end{eqnarray*}
Similar to Lemma \ref{lemma5}, we can show that for any $k\in \{2,\dots,K\}$,
\begin{eqnarray*}
	\left\|\int_x \Phi_L(x)dx \right\|_2^{-1}\ge m^{-1/2} \left\|\int_x \bm{U}_{\widehat{\pi}_{\bar{\mathcal{I}}_{k-1}}}(x)dx\right\|_2^{-1}.
\end{eqnarray*}
It follows that
\begin{eqnarray}\label{someequation}
\sqrt{\frac{nTK}{K-1}}\sum_{k=2}^K \left\|\int_x \bm{U}_{\widehat{\pi}_{\bar{\mathcal{I}}_{k-1}}}(x)dx\right\|_2^{-1}|V(\widehat{\pi}_{\bar{\mathcal{I}}_{k-1}};\mathbb{G})-V(\widehat{\pi};\mathbb{G})|=o_p(1). 
\end{eqnarray}
Using similar arguments in the proof of Theorem \ref{thm1}, we can show
\begin{eqnarray*}
	\max_{k\in \{2,\dots,K\}}\left|\frac{\widehat{\sigma}_{\mathcal{I}_k}^2(\widehat{\pi}_{\bar{\mathcal{I}}_{k-1}};\mathbb{G})}{\sigma^2(\widehat{\pi}_{\bar{\mathcal{I}}_{k-1}};\mathbb{G})}-1\right|=o_p(1).
\end{eqnarray*}
Since
\begin{eqnarray*}
	\left|\frac{\widehat{\sigma}_{\mathcal{I}_k}(\widehat{\pi}_{\bar{\mathcal{I}}_{k-1}};\mathbb{G})}{\sigma(\widehat{\pi}_{\bar{\mathcal{I}}_{k-1}};\mathbb{G})}-1\right|=\left|\frac{\widehat{\sigma}_{\mathcal{I}_k}^2(\widehat{\pi}_{\bar{\mathcal{I}}_{k-1}};\mathbb{G})}{\sigma^2(\widehat{\pi}_{\bar{\mathcal{I}}_{k-1}};\mathbb{G})}-1\right|\left|\frac{\widehat{\sigma}_{\mathcal{I}_k}(\widehat{\pi}_{\bar{\mathcal{I}}_{k-1}};\mathbb{G})}{\sigma(\widehat{\pi}_{\bar{\mathcal{I}}_{k-1}};\mathbb{G})}+1\right|^{-1},
\end{eqnarray*} 
we obtain
\begin{eqnarray}\label{maxsigmadiff}
\max_{k\in \{2,\dots,K\}}\left|\frac{\widehat{\sigma}_{\mathcal{I}_k}(\widehat{\pi}_{\bar{\mathcal{I}}_{k-1}};\mathbb{G})}{\sigma(\widehat{\pi}_{\bar{\mathcal{I}}_{k-1}};\mathbb{G})}-1\right|=o_p(1).
\end{eqnarray}
Similar to \eqref{lowerbound1.5}, we can show there exists some constant $c_6>0$, such that the following occurs wpa1, 
\begin{eqnarray*}
	\sigma(\widehat{\pi}_{\bar{\mathcal{I}}_{k-1}};\mathbb{G})\ge c_6\left\| \int_{x} \bm{U}_{\widehat{\pi}_{\bar{\mathcal{I}}_{k-1}}}(x)\mathbb{G}(dx) \right\|_2,\,\,\,\,\forall k\in \{2,\dots,K\}.
\end{eqnarray*}
This together with \eqref{maxsigmadiff} yields that
\begin{eqnarray*}
	\widehat{\sigma}_{\mathcal{I}_k}(\widehat{\pi}_{\bar{\mathcal{I}}_{k-1}};\mathbb{G})\ge c_6\left\| \int_{x} \bm{U}_{\widehat{\pi}_{\bar{\mathcal{I}}_{k-1}}}(x)\mathbb{G}(dx) \right\|_2,\,\,\,\,\forall k\in \{2,\dots,K\},
\end{eqnarray*}
wpa1. In view of \eqref{someequation}, we obtain
\begin{eqnarray*}
	\left|\sqrt{\frac{nTK}{K-1}} \sum_{k=2}^K \frac{V(\widehat{\pi}_{\bar{\mathcal{I}}_{k-1}};\mathbb{G})-V(\widehat{\pi};\mathbb{G})}{\widehat{\sigma}_{\mathcal{I}_k}(\widehat{\pi}_{\bar{\mathcal{I}}_{k-1}};\mathbb{G})} \right|=o_p(1).
\end{eqnarray*}
Combining this together with \eqref{importanteq1}, we obtain that
\begin{eqnarray*}
	\sqrt{\frac{nTK}{K-1}} \sum_{k=2}^K \frac{\widehat{V}_{\mathcal{I}_k}(\widehat{\pi}_{\bar{\mathcal{I}}_{k-1}};\mathbb{G})-V(\widehat{\pi};\mathbb{G})}{\widehat{\sigma}_{\mathcal{I}_k}(\widehat{\pi}_{\bar{\mathcal{I}}_{k-1}};\mathbb{G})}
	-\sqrt{\frac{nTK}{K-1}} \sum_{k=2}^K \frac{\{\int_x \bm{U}_{\widehat{\pi}_{\bar{\mathcal{I}}_{k-1}}}(x)\mathbb{G}(dx)\}^{\top} \zeta_{1,k}}{\widehat{\sigma}_{\mathcal{I}_k}(\widehat{\pi}_{\bar{\mathcal{I}}_{k-1}};\mathbb{G})}=o_p(1).
\end{eqnarray*}
To prove \eqref{thm2CLT}, it suffices to show
\begin{eqnarray}\label{thm3CLT}
\sqrt{\frac{nTK}{K-1}} \sum_{k=2}^K \frac{\{\int_x \bm{U}_{\widehat{\pi}_{\bar{\mathcal{I}}_{k-1}}}(x)\mathbb{G}(dx)\}^{\top} \zeta_{1,k}}{\widehat{\sigma}_{\mathcal{I}_k}(\widehat{\pi}_{\bar{\mathcal{I}}_{k-1}};\mathbb{G})}\stackrel{d}{\to} N(0,1).
\end{eqnarray}
The LHS of \eqref{thm3CLT} can be further decomposed as
\begin{eqnarray*}
	&&\underbrace{\sqrt{\frac{nTK}{K-1}} \sum_{k=2}^K \frac{\{\int_x \bm{U}_{\widehat{\pi}_{\bar{\mathcal{I}}_{k-1}}}(x)\mathbb{G}(dx)\}^{\top} \zeta_{1,k}}{\sigma(\widehat{\pi}_{\bar{\mathcal{I}}_{k-1}};\mathbb{G})}}_{\eta_7}\\
	&+&\underbrace{\sqrt{\frac{nTK}{K-1}} \sum_{k=2}^K \frac{\{\int_x \bm{U}_{\widehat{\pi}_{\bar{\mathcal{I}}_{k-1}}}(x)\mathbb{G}(dx)\}^{\top} \zeta_{1,k}}{\sigma(\widehat{\pi}_{\bar{\mathcal{I}}_{k-1}};\mathbb{G})} \left\{\frac{\sigma(\widehat{\pi}_{\bar{\mathcal{I}}_{k-1}};\mathbb{G})}{\widehat{\sigma}_{\mathcal{I}_k}(\widehat{\pi}_{\bar{\mathcal{I}}_{k-1}};\mathbb{G})}-1 \right\}}_{\eta_8}.
\end{eqnarray*}
In the following, we show $\eta_7\stackrel{d}{\to} N(0,1)$.  Based on \eqref{someequation}, one can show $\eta_8\stackrel{P}{\to} 0$. Assertion \eqref{thm3CLT} thus follows from Slutsky's theorem. 

Notice that $\eta_7$ equals
\begin{eqnarray*}
	\eta_7=\sqrt{\frac{K}{nT(K-1)}} \sum_{k=2}^K\sum_{(i,t)\in \mathcal{I}_k} \frac{\{\int_x \bm{U}_{\widehat{\pi}_{\bar{\mathcal{I}}_{k-1}}}(x)\mathbb{G}(dx)\}^{\top} \bm{\Sigma}_{\widehat{\pi}_{\bar{\mathcal{I}}_{k-1}}}^{-1} \bm{\xi}_{i,t}\varepsilon_{i,t} }{\sigma(\widehat{\pi}_{\bar{\mathcal{I}}_{k-1}};\mathbb{G})}.
\end{eqnarray*}
For any $1\le g\le nT$, there exists some integer $k(g)$ that satisfies $\{k(g)-1\} n_{\min}T_{\min}+1\le g\le k(g) n_{\min}T_{\min}$. Let $t(g)$ and $i(g)$ be the integers that satisfy
\begin{eqnarray*}
	g-\{k(g)-1\} n_{\min} T_{\min}=\{i(g)-i_0(k(g))\}T_{\min}+t(g)-t_0(k(g))+1.
\end{eqnarray*}
Let $\mathcal{F}^{(0)}=\{X_{1,0},A_{1,0}\}$. Then we iteratively define $\{\mathcal{F}^{(g)}\}_{1\le g\le nT}$ as follows:
\begin{eqnarray*}
	&&\mathcal{F}^{(g)}=\mathcal{F}^{(g-1)}\cup \{Y_{i(g),T-1},X_{i(g),T} \},\,\,\,\,\hbox{if}\,\,\,\,t(g+1)>0~\&~ t(g)=T-1,\\
	&&\mathcal{F}^{(g)}=\mathcal{F}^{(g-1)}\cup \{Y_{i(g), t(g)},X_{i(g),t(g)+1},A_{i(g),t(g)+1}\},\,\,\,\,\hbox{if}\,\,\,\,t(g+1)>0~\&~ t(g)<T-1,\\
	&&\mathcal{F}^{(g)}=\mathcal{F}^{(g-1)}\cup \{Y_{i(g),T-1},X_{i(g),T},X_{i(g+1),0},A_{i(g+1),0} \},\,\,\,\,\hbox{if}\,\,\,\,t(g+1)=0~\&~ t(g)=T-1,\\
	&&\mathcal{F}^{(g)}=\mathcal{F}^{(g-1)}\cup \{Y_{i(g), t(g)},X_{i(g),t(g)+1},A_{i(g),t(g)+1},X_{i(g+1),0},A_{i(g+1),0}\},\,\,\,\,\hbox{otherwise}.
\end{eqnarray*}
Let $\bm{\xi}^{(g)}=\bm{\xi}_{i(g),t(g)}$ and $\varepsilon^{(g)}=\varepsilon_{i(g),t(g)}$. We rewrite $\eta_7$ as
\begin{eqnarray*}
	\eta_7=\sqrt{\frac{K}{nT(K-1)}} \sum_{g=nT/K+1}^{nT} \frac{\{\int_x \bm{U}_{\widehat{\pi}_{\bar{\mathcal{I}}(k(g)-1)}}(x)\mathbb{G}(dx)\}^{\top} \bm{\Sigma}_{\widehat{\pi}_{\bar{\mathcal{I}}(k(g)-1)}}^{-1} \bm{\xi}^{(g)}\varepsilon^{(g)} }{\sigma(\widehat{\pi}_{\bar{\mathcal{I}}(k(g)-1)};\mathbb{G})}.
\end{eqnarray*}
One can show that $\eta_7$ forms a mean-zero martingale with respect to the filtration $\{\sigma(\mathcal{F}^{(g)})\}_{g\ge nT/K}$. Using similar arguments in the proof of Theorem \ref{thm1}, we can show that
\begin{eqnarray}\label{MCLTeq1}
\max_{g\in \{ nT/K+1,\dots,nT \} } \left|\frac{\{\int_x \bm{U}_{\widehat{\pi}_{\bar{\mathcal{I}}(k(g)-1)}}(x)\mathbb{G}(dx)\}^{\top} \bm{\Sigma}_{\widehat{\pi}_{\bar{\mathcal{I}}(k(g)-1)}}^{-1} \bm{\xi}^{(g)}\varepsilon^{(g)} }{\sigma(\widehat{\pi}_{\bar{\mathcal{I}}(k(g)-1)};\mathbb{G})}\right|\stackrel{P}{\to} 0
\end{eqnarray}
Moreover, we have
\begin{eqnarray*}
	&&\left|\frac{K}{nT(K-1)}  \sum_{g=nT/K+1}^{nT} \frac{[\{\int_x \bm{U}_{\widehat{\pi}_{\bar{\mathcal{I}}(k(g)-1)}}(x)\mathbb{G}(dx)\}^{\top} \bm{\Sigma}_{\widehat{\pi}_{\bar{\mathcal{I}}(k(g)-1)}}^{-1} \bm{\xi}^{(g)}\varepsilon^{(g)}]^2 }{\sigma^2(\widehat{\pi}_{\bar{\mathcal{I}}(k(g)-1)};\mathbb{G})}-1 \right|\\
	&\le& \max_{k\in \{2,\dots,K\}} \left| \frac{K}{nT} \sum_{(i,t)\in \mathcal{I}_k} \frac{[\{\int_x \bm{U}_{\widehat{\pi}_{\bar{\mathcal{I}}_{k-1}}}(x)\mathbb{G}(dx)\}^{\top} \bm{\Sigma}_{\widehat{\pi}_{\bar{\mathcal{I}}_{k-1}}}^{-1} \bm{\xi}_{i,t}\varepsilon_{i,t}]^2 }{\sigma^2(\widehat{\pi}_{\bar{\mathcal{I}}_{k-1}};\mathbb{G})}-1 \right|\\
	&\le&  \max_{k\in \{2,\dots,K\}} \frac{\|\{\int_x \bm{U}_{\widehat{\pi}_{\bar{\mathcal{I}}_{k-1}}}(x)\mathbb{G}(dx)\}^{\top} \bm{\Sigma}_{\widehat{\pi}_{\bar{\mathcal{I}}_{k-1}}}^{-1}\|_2^2\|\widehat{\bm{\Omega}}^*_{\mathcal{I}_k, \widehat{\pi}_{\bar{\mathcal{I}}_{k-1}} }-\bm{\Omega}_{ \widehat{\pi}_{\bar{\mathcal{I}}_{k-1}} }\|_2}{\sigma^2(\widehat{\pi}_{\bar{\mathcal{I}}_{k-1}};\mathbb{G})},
\end{eqnarray*}
where $\widehat{\bm{\Omega}}^*_{\mathcal{I}_k, \widehat{\pi}_{\bar{\mathcal{I}}_{k-1}} }=|\mathcal{I}_k|^{-1} \sum_{(i,t)\in \mathcal{I}_k} \bm{\xi}_{i,t}\bm{\xi}_{i,t}^\top \varepsilon_{i,t}^2$.

Similar to the proof of Theorem \ref{thm1}, we can show $\max_{k\in \{2,\dots,K\}} \|\widehat{\bm{\Omega}}^*_{\mathcal{I}_k, \widehat{\pi}_{\bar{\mathcal{I}}_{k-1}} }-\bm{\Omega}_{ \widehat{\pi}_{\bar{\mathcal{I}}_{k-1}} }\|_2=o_p(1)$. Similar to \eqref{lowerbound1}, we can show there exists some constant $c_7>0$ such that
\begin{eqnarray}\label{lowerbound4}
\max_{k\in \{2,\dots,K\}} \frac{\|\{\int_x \bm{U}_{\widehat{\pi}_{\bar{\mathcal{I}}_{k-1}}}(x)\mathbb{G}(dx)\}^{\top} \bm{\Sigma}_{\widehat{\pi}_{\bar{\mathcal{I}}_{k-1}}}^{-1}\|_2^2}{\sigma^2(\widehat{\pi}_{\bar{\mathcal{I}}_{k-1}};\mathbb{G})}\le c_7.
\end{eqnarray}
It follows that
\begin{eqnarray}\label{MCLTeq2}
\left|\frac{K}{nT(K-1)}  \sum_{g=nT/K+1}^{nT} \frac{[\{\int_x \bm{U}_{\widehat{\pi}_{\bar{\mathcal{I}}(k(g)-1)}}(x)\mathbb{G}(dx)\}^{\top} \bm{\Sigma}_{\widehat{\pi}_{\bar{\mathcal{I}}(k(g)-1)}}^{-1} \bm{\xi}^{(g)}\varepsilon^{(g)}]^2 }{\sigma^2(\widehat{\pi}_{\bar{\mathcal{I}}(k(g)-1)};\mathbb{G})}-1 \right|\stackrel{P}{\to} 0.
\end{eqnarray}
Using a martingale central limit theorem for triangular arrays \citep[Corollary 2.8 of][]{McLeish1974}, we have by \eqref{MCLTeq1} and \eqref{MCLTeq2} that $\eta_7\stackrel{d}{\to} N(0,1)$. 
The proof is hence completed. 

\subsection{Proof of Theorem \ref{thm3}}\label{secproofthm3}
Based on the discussions in Section \ref{secconvergeVpi}, it suffices to show
\begin{eqnarray*}
	\Mean \int_x \sum_{a\in \mathcal{A}} Q^{\tiny{opt}}(x,a) \{\pi^{\tiny{opt}}(a|x)-\widehat{\pi}_{\mathcal{I}}(a|x)\}\mathbb{G}(dx)=O(|\mathcal{I}|^{-b_*(1+\alpha)}),
\end{eqnarray*}
and
\begin{eqnarray}\label{proofthm3eq1}
\Mean \int_x \sum_{a\in \mathcal{A}} Q^{\tiny{opt}}(x,a) \{\pi^{\tiny{opt}}(a|x)-\widehat{\pi}_{\mathcal{I}}(a|x)\}dx=O(|\mathcal{I}|^{-b_*(1+\alpha)}).
\end{eqnarray}

We only prove \eqref{proofthm3eq1} for brevity. Under the given conditions, we have $\prob(\mathcal{A}_0)\ge 1-O(|\mathcal{I}|^{-\kappa})$, where
\begin{eqnarray*}
	\mathcal{A}_0=\left\{\sup_{x\in \mathbb{X},a\in \mathcal{A}} |\widehat{Q}_{\mathcal{I}}(x,a)-Q^{\tiny{opt}}(x,a)|\le C|\mathcal{I}|^{-b_*}\right\}.
\end{eqnarray*}
Notice that
\begin{eqnarray}\nonumber
&&\Mean \int_x \sum_{a\in \mathcal{A}} Q^{\tiny{opt}}(x,a) \{\pi^{\tiny{opt}}(a|x)-\widehat{\pi}_{\mathcal{I}}(a|x)\}dx\\ \label{proofthm3eq2}
&=& \Mean \int_x \sum_{a\in \mathcal{A}} Q^{\tiny{opt}}(x,a) \{\pi^{\tiny{opt}}(a|x)-\widehat{\pi}_{\mathcal{I}}(a|x)\}dx\mathbb{I}(\mathcal{A}_0^c)\\ \nonumber
&+&\Mean \int_x \sum_{a\in \mathcal{A}} Q^{\tiny{opt}}(x,a) \{\pi^{\tiny{opt}}(a|x)-\widehat{\pi}_{\mathcal{I}}(a|x)\}dx\mathbb{I}(\mathcal{A}_0).
\end{eqnarray}
Under A1, $Q^{\tiny{opt}}$ is uniformly bounded. Therefore, the first term on the RHS of \eqref{proofthm3eq2} is upper bounded by $O\{\prob(\mathcal{A}_0^c)\}=O(|\mathcal{I}|^{-\kappa})$. Since $\kappa$ can be chosen arbitrarily large, it suffices to show
\begin{eqnarray}\label{proofthm3eq0}
\Mean \int_x \sum_{a\in \mathcal{A}} Q^{\tiny{opt}}(x,a) \{\pi^{\tiny{opt}}(a|x)-\widehat{\pi}_{\mathcal{I}}(a|x)\}dx\mathbb{I}(\mathcal{A}_0)=O\{|\mathcal{I}|^{-b_*(1+\alpha)}\}.
\end{eqnarray}

For any $x\in \mathbb{X}$, suppose
\begin{eqnarray}\label{proofthm3eq3}
\max_{a} Q^{\tiny{opt}}(x,a)-\max_{a\in \mathcal{A}-\argmax_{a'}Q^{\tiny{opt}}(x,a')} Q^{\tiny{opt}}(x,a)>2C|\mathcal{I}|^{-b_*}. 
\end{eqnarray} 
Under the event defined in $\mathcal{A}_0$, we have
\begin{eqnarray*}
	\max_{a} \widehat{Q}_{\mathcal{I}}(x,a)-\max_{a\in \mathcal{A}-\argmax_{a'}Q^{\tiny{opt}}(x,a')} \widehat{Q}_{\mathcal{I}}(x,a)>0,
\end{eqnarray*}
and hence
\begin{eqnarray*}
	\{a\in \mathcal{A}:\widehat{\pi}(a|x)=1\}\subseteq \argmax_{a\in \mathcal{A}} Q^{\tiny{opt}}(x,a). 
\end{eqnarray*}
Thus, we have
\begin{eqnarray*}
	\sum_{a\in \mathcal{A}} Q^{\tiny{opt}}(x,a) \{\pi^{\tiny{opt}}(a|x)-\widehat{\pi}_{\mathcal{I}}(a|x)\}=0,
\end{eqnarray*}
when \eqref{proofthm3eq3} holds. Let $\mathbb{X}_*$ denote the set of $x$ that satisfies \eqref{proofthm3eq3}. It follows that
\begin{eqnarray}\label{proofthm3eq4}
&&\Mean \int_x \sum_{a\in \mathcal{A}} Q^{\tiny{opt}}(x,a) \{\pi^{\tiny{opt}}(a|x)-\widehat{\pi}_{\mathcal{I}}(a|x)\}dx\mathbb{I}(\mathcal{A}_0)\\\nonumber
&=&\Mean \int_x \sum_{a\in \mathcal{A}} Q^{\tiny{opt}}(x,a) \{\pi^{\tiny{opt}}(a|x)-\widehat{\pi}_{\mathcal{I}}(a|x)\}\mathbb{I}(x\in \mathbb{X}_*^c)dx\mathbb{I}(\mathcal{A}_0).
\end{eqnarray}
Let $\widehat{a}_{\mathcal{I}}(x)=\sargmax_{a} \widehat{Q}_{\mathcal{I}}(a,x)$. Similarly, we can show the event $\widehat{a}_{\mathcal{I}}(x)\notin \argmax_{a\in \mathcal{A}} Q^{\tiny{opt}}(x,a)$ occurs only when
\begin{eqnarray}\label{somethinginthm3}
\max_{a} Q^{\tiny{opt}}(x,a)-Q^{\tiny{opt}}(x,\widehat{a}_{\mathcal{I}}(x))\le 2C|\mathcal{I}|^{-b_*}. 
\end{eqnarray}
Since $\max_{a} Q^{\tiny{opt}}(x,a)-Q^{\tiny{opt}}(x,\widehat{a}_{\mathcal{I}}(x))=\sum_{a} Q^{\tiny{opt}}(x,a) \{\pi^{\tiny{opt}}(a|x)-\widehat{\pi}_{\mathcal{I}}(a|x)\}$, we obtain
\begin{eqnarray*}
	&&\Mean \int_x \sum_{a\in \mathcal{A}} Q^{\tiny{opt}}(x,a) \{\pi^{\tiny{opt}}(a|x)-\widehat{\pi}_{\mathcal{I}}(a|x)\}\mathbb{I}(x\in \mathbb{X}_*^c)dx\mathbb{I}(\mathcal{A}_0)\\
	&\le &2C|\mathcal{I}|^{-b_*} \int_x \mathbb{I}(x\in \mathbb{X}_*^c)dx=2C|\mathcal{I}|^{-b_*} O(|\mathcal{I}|^{-b_*\alpha})=O(|\mathcal{I}|^{-b_*(1+\alpha)}),
\end{eqnarray*}
where the first equality follows from A5. Combining this together with \eqref{proofthm3eq4} yields \eqref{proofthm3eq0}. The proof is hence completed. 

\subsection{Proof of Theorem \ref{thm4}}\label{secproofthm4}
For a given $\varepsilon>0$, let $\mathcal{A}_*=\{\max_{a} Q^{\tiny{opt}}(x,a)-Q^{\tiny{opt}}(x,\widehat{a}_{\mathcal{I}}(x))\le \varepsilon\}$. Notice that
\begin{eqnarray}\label{proofthm4eq1}
&&\Mean \int_x \sum_{a\in \mathcal{A}} Q^{\tiny{opt}}(x,a) \{\pi^{\tiny{opt}}(a|x)-\widehat{\pi}_{\mathcal{I}}(a|x)\}dx\\ \nonumber
&=& \Mean \int_x \sum_{a\in \mathcal{A}} Q^{\tiny{opt}}(x,a) \{\pi^{\tiny{opt}}(a|x)-\widehat{\pi}_{\mathcal{I}}(a|x)\}\mathbb{I}(x\in \mathcal{A}_*)dx\\ \nonumber
&+&\Mean \int_x \sum_{a\in \mathcal{A}} Q^{\tiny{opt}}(x,a) \{\pi^{\tiny{opt}}(a|x)-\widehat{\pi}_{\mathcal{I}}(a|x)\}\mathbb{I}(x\in\mathcal{A}_*^c)dx.
\end{eqnarray}
Using similar arguments in the proof of Theorem \ref{thm3}, we can show
\begin{eqnarray}\nonumber
\Mean \int_x \sum_{a\in \mathcal{A}} Q^{\tiny{opt}}(x,a) \{\pi^{\tiny{opt}}(a|x)-\widehat{\pi}_{\mathcal{I}}(a|x)\}\mathbb{I}(x\in \mathcal{A}_*)dx\le \varepsilon \Mean \int_x \mathbb{I}(x\in \mathcal{A}_*)dx\\\label{proofthm4eq2} \le \varepsilon \lambda\left\{ x: \max_{a} Q^{\tiny{opt}}(x,a)-\max_{a\in \mathcal{A}-\argmax_{a'}Q^{\tiny{opt}}(x,a')} Q^{\tiny{opt}}(x,a)\le \varepsilon \right\}=O(\varepsilon^{1+\alpha}). 
\end{eqnarray}

Moreover, similar to \eqref{somethinginthm3}, we can show the event $\widehat{a}_{\mathcal{I}}(x)\notin \argmax_{a\in \mathcal{A}} Q^{\tiny{opt}}(x,a)$ occurs only when
\begin{eqnarray*}
	\max_{a} Q^{\tiny{opt}}(x,a)-Q^{\tiny{opt}}(x,\widehat{a}_{\mathcal{I}}(x))\le 2\max_{a\in \mathcal{A}} |\widehat{Q}_{\mathcal{I}}(x,a)-Q^{\tiny{opt}}(x,a)|.
\end{eqnarray*}
It follows that
\begin{eqnarray*}
	&&\Mean \int_x \sum_{a\in \mathcal{A}} Q^{\tiny{opt}}(x,a) \{\pi^{\tiny{opt}}(a|x)-\widehat{\pi}_{\mathcal{I}}(a|x)\}\mathbb{I}(x\in\mathcal{A}_*^c)dx\\
	&\le& \Mean \int_x \frac{4\max_{a\in \mathcal{A}}|\widehat{Q}_{\mathcal{I}}(x,a)-Q^{\tiny{opt}}(x,a)|^2}{\sum_{a\in \mathcal{A}} Q^{\tiny{opt}}(x,a) \{\pi^{\tiny{opt}}(a|x)-\widehat{\pi}_{\mathcal{I}}(a|x)\}}\mathbb{I}(x\in\mathcal{A}_*^c)dx\\
	&\le& \frac{4}{\varepsilon} \Mean \int_x \max_{a\in \mathcal{A}}|\widehat{Q}_{\mathcal{I}}(x,a)-Q^{\tiny{opt}}(x,a)|^2dx
	\le \frac{4}{\varepsilon} \Mean \int_x \sum_{a\in \mathcal{A}}|\widehat{Q}_{\mathcal{I}}(x,a)-Q^{\tiny{opt}}(x,a)|^2dx\\
	&=&O(\varepsilon^{-1}|\mathcal{I}|^{-2b_*}).
\end{eqnarray*}
Combining this together with \eqref{proofthm4eq1} and \eqref{proofthm4eq2} yields that
\begin{eqnarray*}
	\Mean \int_x \sum_{a\in \mathcal{A}} Q^{\tiny{opt}}(x,a) \{\pi^{\tiny{opt}}(a|x)-\widehat{\pi}_{\mathcal{I}}(a|x)\}dx=O(\varepsilon^{1+\alpha})+O(\varepsilon^{-1} |\mathcal{I}|^{-2b_*} ). 
\end{eqnarray*}
The proof is hence completed by setting $\varepsilon=|\mathcal{I}|^{-2b_*/(2+\alpha)}$. 


\end{document}